# CiteFusion: An Ensemble Framework for Citation Intent Classification Harnessing Dual-Model Binary Couples and SHAP Analyses

**Lorenzo Paolini** (ORCID: 0009-0003-3803-4011)
Department of Computer Science and Engineering, University of Bologna, Bologna, Italy

**Sahar Vahdati** (ORCID: 0000-0002-7171-169X)
Nature-inspired machine intelligence group, SCaDS.AI center, Technical University of Dresden, Germany
Institute for Applied Computer Science, InfAI - Dresden, Germany

**Angelo Di Iorio** (ORCID: 0000-0002-6893-7452)
Department of Computer Science and Engineering, University of Bologna, Bologna, Italy

**Robert Wardenga** (ORCID: 0009-0004-3317-6122)
Institute for Applied Computer Science, InfAI - Dresden, Germany

**Ivan Heibi** (ORCID: 0000-0001-5366-5194)
Research Centre for Open Scholarly Metadata, Department of Classical Philology and Italian Studies, University of Bologna, Bologna, Italy
Digital Humanities Advanced Research Centre (/DH.arc), Department of Classical Philology and Italian Studies, University of Bologna, Bologna, Italy

**Silvio Peroni** (ORCID: 0000-0003-0530-4305)
Research Centre for Open Scholarly Metadata, Department of Classical Philology and Italian Studies, University of Bologna, Bologna, Italy
Digital Humanities Advanced Research Centre (/DH.arc), Department of Classical Philology and Italian Studies, University of Bologna, Bologna, Italy

***Contacts*:** lorenzo.paolini11@unibo.it

## Abstract

Understanding the motivations underlying scholarly citations is essential to evaluate research impact and promote transparent scholarly communication. This study introduces *CiteFusion*, an ensemble framework designed to address the multi-class Citation Intent Classification task on two benchmark datasets: SciCite and ACL-ARC. The framework employs a *one-vs-all* decomposition of the multi-class task into class-specific binary subtasks, leveraging complementary pairs of SciBERT and XLNet models, independently tuned, for each citation intent. The outputs of these base models are aggregated through a feedforward neural network meta-classifier to reconstruct the original classification task. To enhance interpretability, *SHAP* (*SHapley Additive exPlanations*) is employed to analyze token-level contributions, and interactions among base models, providing transparency into the classification dynamics of CiteFusion, and insights about the kind of misclassifications of the ensemble. In addition, this work investigates the semantic role of structural context by incorporating section titles, as framing devices, into input sentences, assessing their positive impact on classification accuracy. CiteFusion ultimately demonstrates robust performance in imbalanced and data-scarce scenarios: experimental results show that *CiteFusion* achieves state-of-the-art performance, with Macro-F1 scores of 89.60% on SciCite, and 76.24% on ACL-ARC. Furthermore, to ensure interoperability and reusability, citation intents from both datasets schemas are mapped to *Citation Typing Ontology* (CiTO) object properties, highlighting some overlaps. Finally, we describe and release a web-based application that classifies citation intents leveraging the *CiteFusion* models developed on SciCite.

# 1 Introduction

Research evaluation is crucial in scholarly communication, as it ensures the quality, relevance, and impact of scientific contributions while fostering an environment of accountability and continuous improvement. It enables scholars and institutions to identify significant advancements and allocate resources to areas with a higher potential for innovation and societal benefit, but also to promote transparency, collaboration, and accessibility – key principles of the *Open Science* movement.

Citation analysis and bibliometrics occupy a central role in this evaluation process (Pride, 2022). However, many researchers have criticized the use of bibliometrics in research assessment, particularly due to the paradoxical nature underlying the abuse of metrics that employ citation counts as proxies for research quality. Indeed, there is insufficient evidence to demonstrate a connection between research quality and citation rates (Pride, 2022). Kuan Lai (2019) also highlights the widespread use of the Journal Impact Factor (JIF) and the h-index, two bibliometric indicators that rely heavily on citation counts, as research quality proxies. Such citation-count-based metrics are used to evaluate researchers, publications, and journals (Aksnes et al., 2019), staying unclear on the differentiation behind the types of citations. Indeed, citations may serve a variety of purposes, ranging from the reuse of a methodology to the mere acknowledgment of prior work (Cohan et al., 2019). This kind of functional distinction is referred to as the *intent* – or *function* – of a citation. The task of understanding and assigning the correct intent to a citation and its context[1] is known in literature as *Citation Intent Classification (CIC)*, represented in ***Figure 1***.

## 1.1 Motivation

Differentiating the functions that citations may serve is thus instrumental in providing more comprehensive and meaningful analyses in research assessment related fields (Small, 2018), and developing tools capable of retrieving influential papers, beyond citation counts, is fundamentally important also to promote more conscious research (Aksnes et al., 2019). Other possibilities within the CIC field are related to the development of applications for enhanced information retrieval, document summarization (Pride, 2022), and finally citations occupy a central role also in studies related to the evolution of scientific fields (Jurgens et al., 2018) where they are employed to construct citation networks and to frame different periods.

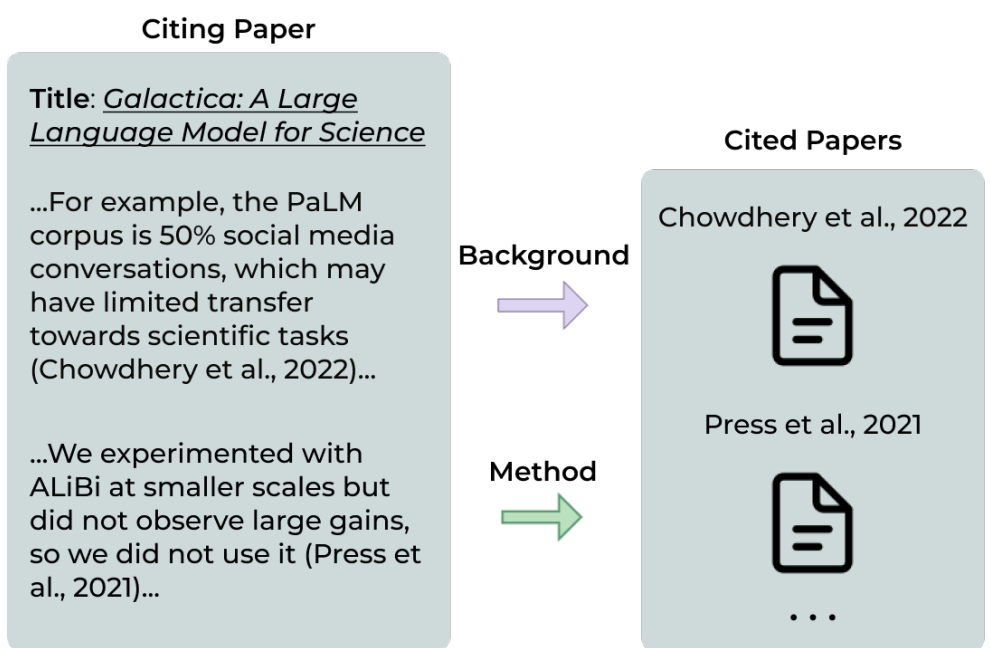

***Figure 1.*** *Example of an entity (citing paper) with two different in-text citations, each referring to distinct entities (cited papers) for different intents: Background and Method.*

Recognizing the importance to meaningfully distinguish between citation intents to advance scholarly communication also highlights the need for automated methods to accomplish this. Indeed, as highlighted by Pride (2022), "*the sheer volume of new research now being produced on an annual basis is far beyond the capacity of a single researcher to investigate even the narrowest of domains without effective search tools*", and this huge amount of new research and consequent citations being produced makes manual classification extremely time consuming and difficult (Lauscher et al., 2021). Instead, automated computational systems can efficiently process large datasets, enabling the discovery of useful insights into citation behaviors and revealing patterns or trends that might be overlooked in manual analyses. In this context, *Machine Learning* (ML) solutions become pivotal.

[1] A citation context refers to the textual segment immediately surrounding a bibliographic reference, providing the linguistic and rhetorical framework through which the cited work is integrated in the discourse.
For instance, in the sentence: "*These results are in contrast with the findings of Santos et al. (16), who reported a significant association between low sedentary time and healthy CVF among Portuguese*", the phrases preceding and following the citation constitute the citation context, while the reference itself – in red – is the bibliographic identifier.

## 1.2 Methodological Overview

Building on recent progress in ML and Natural Language Processing (NLP), in this work we explore how *Pre-trained Language Models* (PLMs) and *Ensemble Strategies* (ES) can enhance performances on *SciCite* (Cohan et al., 2019) and *ACL-ARC* (Jurgens et al., 2018)[2], two benchmark datasets for CIC characterized by a skewed data distribution. We decompose the multi-class CIC problem proposed by the two datasets into multiple binary one-vs-all (OVA) subtasks (Galar et al., 2011), pairing a domain-specialized pretrained language model (PLM) – *SciBERT* (Beltagy et al., 2019) – and a general-purpose PLM – *XLNet* (Yang et al., 2019) – as base classifiers for each class. Each PLM within each couple is independently fine-tuned and produces the probability scores for its assigned binary task. Therefore, every citation context is analyzed from both scientific and general language perspectives across all classes. The positive probabilities from all these models are then concatenated into a single feature vector, which is used as input for a Feed Forward Neural Network (FFNN) meta-classifier that produces the final multi-class prediction. The combination of base PLMs and FFNN aggregator results in ***CiteFusion***, an *Ensemble Classifier* (EC). Alternative aggregation methods, both supervised and unsupervised, are also explored, and original dataset labels are mapped to standardized object properties from the *Citation Typing Ontology* (CiTO), designed by Peroni and Shotton (2012), to enhance reusability.

In this study, we also employ granularly evaluated training loops to overcome the overfitting problem that usually arise when dealing with PLMs fine-tuning. In addition, since full-parameter fine-tuning with PLMs requires substantial time and computational resources, we also utilize mixed-precision to reduce them. We therefore perform computational instability analyses to address both the impact of these techniques and the instability caused by the skewed distributions of the datasets on the reproducibility of *CiteFusion*. Finally, to improve the interpretability and reinforce the reliability on our findings, we incorporate *Explainable AI* (*XAI*) techniques to give a better understanding of the classification dynamics of our models, and to highlight which features, in the form of input tokens, shape the perception of each intent according to different PLM architectures. This methodological overview will be further detailed in *Section 3*.

### 1.2.1 Section Titles as Contextual Features

While prior research highlights structural cues – i.e., *section titles* – as integral components of multitask learning frameworks (Cohan et al., 2019; Qi et al., 2023; Zhang et al., 2022), we argue that the semantic role of these elements warrants further investigation. Specifically, we investigate whether incorporating the explicit section titles – beyond their syntactic function and without explicit markers – into the classification pipeline can enhance a model's ability to accurately perceive and categorize citations. Informed by *frame theory* (Flusberg et al., 2024; Goffman, 1974), we hypothesize that prepending section titles may serve as a framing device: when raw contextual signals, such as individual words or citation context elements, are ambiguous, a model may rely on the inherent semantic associations of section titles (e.g., "*Methodology*" or "*Results*") as contextual cues to infer citation intents and shape its perception of unclear signals. This approach introduces a novel dimension for improving classification accuracy and interpretability by explicitly encoding the functional role of sections as interpretive frames within citation contexts.

To assess the influence of section titles as semantical components and, therefore, framing devices, we develop two ECs for each of the two datasets, resulting in four models: two of them trained with section titles included within the input citation sentences (WS setting), and two of them trained to solve the task through raw citation contexts (WoS setting). In addition, the ECs derived from our experiments on the SciCite dataset are integrated into a web-based application designed to classify citation intents.

[2] We use *ACL-ARC* to refer to the citation dataset constructed by Jurgens et al. (2018) from the *ACL Anthology Reference Corpus (ARC)* (Kan, Min-Yen & Bird, Steven, 2009), which consists of multiple papers.

### 1.3 Contributions

While the individual components employed in this study – such as SciBERT, XLNet, and Ensemble Strategies – are well established within the NLP domain, the core methodological innovation of this work lies in the integration and adaptation of these blocks within a unified framework for CIC. Specifically, we propose a binarized, couple-based ensemble strategy designed to address class imbalance, through OVA decomposition, and to enrich the semantic understanding of citation contexts by integrating two complementary perspectives: one provided by a domain-specific language model, and the other by a general-purpose language model. Furthermore, section titles are employed as framing devices to improve classification performance, while XAI-based analyses provide interpretability at both the base and metaclassifier levels of our ensemble models. The interpretability layer provided by XAI analyses highlights the impact of section titles, and offers insights into the classification dynamics of CiteFusion. Overall, this approach is motivated by the need to capture both specialized scientific context and broader linguistic cues, and is empirically demonstrated to enhance the robustness, interpretability, and generalizability of the resulting classifiers.

In addition to this system-level contribution, our work presents the following outcomes: (i) public release of four ensemble classifiers based on complementary PLMs and FFNNs for CIC – two trained on ACL-ARC and two on SciCite; (ii) a state-of-the-art model for the SciCite benchmark; (iii) a state-of-the-art model for the ACL-ARC benchmark; (iv) empirically demonstrate the utility of section titles as framing devices within citation contexts, discussing how these elements reshape model's perception of input sentences; (v) provision of a mapping between the original annotation schemas of the datasets employed and standardized object properties from the Citation Typing Ontology (CiTO) to enhance interoperability; (vi) release a web-based application for automatic citation intent classification, leveraging the two ensemble classifiers trained on SciCite.

## 2 Background and Related Works

Within the field of Citation Intent Classification (CIC), significant advancements and developments have recently been produced, mainly thanks to an increasing spectrum of available methodologies and theoretical frameworks. This section will present the theoretical groundwork to proficiently understand our work. It begins with an overview of CIC schemas, followed by a review of the datasets and models used within this work. Next, the section will discuss the theoretical rationale for the use of section titles as semantic framing devices within citation contexts. Finally, it will provide a high-level summary of the strategies and the tools employed in this study.

### 2.1 A feasible Citation Intent Classification Schema

To accurately distinguish and categorize citation intents, it is important to consider the selection of an appropriate classification schema. Specifically, the choice of labels and the number of classes included in such a schema play a critical role in determining both the overall utility and impact of the resulting dataset and the outcomes achievable through automated classification methods. This discourse traces its roots back to the seminal work "*Can Citation Indexing Be Automated?*", by Garfield (1964), which identified diverse motivations behind citations and laid the groundwork for subsequent research focused on developing meaningful and accurate citation classification schemes (Kunnath et al., 2021). Well-designed schemas are essential for capturing the various functions of citations – such as *acknowledging the source of ideas*, *evidencing arguments*, *illustrating methodological similarities*, and *connecting related academic discussions* –, thereby serving applications ranging from research evaluation to academic writing and literature review (Pride, 2022).

Furthermore, developing a description logic schema compatible with semantic web technologies can unlock the potential to treat bibliographic references, citation contexts, and even rhetorical elements within scientific publications as semantic metadata, which enables a better organization, search, and integration for web-based scientific tools (Ciccarese et al., 2014). The *Citation Typing Ontology* (CiTO) (Peroni & Shotton, 2012) is a leading example, it provides a comprehensive vocabulary for annotating both factual and rhetorical purposes of citations. CiTO also facilitates the alignment of properties between different classification schemas, enabling interoperability and enriched representations of citation intents. This is particularly relevant for the schemas employed in this work. Indeed, both SciCite (Cohan et al., 2019) and ACL-ARC (Jurgens et al., 2018) datasets characterize their citation contexts into two newly designed schemas (SciCite employs three labels, while ACL-ARC utilizes six distinct categories). By mapping these two schemas to CiTO, it becomes possible to identify overlaps in their respective intents, thereby fostering a clearer and partially unified approach to citation intent classification in this work.

### 2.2 Datasets & Models for Citation Intent Classification

| Dataset | Categories (distribution) | Source | #instances |
|---|---|---|---|
| ACL-ARC Citation Dataset | Background (51%)<br>Extends (4%)<br>Uses (18%)<br>Motivation (5%)<br>Compare/Contrast (18%)<br>Future Work (4%) | Computational Linguistics | 1,941 |
| SciCite | Background (58%)<br>Method (29%)<br>Result Comparison (13%) | Computer Science & Medicine | 11,020 |

***Table 1***. *Comparison of SciCite and ACL-ARC citation datasets.*

A key consequence of designing an effective citation intent classification schema is the development of robust and comprehensive datasets. Most of the current datasets designed for classifying citation contexts use their own specific and newly developed schema. While creating multiple specialized schemas drives innovation, it complicates the integration of different resources (Cohan et al., 2019). Moreover, the field still suffers from a shortage of extensive and diverse datasets, mainly because manual annotation of citation contexts according to their intents is labor-intensive, and also requires domain-specific expertise (Pride, 2022). As a result, the limited number of manually annotated datasets currently available cannot yet serve as a definitive gold standard for the CIC task, though they offer important starting points.

An early contribution to CIC is the *ACL-ARC* citations dataset (Jurgens et al., 2018), which provides nearly 2,000 manually annotated citation contexts from the NLP domain. Despite employing a useful six-labels schema, ACL-ARC's narrow disciplinary focus and relatively small size limit its generalizability and overall utility. In contrast, the development of *SciCite*, by Cohan and colleagues (2019), marks a significant advancement in the field. This dataset spans both computer science and medicine related domains, featuring about 11,000 manually annotated citation contexts divided into three broader categories. The wider coverage, in both size and included domains, makes SciCite more versatile, and positions it as a valuable dataset for CIC.

The main problem of these datasets is represented by their imbalanced distribution (see ***Table 1***), which is a major issue when trying to solve the CIC task with automated methods (Kunnath et al., 2021). Previous works tried to overcome the skewness of datasets for CIC by applying SMOTE or by re-balancing the original corpus distribution (Jurgens et al., 2018; Nazir et al., 2020; Qayyum & Afzal, 2019). In contrast, our approach targets each class individually with dedicated and class-specific models, thereby facilitating a focused and independent processing of each intent category. This design choice, aimed at mitigating the challenges posed by data imbalance, enables our classifiers to effectively recognize and characterize even underrepresented classes.

### 2.2.1 Promising PLMs for CIC

The evolution of methodologies for addressing the CIC task mirrors advancements in machine learning (ML) and natural language processing (NLP). The introduction of transformer-based pretrained language models (PLMs) has significantly advanced the field. *SciBERT* (Beltagy et al., 2019), pretrained on scientific corpora and equipped with a specialized scientific vocabulary – *SciVOCAB* –, consistently outperforms general-purpose models in processing academic texts (Cohan et al., 2019). Its applications in CIC, including prompt-based frameworks such as *CitePrompt* (Lahiri et al., 2023), demonstrated robust performances, in particular on the ACL-ARC citation dataset on which the authors obtained state-of-the-art (SOTA) results. Moving on from SciBERT, which employed a masked language modelling objective, other training strategies and objectives resulted beneficial in understanding citation contexts despite the use of general training corpora. *XLNet* (Yang et al., 2019) is an example of such models, it combines autoregressive and autoencoding objectives, further improving the ability to understand citation contexts without specialized vocabularies. A fine-tuned version of XLNet, *ImpactCite* (Mercier et al., 2021), achieved SOTA results on SciCite.

Therefore, SciBERT's domain-specific pretraining makes it particularly effective for academic text, while XLNet's general-purpose pretraining allows it to adapt well to diverse language contexts. Given the distinct strengths of XLNet and SciBERT, we argue and demonstrate in this work how their integration can yield robust and reliable performance by capturing complementary aspects of citation contexts. By combining these architectures, we aim to highlight their respective capabilities and leverage their synergies to address diverse challenges in citation intent classification effectively.

The studies presented here, which employ these two PLMs, are mainly focused on classifying citation contexts, without expanding on the semantic role of their components. These approaches tend to prioritize performance metrics, and do not provide explanations of the underlying reasons for which an intent is preferred over another for a particular citation context. This emphasis on performance therefore results in no interpretability of the rationale behind models' decisions, underscoring the need for methodologies that balance accuracy with a deeper understanding of model behavior.

## 2.3 Ensemble Strategies and Explainable AI for Interpretable Classification

An interesting research direction in classification tasks is related to ensemble strategies (ES), which employ multiple baseline learners (base models) whose predictions are aggregated to produce a single output (Mohammed & Kora, 2023). ES can utilize either homogeneous or heterogeneous models and employ a variety of aggregation methods, including simple and weighted voting, but also more advanced meta-learning approaches. Prominent ensemble techniques such as *bagging*, *boosting*, and *stacking* – along with their variants – have been extensively explored in recent surveys (Khan et al., 2024; Mohammed & Kora, 2023).

In *Stacking*, multiple base models (sometimes referred to as *level-0* models) of the same or of different types are trained on the same or on different subsets of data. The predictions of these base models, either in the form of probabilities or class labels, are then combined and used to train a meta-model (sometimes referred to as *level-1* model), or aggregated through a more traditional voting strategy. *StackingC* (Seewald, 2002) is a notable variant of Stacking, in which *Multiple Linear Regression* (MLR)[3] is used to predict per-class specific probabilities from each set of level-0 models, reducing the complexity of the overall aggregation function and leading to better performances in multiclass settings. Another alternative to traditional stacking is the *Geometric Framework* described by Wu and colleagues (2023), in which MLR is used to minimize the Euclidean Distance (ED) from the predicted and the ideal points in $n$-dimensional spaces, thereby finding optimal weights at dataset level to apply to base models predictions as weighting schemes for weighted voting strategies.

In addition, when dealing with multi-class classification – such as the CIC task with the SciCite and ACL-ARC citation datasets –, some decomposition strategies may be adopted. Notably, *One-vs-One* (OVO) and *One-vs-All* (OVA) are two leading decomposition approaches in ensemble learning. OVO splits a problem into many binary subtasks, each of which represents a discrimination task between a pair of classes, resulting into $m(m-1)/2$ binary classifiers for a problem with $m$ classes (Galar et al., 2011; Gao et al., 2021). Instead, for the same $m$-classes problem, OVA creates a binary classifier for each of the $m$ classes, treating the target class of each binary classifier as the positive case and grouping all other classes together as the negative case, without differentiating among them (Galar et al., 2011).

Finally, ES demonstrated to improve classification performances in a wide range of tasks from various domains (Mohammed & Kora, 2023), and in many cases they perform better than more traditional methods when dealing with imbalanced classification tasks. Indeed, by harnessing the power of multiple classifiers, it is possible to deal more efficiently with underrepresented classes, as demonstrated in recent studies (Khan et al., 2024; Liu et al., 2022). Within ES it is also possible to harness the power of multiple baseline PLMs to produce different outputs to fuse through an aggregator (Huang et al., 2024; Jiang et al., 2023). To build such a framework there are various possibilities which may involve differentiated folds of data, heterogeneous or homogeneous baseline PLMs, and multiple learning stages (Monteiro et al., 2021).

Motivated by the efficacy of ES on imbalanced data, in this study we focus on Stacking and its variants, also driven by the objective of harnessing the complementary strengths of the PLMs introduced in the previous section when stacked as level-0 models in an ensemble framework. Furthermore, in designing our ES, we also apply OVA decomposition, as it requires fewer classifiers w.r.t. OVO, while ensuring that each base classifier directly addresses the separation between one intent and all the alternatives. Consequently, our framework includes two base PLM learners for each intent – one domain-specific and one general –, each trained to discriminate its target class against the rest through different linguistic perspectives. These pairs are then combined through stacking, allowing the ensemble to effectively integrate the unique contributions of both learners for each class.

#### 2.3.1 On the Opacity and Complexity of PLMs and ES: XAI for Citation Analysis

The complexity of ES and the opacity of PLMs – particularly in their internal dynamics and latent representations (e.g., transformer layers) – pose significant challenges for interpretability (Longo et al., 2024). *Explainable AI* (XAI) tries to address this by making classification processes more transparent, consequently fostering trust in systems developed to solve critical tasks such as CIC.

While XAI methods have been increasingly adopted in various academic domains, their application to scholarly citation tasks remains relatively limited. Weber and colleagues (2018) introduced an explanation-augmented citation recommender that leverages citation categories (such as "*substantiation*" or "*background*") to justify recommended citations in scientific literature. Their approach uses case-based reasoning to produce explanations that clarify the relationship between recommended citations and the original query document, thus supporting trust in automated citation recommendation systems. Similarly, Luo et al. (2023) proposed an interpretable architecture for legal citation prediction, allowing decisions to be traced to specific precedents and provisions in the legal text. Their work emphasizes that interpretability is essential in domains such as law, and highlights the lack of prior work on inherent interpretability for legal citation prediction.

[3] Also denoted as *Multi-Response Linear Regression* (Seewald, 2002).

These contributions demonstrate that XAI can play a significant role in increasing transparency and trust in citation-related prediction tasks, both in scientific and legal contexts. However, to date, the use of XAI to directly interpret and analyze citation intent classification models remains unexplored, as far as we know. Our work therefore seeks to address this gap by integrating post-hoc XAI methods into ensemble architectures for citation intent classification, thereby aiming to advance interpretability and reliability in this area.

XAI techniques like SHAP (*SHapley Additive exPlanations*) (Lundberg & Lee, 2017), which quantify feature contributions using Shapley values, offer valuable insights into model predictions and dataset characteristics. Shapley values help in identifying model- and class-specific relevant features – as token-level contributions for PLMs, and as "reliance" scores for the aggregator –, possibly helping in providing different viewpoints on the inherent semantic of various citation intents, while also contributing to explain CiteFusion's classifications. In this work, we apply SHAP to both levels of our ensemble: at level-0, SHAP highlights at token-level the features shaping the binary predictions of each base model; at level-1, SHAP values are instead employed to understand how the metaclassifier utilizes the output of level-0 models, showing typical patterns of trust and reliance in predicting different citation intents.

Since XAI methods help in demystifying "*black-box*" outputs (Longo et al., 2024) while enhancing reliability for downstream applications – as observed in other domains (Leichtmann et al., 2023; Trindade Neves et al., 2024) –, by coupling ensemble robustness with XAI transparency, we aim to achieve a balance between accuracy and interpretability, both essential to advance reproducible and trustworthy AI systems in academic settings.

## 2.4 Section Titles as Framing Devices

As briefly mentioned in the introductory part of this work, section titles have frequently been used as auxiliary or additional features within multitask learning frameworks for CIC (Cohan et al., 2019; Qi et al., 2023; Zhang et al., 2022), where their role was restricted to that of structural metadata. Although empirical studies consistently report improved classification performance when section titles are included in these frameworks, their utility has been interpreted in a purely functional sense – as additional input variables, or as targets of auxiliary tasks that enhance model accuracy. However, we argue that their function can also be interpreted through the lens of *framing theory*, which emphasizes how contextual cues shape interpretation by providing semantic scaffolding that guides readers' expectations and comprehension. According to this perspective, the meaning of a word, or phrase, *emerges* and varies according to the context surrounding it (Goffman, 1974).

This view is theoretically supported through the framework developed by Clarke (2009), which demonstrates that the meaning of linguistic expressions in NLP can be derived from any relevant set of contexts – including those defined at the sentence, section, or document level – thus motivating the explicit modeling of high-level structural cues, such as section titles, as meaningful contextual signals. This theoretical framework also aligns with recent works in NLP, where context is recognized as an essential element to address uncertainty. By reducing ambiguity and shaping the interpretation of sentences and texts for both humans and machines, the inclusion of contextual elements lead to improved decision making, error reduction, and enhanced model performance in NLP-related tasks (Garten et al., 2019; Khem et al., 2023). Matta and colleagues (2024) further reinforce this view by arguing that relying only on sentence-level analysis is insufficient for accurate interpretation in complex domains and that high-level structural cues may serve as situational context to guide both semantic extraction and intent understanding.

In addition, according to Bertin & Atanassova (2014), in the context of scientific writing, the role and function of a citation are closely linked to its position within the rhetorical structure of a research article. Their analyses revealed that specific citation act verbs are linked with specific sections within scientific articles, thereby concluding that the role that a citation may serve is defined also by its position within the rhetorical structure of a document. Thus, section titles may serve as immediate contextual cues that can help infer citation intent. This is especially relevant for PLMs trained on large corpora of scientific articles, as they are likely to internalize typical associations between section titles and specific language or discourse patterns (such as the act verbs presented by Bertin & Atanassova (2014)). Additionally, titles and headings not only bias comprehension toward particular topics (Lemarié et al., 2012), but also facilitate the organization and integration of information at the discourse level (Wiley & Rayner, 2000).

Beyond their conventional use as structural features, section titles may therefore be explicitly incorporated – either as prefixes or postfixes – into model inputs to encode contextual information within citation sentences. This approach leverages the semantic potential of section titles as framing devices, shaping both the model's attention and interpretation of citation contexts. Prior works further suggest that contextual cues are particularly effective when placed at the beginning of a sentence, therefore as prefixes (Pimentel et al., 2021). In this study, we implement this approach by prepending section titles as the initial tokens in the input sequence. By doing so, we enable section titles to work not only as organizational markers, but also as immediate contextual cues at the lexical and semantic level, making their influence directly accessible to the model throughout the classification process (Matta et al., 2024). Building on these insights, we argue that exploiting the role of section titles as framing devices can enhance model performance on citation intent classification tasks.

## 3 Models and Experiments

This section outlines the implementation and training dynamics of *CiteFusion*, the Ensemble Strategy (ES) designed to enhance the performances, as measured by Macro-F1[4] and accuracy scores, in the classification of citation intents on the *SciCite* and *ACL-ARC* citation datasets[5]. In developing *CiteFusion*, we decided to leverage the Pretrained Language Model (PLM) architectures outlined in *Section 2.2.1*: *SciBERT* (*Cased*) (Beltagy et al., 2019) and *XLNet* (*base-cased*) (Yang et al., 2019). The decision to employ both architectures was driven by the aim to harness the complementary strengths of domain specialization and generalized adaptability. Finally, to enhance the interoperability of the classifications performed by our models, we designed two mapping schemes that translate the original labels of the datasets employed into object properties selected from CiTO (Peroni & Shotton, 2012), summarized in ***Table 2***.

| Dataset | Original Labels | Mapping |
|---|---|---|
| ACL-ARC | Background | http://purl.org/spar/cito/obtainsBackgroundFrom |
| | Uses | http://purl.org/spar/cito/usesMethodIn |
| | CompareOrContrast | http://purl.org/spar/cito/discusses |
| | Extends | http://purl.org/spar/cito/extends |
| | Motivation | http://purl.org/spar/cito/obtainsSupportFrom |
| | Future | http://purl.org/spar/cito/citesAsPotentialSolution |
| SciCite | Method | http://purl.org/spar/cito/usesMethodIn |
| | Background | http://purl.org/spar/cito/obtainsBackgroundFrom |
| | Result | http://purl.org/spar/cito/usesConclusionsFrom |

***Table 2****. Mapping of the original schemes of SciCite and ACL-ARC citation datasets with object properties from CiTO.*

### 3.1 Problem Definition and Notation

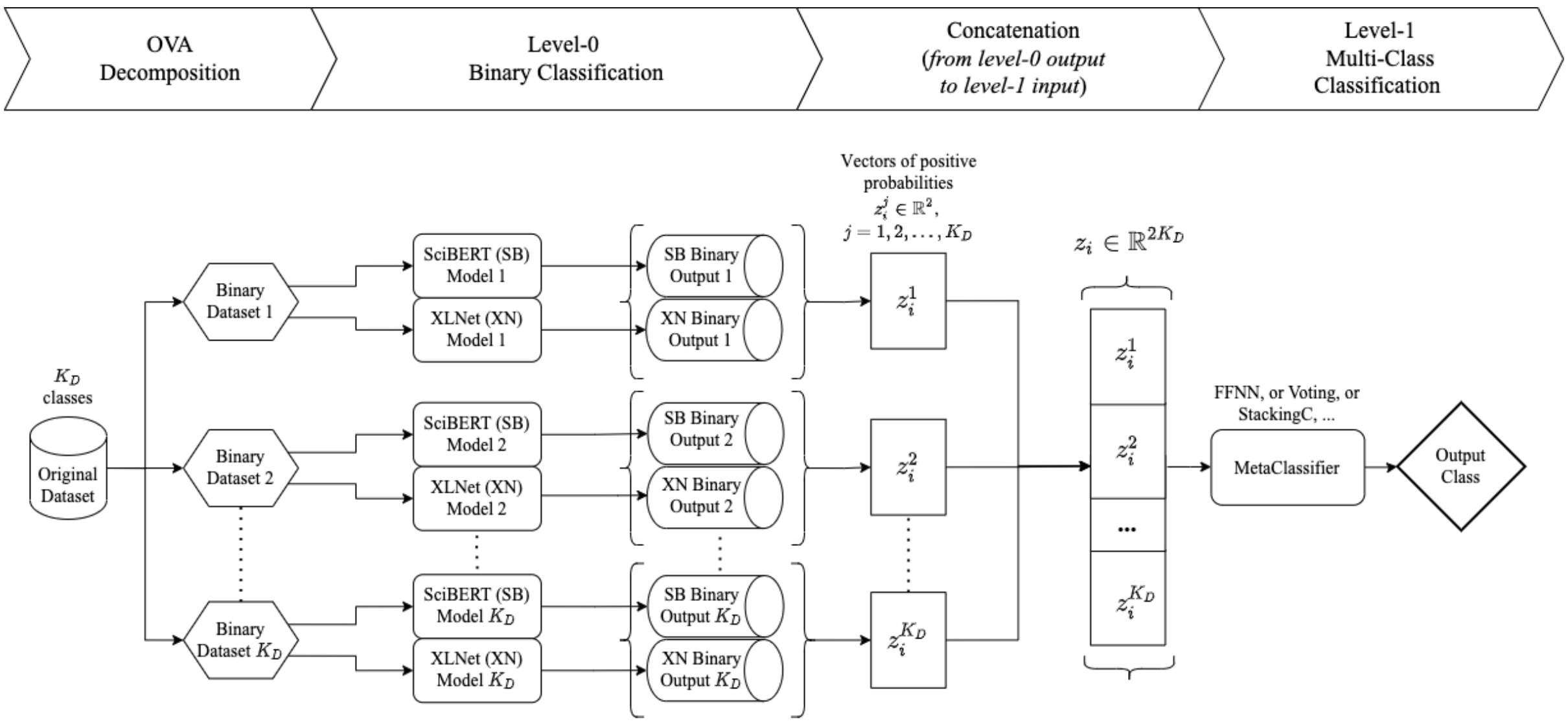

***Figure 2****. Overview of the CiteFusion Ensemble Architecture. The original dataset is decomposed into $K_D$ one-vs-all (OVA) binary classification tasks. For each binary task, two separate models (SciBERT and XLNet – identified as SB and XN, respectively) are trained. Their positive probabilities for each instance are concatenated into a feature vector, which is then passed to a level-1 meta-classifier to produce the final multi-class prediction.*

As outlined in *Section 1*, to address the CIC task, we designed *CiteFusion* (see ***Figure 2***), an ensemble strategy that leverages two different PLM architectures and fuses their respective strengths through a FFNN aggregator. This approach was applied to the two datasets presented. The design of CiteFusion was motivated by two key factors: (1) the relatively limited number of data points in each dataset and (2) the significant class imbalance present in the data, both of which restrict the ability to fully exploit the potential of a single PLM for this task. These challenges suggested the use of a stacked ensemble approach and a one-vs-all (OVA) decomposition strategy to effectively address and mitigate their impact. To provide a clear and concise description of the ensemble strategy, we briefly formalize its general structure.

[4] Macro-F1 is the reference metric used to compare the results obtained in CIC.
[5] While we utilized SciCite from the official Huggingface release (https://huggingface.co/datasets/allenai/scicite), we took the preprocessed ACL-ARC citation dataset from https://github.com/allenai/scicite (Cohan et al., 2019).

Let $\mathcal{D} = \{(s_i, x_i, y_i)\}_{i=1}^{N}$ be a dataset[6] of citation contexts $x_i$, section titles $s_i$, and labels $y_i \in \mathcal{L}_{\mathcal{D}}$, where $\mathcal{L}_{\mathcal{D}}$ is the set of all possible citation intent classes $c_j$ in $\mathcal{D}$, $K_{\mathcal{D}} = |\mathcal{L}_{\mathcal{D}}|$ is the number of classes, and $N$ is the number of samples $i$ contained in $\mathcal{D}$.

For each datapoint $i$, we define two input formats:

- $\beta_i^{\text{WS}} = s_i + ".\ " + x_i \rightarrow$ input sentence ***with*** section title prepended (*WS setting*)[7];
- $\beta_i^{\text{WoS}} = x_i \rightarrow$ input sentence ***without*** section title (*WoS setting*).

The use of two different input formats is instrumental in assessing the contribution of section titles as framing devices[8].

| Symbols | Meaning |
|---|---|
| $\mathcal{D}$ | Citation Intent Dataset |
| $N$ | Number of datapoints in $\mathcal{D}$ |
| $\mathcal{L}_{\mathcal{D}}$ | Set of class labels in $\mathcal{D}$ |
| $K_{\mathcal{D}}$ | Number of classes in $\mathcal{L}_{\mathcal{D}}$ and number of binary subtasks derived from the OVA decomposition of dataset $\mathcal{D}$ |
| $x_i$ | $i$-th citation context in $\mathcal{D}$ |
| $s_i$ | $i$-th section title in $\mathcal{D}$ |
| $y_i$ | $i$-th (multi-class) label in $\mathcal{D}$, contained in $\mathcal{L}_{\mathcal{D}}$ |
| $\beta_i^{\text{WS}}$, $\beta_i^{\text{WoS}}$ | $i$-th PLM's input formatted in WS or WoS settings |
| $c_j$ | $j$-th class label in $\mathcal{L}_{\mathcal{D}}$ |
| $T_{\mathcal{D}}$ | Set of OVA binary tasks for dataset $\mathcal{D}$ |
| $\tau_j$ | $j$-th binary task in $T_{\mathcal{D}}$ ($\tau_j$ is derived from the corresponding $c_j$) |
| $\mathcal{B}_j$ | $j$-th binary dataset derived from OVA decomposition, and structured to solve $\tau_j$ |
| $k_{i,j}$ | $i$-th binary label mapped from $c_j$ in OVA decomposition: can be either $k_{i,j}^{(0)}$ (negative) or $k_{i,j}^{(1)}$ (positive) |
| $A$ | Set of PLM architectures employed in this study |
| $a$ | PLM architecture, either SciBERT or XLNet: $a \in A$ |
| $f_{c_j,a}$ | Binary classifier trained to recognize $c_j$ and built as a tuned version of the PLM architecture $a$ |
| $\mathbf{P}_{i,c_j,a}$ | Softmax distribution of positive and negative probabilities for the attribution of the $i$-th sample to class $c_j$ from $f_{c_j,a}$ |
| $\rho_{i,c_j,a}^{(0)}$ | Probability that $\beta_i$ does not belong to class $c_j$ (negative attribution probability), in $\mathbf{P}_{i,c_j,a}$ |
| $\rho_{i,c_j,a}^{(1)}$ | Probability that $\beta_i$ belongs to class $c_j$ (positive attribution probability), in $\mathbf{P}_{i,c_j,a}$ |
| $M_{\mathcal{D}}$ | Set of $f_{c_j,a}$ used for dataset $\mathcal{D}$, with $\lvert M_{\mathcal{D}}\rvert = 2K_{\mathcal{D}}$ |
| $\mathbf{z}_i$ | $2K_{\mathcal{D}}$-dimensional vector of positive probabilities for the $i$-th datapoint |
| $z_i^j$ | $i$-th class-specific 2-dimensional sub-vector of $\rho_{i,c_j,a}^{(1)}$ produced by the PLMs in $a$ when tuned on $\mathcal{B}_j$ to solve $\tau_j$ |

***Table 3***. *Notation used in this paper.*

[6] Each dataset $\mathcal{D}$ is divided into three disjoint subsets: *training* ($\mathcal{D}_{(\text{train})}$ with $N_{(\text{train})}$ samples), *validation* ($\mathcal{D}_{(\text{val})}$ with $N_{(\text{val})}$ samples), and *test* ($\mathcal{D}_{(\text{test})}$ with $N_{(\text{test})}$ samples).

[7] The + sign here represents a concatenation of strings.

[8] Examples of the same instance $i$, from *SciCite*, when structured as input in both *WS* and *WoS* settings (we highlighted $s_i$):

- Example of $\beta_i^{\text{WS}}$: "**Introduction**. However, how frataxin interacts with the Fe-S cluster biosynthesis components remains unclear as direct one-to-one interactions with each component were reported (IscS [12,22], IscU/Isu1 [6,11,16] or ISD11/Isd11 [14,15])."
- Example of $\beta_i^{\text{WoS}}$: "However, how frataxin interacts with the Fe-S cluster biosynthesis components remains unclear as direct one-to-one interactions with each component were reported (IscS [12,22], IscU/Isu1 [6,11,16] or ISD11/Isd11 [14,15])."

### 3.1.1 OVA Decomposition

To address class imbalance, we employ a one-vs-all (OVA) decomposition for each dataset $\mathcal{D}$. Therefore, given the set of classes $\mathcal{L}_{\mathcal{D}} = \{c_1, c_2, \ldots, c_{K_{\mathcal{D}}}\}$, we construct $K_{\mathcal{D}}$ binary classification sub-tasks $T_{\mathcal{D}} = \{\tau_1, \tau_2, \ldots, \tau_{K_{\mathcal{D}}}\}$.

Therefore, for each class $c_j$ (with $j = 1, \ldots, K_{\mathcal{D}}$), a binary sub-task $\tau_j$ is defined by mapping the original multi-class labels $y_i \in \mathcal{L}_{\mathcal{D}}$, for each instance $i$, to binary labels $k_{i,j}$ using an indicator function[9]:

$$k_{i,j} = \mathbb{I}_{\{y_i = c_j\}} = \begin{cases} 1 \text{ if } y_i = c_j \to k_{i,j}^{(1)} \text{(positive class)} \\ 0 \text{ otherwise} \to k_{i,j}^{(0)} \text{(negative class)} \end{cases}$$

This generates a binary dataset $\mathcal{B}_j = \{(s_i, x_i, k_{i,j})\}_{i=1}^{N}$, where $\mathcal{B}_j$ isolates $c_j$ as the positive class and groups all other classes as negatives. The OVA decomposition therefore produces $K_{\mathcal{D}}$ binary datasets, each corresponding to a binary classification sub-task $\tau_j$. Each $\tau_j$ is then used to independently tune a couple of base PLMs on the corresponding $\mathcal{B}_j$. The resulting set of base (or *level-0*) binary classifiers serve as the first layer of CiteFusion, providing class-specific predictions that are later aggregated by higher-level strategies to obtain the final multi-class predictions.

## 3.2 Binary Classifications: Level-0 PLMs

Let $A = \{\text{SciBERT}, \text{XLNet}\}$ denote the set of PLMs employed.

For each binary task $\tau_j$, each architecture $a \in A$ is independently fine-tuned[10] on $\mathcal{B}_j$, resulting in a set $M_{\mathcal{D}}$ of *class-architecture-specific* (*base*, or *level-0*) classifiers $f_{c_j,a}$ for each instance $i$ of the original dataset $\mathcal{D}$:

$$M_{\mathcal{D}} = \left\{ f_{c_j,a} \;\middle|\; c_j \in \mathcal{L}_{\mathcal{D}}, a \in \mathrm{A} \right\}$$

yielding $|M_{\mathcal{D}}| = 2K_{\mathcal{D}}$ base binary classifiers. Each $f_{c_j,a}$ is trained only on $\mathcal{B}_j$ to distinguish $c_j$ (positive) from non-$c_j$ (negative), according to $\tau_j$. Thus, for each classifier $f_{c_j,a}$, the corresponding binary task $\tau_j$ is to distinguish whether an input[11] $\beta_i$ belongs to class $c_j$ ($k_{i,j}^{(1)}$) or not ($k_{i,j}^{(0)}$), as defined[12].

---

[9] Where $\mathbb{I}_{\{\cdot\}}$ denotes the indicator function, which returns 1 if the condition $\{\cdot\}$ holds, and 0 otherwise.

[10] Unless specified differently, we denote that all training and fine-tuning operations are conducted on the training split of their reference dataset. This convention also holds for the binary datasets derived from the OVA decomposition. In this case, tuning operations are therefore performed on $\mathcal{B}_{j(\text{train})}$.

[11] Either $\beta_i^{\text{WS}}$ or $\beta_i^{\text{WoS}}$, according to the setting in which $f_{c_j,a}$ was fine-tuned.

[12] Here we provide an **example**.

Consider a dataset $\mathcal{D}$ with classes $\mathcal{L}_{\mathcal{D}} = \{c_0, c_1, c_2\}$ ($K_{\mathcal{D}} = 3$). The OVA decomposition generates three binary datasets:

- $\mathcal{B}_0$: $\tau_0$ isolates $c_0$ (positive class $k_{i,0}^{(1)}$) vs. $\{c_1, c_2\}$ (negative class $k_{i,0}^{(0)}$),
- $\mathcal{B}_1$: $\tau_1$ isolates $c_1$ (positive class $k_{i,1}^{(1)}$) vs. $\{c_0, c_2\}$ (negative class $k_{i,1}^{(0)}$),
- $\mathcal{B}_2$: $\tau_2$ isolates $c_2$ (positive class $k_{i,2}^{(1)}$) vs. $\{c_0, c_1\}$ (negative class $k_{i,2}^{(0)}$).

For each $\mathcal{B}_j$, two PLMs are independently fine-tuned. Therefore:

- Task $\tau_1$ ($\mathcal{B}_1$):
    - $f_{c_1,\text{XLNet}}$ is fine-tuned to predict $k_{i,1}$, distinguishing $c_1$ from non-$c_1$;
    - $f_{c_1,\text{SciBERT}}$ is fine-tuned to predict $k_{i,1}$, distinguishing $c_1$ from non-$c_1$.

Similarly, classifiers $f_{c_0,\text{SciBERT}}$ and $f_{c_0,\text{XLNet}}$ are tuned on $\mathcal{B}_0$ for the task $\tau_0$, while $f_{c_2,\text{SciBERT}}$ and $f_{c_2,\text{XLNet}}$ are tuned on $\mathcal{B}_2$ to learn how to solve the task $\tau_2$.

This results in $|M_{\mathcal{D}}| = 6$ base binary classifiers:

$$M_{\mathcal{D}} = \{f_{c_0,\text{SciBERT}}, f_{c_0,\text{XLNet}}, f_{c_1,\text{SciBERT}}, f_{c_1,\text{XLNet}}, f_{c_2,\text{SciBERT}}, f_{c_2,\text{XLNet}}\}$$

Therefore, for an input sequence $\beta_i$ (either $\beta_i^{\text{WS}}$ or $\beta_i^{\text{WoS}}$), the output of a classifier $f_{c_j,a}$ is:

$$\mathbf{P}_{i,c_j,a} = \text{softmax}\left(f_{c_j,a}(\beta_i)\right) = \left(\rho_{i,c_j,a}^{(0)}, \rho_{i,c_j,a}^{(1)}\right)$$

where $\rho_{i,c_j,a}^{(1)}$ is the softmax probability[13] that the input $\beta_i$ belongs to class $c_j$, and $\rho_{i,c_j,a}^{(0)}$ is the softmax probability that it does not. Given the softmax, we have $\rho_{i,c_j,a}^{(1)} = 1 - \rho_{i,c_j,a}^{(0)}$.

#### 3.2.1 Concatenation: From Level-0 Outputs to Level-1 Inputs

For each datapoint $i$, we thereby process its representation $\beta_i$ with every $f_{c_j,a}$ in $M_{\mathcal{D}}$, and retain only the positive probabilities $\rho_{i,c_j,a}^{(1)}$ returned by this level-0 process. The entire set of $\rho_{i,c_j,a}^{(1)}$ for the $i$-th sample is then concatenated into a vector:

$$\boldsymbol{z}_i = \left[ \underbrace{\rho_{i,c_1,\text{SciBERT}}^{(1)}, \rho_{i,c_1,\text{XLNet}}^{(1)}}_{z_i^1} \,\Big\|\, \underbrace{\rho_{i,c_2,\text{SciBERT}}^{(1)}, \rho_{i,c_2,\text{XLNet}}^{(1)}}_{z_i^2} \,\Big\|\, \cdots \,\Big\|\, \underbrace{\rho_{i,c_{K_\mathcal{D}},\text{SciBERT}}^{(1)}, \rho_{i,c_{K_\mathcal{D}},\text{XLNet}}^{(1)}}_{z_i^{K_\mathcal{D}}} \right]$$

where $\boldsymbol{z}_i \in \mathbb{R}^{2K_\mathcal{D}}$ is the vector of concatenated positive class probabilities for datapoint $i$.

In addition, let $z_i^j = \left[ \rho_{i,c_j,\text{SciBERT}}^{(1)}, \rho_{i,c_j,\text{XLNet}}^{(1)} \right] \in \mathbb{R}^2$ be the class-specific sub-vector for sub-task $\tau_j$, containing the positive predictions $\rho_{i,c_j,a}^{(1)}$ from each $a \in A$ tuned on the dataset derived from $\tau_j$. Therefore:

$$\boldsymbol{z}_i = \left[\, z_i^1 \,||\, z_i^2 \,||\, \dots \,||\, z_i^{K_\mathcal{D}} \,\right]$$

Each vector $\boldsymbol{z}_i$, $\forall i \in \mathcal{D}$, is then passed to various aggregators to reconstruct the original multi-class setting.

### 3.3 MetaClassification: Multi-Class Level-1 Aggregators

The vector of positive class probabilities $\boldsymbol{z}_i \in \mathbb{R}^{2K_\mathcal{D}}$, for each datapoint $i$, serves as input to several meta-classification strategies, each aiming to reconstruct the original multi-class label from the outputs of the base (level-0) binary classifiers. We investigated both unsupervised and supervised aggregation approaches, starting with traditional voting mechanisms, and progressing to more advanced supervised strategies and meta-classifiers.

#### 3.3.1 Unsupervised Voting-Based Approaches

**Max Voting:** For each sample $i$, the predicted label $\hat{y}_i$ is determined by selecting the class $c_j$ corresponding to the highest positive probability in $\boldsymbol{z}_i$, therefore:

$$\hat{y}_i = c_{\hat{j}}, \quad \text{where } (\hat{j}, \hat{a}) = \underset{j \in \{1,\dots,K_\mathcal{D}\},\ a \in A}{\text{argmax}} \rho_{i,c_j,a}^{(1)}$$

**Average Voting:** For each instance $i$, the predictions from *SciBERT* and *XLNet* within each class-specific ($j$-th) sub-vector $z_i^j$ of $\boldsymbol{z}_i$, where $z_i^j \in \mathbb{R}^2$, are averaged to compute a *consensus score* $\bar{z}_i^j$ for class $c_j$. Formally:

$$\bar{z}_i^j = \frac{1}{2}\left(\rho_{i,c_j,\text{SciBERT}}^{(1)} + \rho_{i,c_j,\text{XLNet}}^{(1)}\right) = \frac{1}{2}\left(z_i^j[0] + z_i^j[1]\right), \ \ \forall j \in \{1, \dots, K_\mathcal{D}\}$$

[13] Each base PLM classifier is implemented with a two-logit (binary) softmax output layer, producing explicit probabilities for both the positive (class $c_j$) and negative (all other classes) outcomes. This approach is equivalent to using a sigmoid in terms of class prediction, but enables direct access to both positive and negative probabilities for each input.

where $z_i^j[0]$ and $z_i^j[1]$, contained in $z_i^j$, denote the positive probabilities produced by SciBERT and XLNet, respectively, when tuned on the binary sub-task $\tau_j$.

The scores obtained are then collected into a single consensus vector $\bar{\mathbf{z}}_i$ for the $i$-th instance of $\mathcal{D}$:

$$\bar{\mathbf{z}}_i = \left[\bar{z}_i^1, \bar{z}_i^2, \dots, \bar{z}_i^{K_\mathcal{D}}\right] \in \mathbb{R}^{K_\mathcal{D}}$$

and the final prediction $\hat{y}_i$ is retrieved by selecting the class $c_j$ corresponding to the higher consensus score in $\bar{\mathbf{z}}_i$, as follows:

$$\hat{y}_i = c_{\hat{j}}, \text{ where } \hat{j} = \underset{j \in \{1,\dots,K_\mathcal{D}\}}{\operatorname{argmax}} \bar{z}_i^j$$

**Majority Voting:** Each model $f_{c_j,a}$ outputs a positive vote for class $c_j$ if $\rho_{i,c_j,a}^{(1)} \geq \gamma$ (threshold $\gamma = 0.5$), with $\rho_{i,c_j,a}^{(1)} \in [0, 1]$:

$$v_{i,a}^{c_j} = \mathbb{I}_{\left\{\rho_{i,c_j,a}^{(1)} \geq \gamma\right\}}$$

where $v_{i,a}^{c_j}$ is the vote expressed for the $i$-th instance of $\mathcal{D}$ by the model $f_{c_j,a}$, and $\mathbb{I}_{\{\cdot\}}$ is the indicator function mapping the vote for class $c_j$ to 1 if the condition is met, and to 0 otherwise.

Let $V_{i,j}$ be the total number of votes for class $c_j$ for instance $i$:

$$V_i^{c_j} = \sum_{a \in A} v_{i,a}^{c_j} = v_{i,\text{SciBERT}}^{c_j} + v_{i,\text{XLNet}}^{c_j}$$

with $V_i^{c_j} \in \{0, 1, 2\}$, given $|A| = 2$.

Define then the set of classes with maximum votes as:

$$\mathcal{J}_i^* = \left\{j : V_i^{c_j} = \max_k V_{i,k}\right\}$$

then:

- if $|\mathcal{J}_i^*| = 1$, we assign $\hat{y}_i = c_{j^*}$, where $j^* \in \mathcal{J}^*$.
- If $|\mathcal{J}_i^*| > 1$ there is a tie, therefore *average voting* is applied between the set of classes in $\mathcal{J}^*$.

### 3.3.2 Supervised MetaClassification Strategies and Models

**Geometric Framework for Optimal Weights:** Building on traditional voting strategies, we adapted the supervised *geometric framework* introduced by Wu and colleagues (2023) (see *Section 2.3*) to our context. Specifically, for each class $c_j$, we estimate a class-specific weights vector via *Multiple Linear Regression* (MLR) on the binary $\mathcal{B}_{j(\text{val})}$, mapped from the validation split ($\mathcal{D}_{(\text{val})}$) of $\mathcal{D}$ according to the task $\tau_j \in T_\mathcal{D}$. The computed weight vectors are then applied to the corresponding class-specific positive prediction sub-vectors $z_i^j \in \mathbb{R}^2$ within each $\mathbf{z}_i$, such that for each class $c_j$, the predictions for $c_j$ are combined with the weight vector $\mathbf{w}_j$, yielding the weighted prediction vector $\dot{\mathbf{z}}_i = [\langle z_i^1, w_1\rangle \,||\, \langle z_i^2, w_2\rangle \,||\, \dots \,||\, \langle z_i^{K_\mathcal{D}}, w_{K_\mathcal{D}}\rangle]$ for each instance $i$.

Formally, for each datapoint $i$ of $\mathcal{D}_{(\text{val})}$, and for each class $c_j$, we computed $z_i^j = \left[\rho_{i,c_j,\text{SciBERT}}^{(1)}, \rho_{i,c_j,\text{XLNet}}^{(1)}\right] \in \mathbb{R}^2$, which denotes the vector of positive probabilities assigned by the two $\tau_j$-tuned level-0 models, as defined above. The corresponding true OVA binary label for class $c_j$ in $\tau_j$ is $k_{i,j} \in \{0, 1\}$.

For each class $c_j$, we therefore fit a weight vector $\boldsymbol{w}_j = \left[w_j^{\text{SciBERT}}, w_j^{\text{XLNet}}\right] \in \mathbb{R}^2$ by solving the following minimization problem[14] via MLR on the $\mathcal{D}_{(\text{val})}$:

$$\min_{\boldsymbol{w}_j} \sum_{i=0}^{N_{(\text{val})}} \left(\langle z_i^j, \boldsymbol{w}_j \rangle - k_{i,j}\right)^2$$

where $\langle \cdot,\cdot \rangle$ is the standard inner product in $\mathbb{R}^2$. Notably, following the original implementation of this geometric framework, we minimized the squared Euclidean Distance (ED) in $\mathbb{R}^2$ which, in the context of finding the optimal weights, is equivalent to the minimization of the ED since it cannot be negative (Wu et al., 2023).

After fitting, each $\boldsymbol{w}_j$ is normalized via a softmax normalization, to ensure interpretability and numerical stability. This normalization guarantees that the contributions of each model are directly comparable not only within a single weight vector $\boldsymbol{w}_j$ (for a given class), but also across different classes in the stacked weight matrix $W = \left[\boldsymbol{w}_1, \boldsymbol{w}_2, \dots, \boldsymbol{w}_{K_{\mathcal{D}}}\right]$.

Then, at test time, the final prediction score for class $c_j$ and instance $i$ is computed as

$$\dot{z}_i^j = \langle z_i^j, \boldsymbol{w}_j \rangle$$

where $z_i^j$ is the test-time positive probability vector for class $c_j$, and $\boldsymbol{w}_j$ is the corresponding class-specific weight vector estimated on $\mathcal{B}_{j(\text{val})}$ and normalized as described above.

The weighted predictions vectors $\dot{\boldsymbol{z}}_i = [\langle z_i^1, \boldsymbol{w}_1 \rangle \,||\, \langle z_i^2, \boldsymbol{w}_2 \rangle \,||\, \dots \,||\, \langle z_i^{K_{\mathcal{D}}}, \boldsymbol{w}_{K_{\mathcal{D}}} \rangle], \forall i \in \mathcal{D}_{(\text{test})}$, are then passed to the previously defined voting-based aggregation strategies to produce the final multi-class prediction for each instance.

**StackingC:** We investigated another extension over traditional stacking voting strategies by implementing the *StackingC* approach (Seewald, 2002), which leverages MLR to learn class-specific probabilities directly from the level-0 model predictions. As previously described, in adapting this method to our OVA framework, for each class $c_j$, we construct the binary validation set $\mathcal{B}_{j(\text{val})}$ corresponding to the binary task $\tau_j \in T_{\mathcal{D}}$, as derived from the validation split $\mathcal{D}_{(\text{val})}$ of the original dataset $\mathcal{D}$.

For every datapoint $i$ in $\mathcal{B}_{j(\text{val})}$, we extract the class-specific positive prediction vector $z_i^j$ and use the corresponding true binary label $k_{i,j}$ as the regression target. We then fit a class-specific MLR model for each $c_j$ by minimizing the sum of squared errors:

$$\min_{\theta_j, b_j} \sum_{i=0}^{N_{(\text{val})}} \left(\theta_j^\top z_i^j + b_j - k_{i,j}\right)^2$$

[14] No intercept term is included, in line with our code implementation (Paolini, 2024d).

where $\theta_j \in \mathbb{R}^2$ and $b_j \in \mathbb{R}$ are the regression coefficients and bias for class $c_j$ as computed from $\mathcal{B}_{j(\text{val})}$.[15]

At test time, for each instance $i$, with prediction vector $\boldsymbol{z}_i = [\, z_i^1 \,||\, z_i^2 \,||\, \dots \,||\, z_i^{K_\mathcal{D}} \,]$, we apply each trained $\text{MLR}_j$ model to the corresponding sub-vector $z_i^j$. For each class we thereby compute the class-specific logit score:

$$\text{logit}_{i,j} = \theta_j^\top z_i^j + b_j$$

Thus, for each $i$, we obtain a vector of $K_\mathcal{D}$ logit scores $\text{logit}_i = [\text{logit}_{i,1}, \text{logit}_{i,2}, \dots, \text{logit}_{i,K_\mathcal{D}}]$. We then apply a softmax normalization over this vector of logits to obtain a vector of class probabilities:

$$\mathbf{P}_i = [\rho_{i,1}, \rho_{i,2}, \dots, \rho_{i,K_\mathcal{D}}] = \text{softmax}(\text{logit}_i)$$

The final predicted label is then assigned via max voting, as previously defined, but this time on $K_\mathcal{D}$ votes.

**Meta Models – Direct Aggregators:** We also investigated a suite of Machine Learning (ML) models and algorithms, explicitly used as meta-classifiers, to aggregate the predictions of level-0 base models. For each datapoint $i$, the concatenated vector of positive class probabilities $\boldsymbol{z}_i \in \mathbb{R}^{2K_\mathcal{D}}$ serves as the input features vector for the meta-classifiers. We implemented Random Forest (RF), Support Vector Machine (SVM), Logistic Regression (LR), and K-Nearest Neighbors (KNN) through an extensive grid-search over possible combinations of hyperparameters, and defined a Feed Forward Neural Network (FFNN) for the aggregation task. For simplicity we aggregate the definition of these models and algorithms as one, and provide a brief formal description of the common working dynamics. Full details on hyperparameter optimization, model configurations, and implementation are available in the public computational notebooks (Paolini, 2024d).

Let $\mathcal{M}_{\text{meta}} : \mathbb{R}^{2K_\mathcal{D}} \to \mathbb{R}^{K_\mathcal{D}}$ denote a generic meta-classifier trained on the train split to map $\boldsymbol{z}_i$ to a vector of class logits in $\mathbb{R}^{K_\mathcal{D}}$:

$$\text{logit}_\text{i} = \mathcal{M}_{\text{meta}}(\boldsymbol{z}_i), \qquad \text{where } \boldsymbol{z}_i \in \mathbb{R}^{2K_\mathcal{D}} \wedge \text{logit}_\text{i} \in \mathbb{R}^{K_\mathcal{D}}$$

At test and inference time, a softmax is applied to obtain a probability distribution over $\text{logit}_\text{i}$:

$$\mathbf{P}_i = [\rho_{i,1}, \rho_{i,2}, \dots, \rho_{i,K_\mathcal{D}}] = \text{softmax}(\text{logit}_i)$$

and the final predicted class for instance $i$ is then selected as:

$$\hat{y}_i = \underset{j \in \{1,\dots,K_\mathcal{D}\}}{\text{argmax}} \, \mathbf{P}_i[j]$$

This approach allows the system to capture complex interactions between the level-0 binary predictions, ideally improving classification scores.

[15] Even though the settings and procedures of the geometric framework previously defined and of StackingC appear similar, it is important to highlight their difference in our context. Specifically, in the geometric framework, for each class $c_j$, MLR is used to learn a class-specific weight vector $\boldsymbol{w}_j$ by minimizing the squared Euclidean distance between the predicted probability vector $z_i^j$ and the OVA label $k_{i,j}$ over the validation split of the corresponding binary dataset $\mathcal{B}_{j(\text{val})}$. The resulting weights are then used to combine base model predictions at test time, and the final label is determined by applying the previously defined voting-based aggregation strategies. In contrast, StackingC employs MLR to fit a regression model (with coefficients $\theta_j$ and intercept $b_j$) for each class $c_j$, mapping the base prediction vectors $z_i^j$ in $\mathcal{B}_{j(\text{val})}$ directly to class probabilities. At test time, these meta-models output class logits for each $c_j$, which are then normalized via softmax across all classes, and the predicted label is assigned to the class with the highest probability. In summary, while StackingC operates as a meta-classification ensemble, the geometric framework is utilized to derive an optimal weighting scheme for the base model predictions.

### 3.4 Training Dynamics and Computational Instability

To employ the baseline models for the binary task we developed a fine-tuning loop structured to overcome the overfitting problem that usually arises when adapting LMs to downstream tasks. This loop allows for detailed assessments of the model's performance on the validation set, and it is accompanied by a scheduler to decrease learning rate on validation loss plateaus. Specifically, each model is evaluated every 10 batches of training data within each epoch. During each evaluation, the model's performance, in terms of validation loss, is compared to the best performance recorded up to that point and the best model's state in terms of validation loss is saved as a checkpoint. This process ensures that the model is continuously monitored, and the best performing version is preserved and finally retrieved at the end of the loop. Additionally, an early stopping mechanism has been implemented to stop the fine-tuning process after 50 evaluations without performance increase.

Level-0 models were fine-tuned using constant learning rate ($2e^{-5}$), weight decay ($0.01$), and batch size ($32$) values across experiments[16]. The *AdamW* optimizer with *cross-entropy loss* was employed, with weight decay applied to all parameters except for *biases* and *LayerNorm* weights. Excluding these terms is a standard practice, as regularizing bias and normalization parameters has been shown to hinder model performance and training stability (Ba et al., 2016; Lahiri et al., 2023; Loshchilov & Hutter, 2017).

Mixed-precision training was also employed, utilizing both 16-bit and 32-bit floating-point numbers to reduce memory usage and accelerate computations without compromising model performance. To maintain coherence with the ensemble strategy, level-0 models were not evaluated as individual classifiers on the test sets of the binary datasets used for fine-tuning.

**On the Reproducibility/Replicability of CiteFusion:** Ensuring the replicability of our results, and a robust and transparent description that enables the reproducibility of our methodology are a key focus of this work. While we follow best practices for both reproducibility and replicability – using fixed random seeds, deterministic algorithmic settings, and controlled hardware/software environments[17] – recent works and official documentation emphasize that some degree of run-to-run variability is unavoidable in deep learning frameworks (Desai et al., 2025; Gundersen et al., 2018), in particular when dealing with Language Models (Semmelrock et al., 2025). Also, the PyTorch documentation[18] notes that certain sources of non-determinism, such as hardware (e.g. GPU vs CPU executions) or library (e.g. PyTorch vs TensorFlow) differences, can introduce minor variations between runs even when identical seeds and code are used (Semmelrock et al., 2025). These factors are inherent to modern computational pipelines in deep learning (Desai et al., 2025; Goodman et al., 2016; Semmelrock et al., 2025), and in our case, the use of imbalanced datasets and mixed-precision training, mainly when implemented with fine-grained evaluations, may also contribute to additional small numerical differences across repeated experiments.

To quantify the robustness and computational instability of CiteFusion, we systematically repeated all training and evaluation procedures for every experimental condition over 10 additional and independent runs, each with fixed seeds, consistent data splits, and within the same environment[19]. This protocol was uniformly applied to all base PLMs fine-tuning, OVA decompositions, and to the FFNN ensemble meta-classifier training and evaluation. By reporting the distribution of results (mean, variance, and full scores) for all major metrics, we provide a direct empirical measure of the variability induced by our strategy in imbalanced settings (see *Sections 4.1* and *4.2.1*). This approach ensures procedural transparency and allows readers to accurately assess the practical robustness of our framework.

[16] To inspect the full hyperparameters setting refer to the publicly available code (Paolini, 2024d).
[17] All libraries and packages utilized in this project are publicly available, and their specific versions are detailed in (Paolini, 2024d).
[18] https://docs.pytorch.org/docs/stable/notes/randomness
[19] All experiments were conducted in the Google Colab environment, in line with reported best-practices (Semmelrock et al., 2025), with the same versions of the same libraries.

Furthermore, in accordance with contemporary standards for scientific rigor, our methodology is intentionally designed to address three major axes of empirical reliability (Desai et al., 2025; Goodman et al., 2016):

1. **Repeatability:** We systematically execute all experiments ten times under identical settings, with the goal of verifying internal consistency and quantifying the extent of variability attributable to random fluctuations.
2. **Dependent reproducibility:** We provide comprehensive descriptions of code, hyperparameters, and processed data sources and splits, thereby enabling independent researchers to fully reproduce our results using the same data and implementation.
3. **Conceptual replicability:** To evaluate the adaptability and generalizability of our approach, we apply CiteFusion not only to our primary citation intent dataset, but also to the independent ACL-ARC benchmark.

Finally, we did not evaluate reproducibility or replicability of our framework on alternative platforms, hardware configurations, or with different library versions. Expanding our analyses in this direction would have required many additional resources and time, and is outside the intended scope of this study, which is focused on developing and evaluating a framework for CIC, and in assessing the contribution of section titles as framing devices (as detailed in *Section 2.4*). Nonetheless, we acknowledge the value of future work investigating cross-platform and cross-library robustness, and we encourage the community to follow these directions in dedicated studies.

### 3.5 Explainers

Following the development of CiteFusion to address the CIC task, we also conducted a series of experiments using SHAP (Lundberg & Lee, 2017) to enhance the interpretability and trustworthiness of our classifiers. To ensure consistency, SHAP was applied systematically across all the ECs, and SHAP values were computed at both levels (level-0 and level-1) of the ensemble framework for both datasets, SciCite and ACL-ARC. At level-0, SHAP was employed to explain the contributions of individual tokens, considered as features influencing the predictions made by the base models. The analysis of the level-0 results of these experiments provides preliminary insights into how specific words or phrases impact classification outcomes in binary settings, also highlighting the differences between the two PLM architectures employed.

At level-1, SHAP was instead employed to clarify the aggregation process of base model predictions within the FFNN metaclassifier. This approach facilitates an understanding of which base models contribute most significantly to shaping the final predictions for each class. By applying SHAP at both levels, we gain a comprehensive view of the classification dynamics of our ES while also uncovering dataset-specific characteristics inherent to both SciCite and ACL-ARC. In the context of this research work, SHAP serves as a valuable tool for enhancing transparency and fostering a deeper understanding of models' behavior and underlying data properties.

## 4 Experimental Results and Discussion

This section presents and discusses the results obtained when classifying citation intents from both SciCite and ACL-ARC citation datasets. Performance, robustness, and reliability of our framework are investigated, together with a discussion about the underlying classification dynamics of our ensemble. In addition, the effect of section titles when used as framing devices is assessed through different viewpoints – performance and overall impact on CiteFusion's classification dynamics –, with a focus on how their integration reshapes the way in which the ensemble perceives citation contexts. This evaluation highlights key differences between the different kind of features considered by the two Pretrained Language Model (PLM) architectures when performing binary classifications. The effect of this perceptual shift is also examined for the FFNNs metaclassifiers, discussing the contributions of each model for each class-specific output of both datasets in both settings. These analyses help in understanding the classification dynamics of our Ensemble Classifiers (ECs), thereby enhancing the overall interpretability of CiteFusion. Finally, error analyses conducted through SHAP and more traditional methods are presented, to highlight strengths and weaknesses of our framework.

The results of computational instability analyses are also presented in this section. These are intended to demonstrate the robustness of our Ensemble Strategy (ES) and its consistency under imbalanced data settings. The same analyses are utilized to assess the effect of mixed precision and fine-grained evaluations on reproducibility and replicability.

### 4.1 Results

Experimental results revealed a positive influence of incorporating section titles. Across all experiments conducted on both datasets, the WS (with section titles) setting consistently outperformed the WoS (without section titles) setting. For SciCite, the mean improvements across all the aggregation functions and models utilized in the experiments were $\Delta_A = 0.73$ and $\Delta_{MF1} = 0.93$. Similarly, the mean improvements w.r.t. the final classification scores for ACL-ARC were $\Delta_A = 1.86$ and $\Delta_{MF1} = 1.55$. Furthermore, as evident from ***Table 4***, FFNNs consistently outperformed all other aggregators, except for the WoS setting in ACL-ARC, in which Random Forest (RF) performed better.

| Setting | Aggregation Function | SciCite | | ACL-ARC | |
|---|---|---|---|---|---|
| | | Accuracy | Macro-F1 | Accuracy | Macro-F1 |
| WS | Max | 89.94 | 88.51 | 79.86 | 71.90 |
| | Avg | 90.16 | 88.98 | 78.42 | 68.58 |
| | Maj | 90.10 | 88.92 | 78.42 | 68.58 |
| | W-Max | 89.99 | 88.56 | 79.14 | 71.33 |
| | W-Avg | 90.21 | 88.99 | 78.42 | 68.35 |
| | W-Maj | 90.05 | 88.74 | 78.42 | 70.28 |
| | StackingC | 90.10 | 88.81 | 82.73 | 73.91 |
| | RF | 88.76 | 87.36 | 80.58 | 73.02 |
| | SVM | 89.89 | 88.75 | 78.42 | 68.16 |
| | LR | 90.16 | 89.08 | 79.14 | 70.86 |
| | KNN | 88.70 | 87.25 | 79.86 | 69.20 |
| | FFNN | 90.08 | 89.60 | 81.29 | 76.24 |
| | Mean | 89.85 | 88.63 | 79.62 | 70.81 |
| WoS | Max | 88.81 | 87.30 | 76.98 | 69.68 |
| | Avg | 89.46 | 88.04 | 77.70 | 69.42 |
| | Maj | 89.46 | 88.04 | 78.42 | 70.10 |
| | W-Max | 88.92 | 87.67 | 77.70 | 69.78 |
| | W-Avg | 89.46 | 88.11 | 78.42 | 67.74 |
| | W-Maj | 89.46 | 88.11 | 79.14 | 68.38 |
| | StackingC | 89.24 | 87.91 | 76.98 | 62.15 |
| | RF | 88.54 | 86.97 | 79.86 | 72.44 |
| | SVM | 89.24 | 87.75 | 74.82 | 69.76 |
| | LR | 89.13 | 87.72 | 79.14 | 71.03 |
| | KNN | 88.17 | 86.56 | 76.26 | 68.42 |
| | FFNN | 89.56 | 88.22 | 76.98 | 71.46 |
| | Mean | 89.12 | 87.70 | 77.76 | 69.26 |

***Table 4***. *The table presents Accuracy and Macro-F1 scores for different aggregation strategies of level-0 models on both SciCite and ACL-ARC datasets. It includes mean results across all level-1 models (heads) for both WS and WoS settings. Individual scores are reported for three voting strategies – Max, Average (Avg), and Majority (Maj) – along with their weighted versions (W-Max, W-Avg, W-Maj). Scores for machine learning (ML) algorithms are also provided. For FFNN, the reported scores reflect the best results from 10 runs of CiteFusion in the WS setting, and single-run results in the WoS setting.*

The best-performing model for the SciCite dataset was trained in WS setting. Specifically, it attains a Macro-F1 score of 89.60%, a Micro-F1 and accuracy scores of 90.80%, and a weighted F1 score of 90.87%. Additionally, it demonstrates a precision (macro average) of 88.57% and a recall (macro average) of 90.95%. The class-specific performance metrics are as follows: for the *Method* class, accuracy and F1 scores are 89.60% and 91.65%, respectively; for the *Background* class, 91.06% and 91.85%; and for the *Result* class, 91.89% and 85.30%.

For the ACL-ARC dataset, the best-performing model is again trained in WS setting. This achieves a Macro-F1 score of 76.24%, a Micro-F1 and accuracy scores of 81.29%, and a weighted F1 score of 80.77%. The precision (macro average) is 80.91%, and the recall (macro average) is 72.84%. Class-specific results are as follows: for the *Background* class, accuracy and F1 scores are 92.96% and 86.84%, respectively; for the *Uses* class, 69.23% and 73.47%; for the *CompareOrContrast* class, 72.00% and 78.26%; for the *Extends* class, 80.00% and 88.89%; for the *Motivation* class, 42.86% and 50.00%; and for the *Future* class, 80.00% and 80.00%.

| Setting | Run | SciCite | | ACL-ARC | |
|---|---|---|---|---|---|
| | | Accuracy | Macro-F1 | Accuracy | Macro-F1 |
| WS | 0 | 90.32 | 89.01 | 80.58 | 75.28 |
| | 1 | 90.53 | 89.39 | 80.58 | 73.19 |
| | 2 | 90.64 | 89.50 | 79.14 | 72.12 |
| | 3 | 90.53 | 89.35 | 79.86 | 73.35 |
| | 4 | 90.59 | 89.42 | 79.86 | 70.76 |
| | 5 | 90.32 | 89.01 | 79.86 | 72.81 |
| | 6 | 90.80 | 89.60 | 79.86 | 72.36 |
| | 7 | 90.64 | 89.50 | 81.29 | 76.24 |
| | 8 | 90.59 | 89.38 | 80.58 | 73.91 |
| | 9 | 90.59 | 89.43 | 81.29 | 75.22 |
| | 10 | 90.37 | 89.22 | 82.01 | 75.54 |
| | Mean (Std) | 90.53 (0.14) | 89.34 (0.18) | 80.44 (0.79) | 73.71 (1.61) |

***Table 5***. *Results of computational instability analyses on both datasets. Accuracy and Macro-F1 scores of each run are reported, together with score and dataset specific means and standard deviations (grey row). Run 10 represents the first (base) experiment, while runs from 0 to 9 represent repeated runs of it.*

As previously mentioned, we conducted additional analyses to assess the computational instability of our Ensemble Strategy. These analyses were performed exclusively under the WS setting[20] by repeating the same experiment 10 additional times for both ACL-ARC and SciCite. For both the analyses, we employed the same FFNN architecture as aggregator for the level-0 models. The results of these investigations are summarized in ***Table 5***.

Finally, for plots detailing the results involving SHAP to explain the predictions of our ECs, we refer the reader to ***Figures A.1, A.2, A.3, A.4, A.5, and A.6*** in ***Appendix A.1*** and to the following sections. SHAP analyses reveal the most influential model- and class-specific tokens for level-0 binary predictions, while also identifying the most impactful models for class-specific level-1 metaclassifications via FFNNs. The paper presents the most interesting findings related to SHAP while discussing them, in *Section 4.3*.

### 4.2 Discussion

CiteFusion (WS) models surpass the previous state-of-the-art (SOTA) results in both SciCite and ACL-ARC benchmarks, as depicted in ***Tables 5 and 6***. Furthermore, the FFNN metaclassifier head consistently outperforms the previous SOTA models for both SciCite and ACL-ARC across all the computational instability runs.

In addition, the inclusion of section titles in input sentences consistently improves the performance of our ECs in CIC, as highlighted by the gains obtained in both Accuracy and Macro-F1 scores across the datasets w.r.t. WoS setting. Nevertheless, even in WoS setting the robust performance of the ECs highlights the effectiveness of CiteFusion in dealing with underrepresented classes in both datasets. By training complementary couples of level-0 models on OVA binary tasks, we enabled a direct and independent processing of each class, effectively mitigating class imbalance.

[20] Due to limited computational resources.

Results are summarized in ***Table 6*** for SciCite, and in ***Table 7*** for the ACL-ARC dataset. These two tables compare the scores of our models (*CiteFusion*) against the two benchmarks[21]. In ***Table 6*** the comparison is against the two SOTA models for SciCite: *ImpactCite* (Mercier et al., 2021) and *CitePrompt* (Lahiri et al., 2023).

| Model | Per-Class Accuracy | | | Per-Class F1 Scores | | | Accuracy | Precision | Recall | Micro-F1 | WeightedF1 | Macro-F1 |
|---|---|---|---|---|---|---|---|---|---|---|---|---|
| | MET | BKG | RES | MET | BKG | RES | | | | | | |
| CiteFusion (WS) | 94.67 | 91.34 | 95.59 | 91.65 | 91.85 | 85.30 | 90.80 | 88.57 | 90.95 | 90.80 | 90.87 | 89.60 |
| CiteFusion (WoS) | 93.65 | 89.89 | 95.59 | 89.58 | 91.00 | 84.86 | 89.73 | 89.77 | 87.40 | 89.73 | 89.68 | 88.48 |
| ImpactCite[22] | 85.79 | 88.34 | 92.67 | 87.00 | 90.00 | 85.00 | 88 | NA | NA | 88.13 | 88 | 88.93 |
| CitePrompt | NA | NA | NA | NA | NA | NA | 87.56 | NA | NA | NA | NA | 86.33 |

***Table 6****. Results for the SciCite dataset: This table compares our models, CiteFusion (in both WS and WoS settings), with the current state-of-the-art models for the CIC task on this dataset.*

In ***Table 7*** the comparison is against *CitePrompt* (the current best-performing model for this task), and the model obtained by Cohan and colleagues (2019), which is presented as *Structural Scaffolds* in the table.

| Model | Background | | Uses | | ComOrCon | | Extends | | Motivation | | Future | | A | P | R | MiF1 | W-F1 | MaF1 |
|---|---|---|---|---|---|---|---|---|---|---|---|---|---|---|---|---|---|---|
| | A | F1 | A | F1 | A | F1 | A | F1 | A | F1 | A | F1 | | | | | | |
| CiteFusion (WS) | 85.61 | 86.84 | 90.65 | 73.47 | 92.81 | 78.26 | 99.28 | 88.89 | 95.68 | 50.00 | 98.56 | 80.00 | 81.29 | 80.91 | 72.84 | 81.29 | 80.77 | 76.24 |
| CiteFusion (WoS) | 81.29 | 83.33 | 89.93 | 72.00 | 89.21 | 63.41 | 97.84 | 66.67 | 97.12 | 60.00 | 98.56 | 83.33 | 76.98 | 79.86 | 69.27 | 76.98 | 75.86 | 71.46 |
| Structural Scaff. | NA | 83.5 | NA | 75.0 | NA | 71.1 | NA | 66.7 | NA | 44.4 | NA | 66.7 | NA | 81.3 | 62.5 | NA | NA | 67.9 |
| CitePrompt | NA | NA | NA | NA | NA | NA | NA | NA | NA | NA | NA | NA | 78.42 | NA | NA | NA | NA | 68.39 |

***Table 7****. Results for the ACL-ARC Dataset: This table compares our proposed models, CiteFusion (in both WS and WoS settings), with the current state-of-the-art models for the CIC task on this dataset (note that Structural Scaffolds name is abbreviated). For clarity, we use abbreviations to represent evaluation metrics. Evaluation scores are denoted as follows:* ***A*** *for Accuracy,* ***P*** *for Precision,* ***R*** *for Recall,* ***MiF1*** *for Micro-F1,* ***W-F1*** *for Weighted-F1, and* ***MaF1*** *for Macro-F1.*

This table, together with ***Table 6***, demonstrate the superior performance of our strategy compared to current state-of-the-art models, underscoring its effectiveness in handling imbalanced datasets, and its ability in targeting underrepresented classes. This is especially evident when examining the significant improvements achieved by CiteFusion (WS) on the ACL-ARC dataset, with nearly an overall 8% increase in Macro-F1 and a 3% rise in accuracy compared to the previous SOTA, and also when looking at the higher class-specific Macro-F1 scores obtained for the underrepresented *Extends*, *Motivation*, and *Future* intents, when compared to previous SOTA.

| Model | Method | | Background | | Result | |
|---|---|---|---|---|---|---|
| | P | R | P | R | P | R |
| CiteFusion (WS) | 93.46 | 89.90 | 92.65 | 91.06 | 79.60 | 91.86 |
| CiteFusion (WoS) | 92.04 | 88.08 | 89.22 | 92.27 | 85.26 | 82.63 |

***Table 8****. Class-specific Precision (P) and Recall (R) scores for the SciCite Dataset.*

We report in ***Tables 8*** and ***9*** class-specific *precision* and *recall* scores for SciCite and ACL-ARC, respectively, in both WS and WoS settings. The integration of section titles in SciCite lead to performance differences across classes. For the underrepresented *Result* class, CiteFusion in WS setting achieves a recall of 91.86% (+9.23% over WoS), while the precision drops to 79.60% (from 85.26% in WoS). This demonstrates that the inclusion of section titles helps in identifying more instances from the *Result* class, while also increasing the number of false positives for this same class. This is in line with our expectations since section headers, such as "*Results*", likely provide strong signals for base models and while these enhance the detection of the *Result* class, may also introduces biases seen the underrepresentation of this intent in SciCite.

For the majority classes instead – i.e., *Method* and *Background* – CiteFusion in WS setting demonstrates improved precision, achieving 93.46% compared to 92.04% in WoS for *Method*, and 92.65% compared to 89.22% in WoS for *Background*, while increasing the recall from 88.08% in WoS to 89.90% for the *Method* class, and decreasing it from 92.27% in WoS to 91.06% in WS for the *Background* class. These results indicate that section titles, when employed as contextual elements within input sentences, help models in better identifying and weighing relevant features for each class. The improved detection of class-specific features, in the ensemble context, leads to better performances across the entire dataset, without compromising the coverage for single classes.

[21] The tables, for *CiteFusion*, report the best scores obtained through 10 runs of the ECs in WS settings, and the scores obtained through a single run in WoS settings. The other models reported in these tables are the current state-of-the-art models for the two datasets in CIC.

[22] Even if reference scores of ImpactCite (Mercier et al., 2021) for Accuracy, Weighted-F1, and class-specific F1s are not directly reported in the paper, they are present in the code released by the authors, publicly available at https://github.com/DominiqueMercier/ImpactCite/blob/main/ImpactCite_Intent/Tester.ipynb.

In ACL-ARC, the inclusion of section titles has a significant but uneven impact on underrepresented classes. For the *Extends* class, CiteFusion in WS setting achieves perfect precision (100% compared to 75% for WoS) and higher recall (80% compared to 60%), suggesting that section titles help resolve ambiguities in identifying this rare class. Differently, for the *Future* class, in WS setting CiteFusion has increased precision (80% vs. 71.43%) but decreased recall (80% vs. 100%) w.r.t. WoS setting. This suggests that when section titles are included, the EC becomes more selective in assigning the *Future* label – i.e., makes fewer predictions for this class, but those predictions are more likely to be correct (higher precision). Here, the recall decline indicates that some *Future* instances are now missed.

In contrast, the *Motivation* class sees a decline in precision (60% vs. 100% in WoS), despite maintaining identical recall (42.86%), suggesting that section titles may introduce noise for this class. This noise likely derives from the indirect interaction between section titles and the metaclassifier's base-models weighing mechanism. Indeed, *Motivation*-type citations in ACL-ARC appear in sections with ambiguous titles, such as "*Introduction*", which lack a clear semantic association with motivational intent and are instead more closely related to other classes, such as *Background*. This missing association may be particularly relevant when considering the severe imbalance of ACL-ARC, which do not provide the model with enough datapoints to infer it. Additionally, due to the scarcity of *Motivation*-type citations, the metaclassifier learns to give higher weights to positive probabilities from *Motivation*-based level-0 models in classifying *Motivation* instances (see *Section 4.3*, and ***Figure 8***). Therefore, in the WS setting, the presence of noisier signals derived from the ambiguity of the relation between section titles and class-specific cues, combined with the metaclassifier's tendency to overweight predictions from *Motivation*-specific models, results in the misclassification of these mixed signals as *Motivation*. This, in turn, leads to an increase in false positives.

| Model | Background | | Uses | | ComOrCon | | Extends | | Motivation | | Future | |
|---|---|---|---|---|---|---|---|---|---|---|---|---|
| | P | R | P | R | P | R | P | R | P | R | P | R |
| CiteFusion (WS) | 81.48 | 92.96 | 78.26 | 69.23 | 85.71 | 72.00 | 100.0 | 80.00 | 60.00 | 42.86 | 80.00 | 80.00 |
| CiteFusion (WoS) | 76.47 | 91.55 | 75.00 | 69.23 | 81.25 | 52.00 | 75.00 | 60.00 | 100.0 | 42.86 | 71.43 | 100.0 |

***Table 9****. Class-specific Precision (P) and Recall (R) scores for the ACL-ARC Dataset.*

Among majority classes in ACL-ARC, CiteFusion in WS setting has consistently higher performances w.r.t. WoS setting. For *Background*, precision increases from 76.47% to 81.48%, and recall improves slightly from 91.55% to 92.96%. Similarly, for *CompareOrContrast*, precision rises from 81.25% to 85.71%, and recall increases from 52% to 72%. Finally, for *Uses*, precision increases from 75.00% to 78.26%, and recall remains the same. These improvements likely derive from the stronger contextual alignment provided by section titles. These findings on class-specific precision and recall scores highlight that section titles disproportionately benefit rare classes associated with specific sections, such as *Extends*. However, their utility varies depending on the semantics of the class and the relevance of the titles. Overall, including section titles provide a benefit for most classes in ACL-ARC, as demonstrated by the increased overall performances in Macro-F1 and accuracy scores (***Table 7***).

**Supervised Approaches – Comparison Between Heads**

As can be seen in ***Table 4***, the FFNN meta-classifier consistently outperforms all other aggregation strategies in both accuracy and Macro-F1 scores across both citation datasets (SciCite and ACL-ARC) and settings (WS and WoS). The only exception is observed in the WoS setting for ACL-ARC, where *Random Forests* (RF) achieve marginally higher performance. A closer inspection of class-specific results demonstrates that RF excels in majority classes (e.g., *Background* and *CompareOrContrast*) but struggles in classifying underrepresented intents. This behavior is attributable to the inherent construction of RFs which, as models designed to minimize the overall error rate, prioritize identifying reliable predictions for majority classes at the expense of accuracy for less represented intents (Chen et al., 2004). Consequently, even if RF obtains the best classification performances in the WoS setting for the ACL-ARC dataset, its limited focus on underrepresented classes makes it less suitable as the final aggregator for CiteFusion. Considering this, we decided to employ the FFNN as the preferred meta-classifier, given its ability to balance performance across both majority and minority classes effectively, and given the will to aggregate in a synergistic and complementary manner the independent level-0 predictions.

*StackingC* (Seewald, 2002), as described in *Section 3*, demonstrates robust performance in both datasets and settings, effectively leveraging class-specific binary predictions to mitigate data scarcity challenges. However, its lower performance relative to FFNN can be attributed to its inability to account for interdependencies between base model predictions across different classes. Unlike the FFNN, which synthesizes cross-class interactions through a unified architecture, StackingC operates independently on each class, potentially limiting its capacity to resolve ambiguities arising from the overlapping of level-0 predictions of citation intents. This limitation is particularly evident in complex datasets like ACL-ARC, where the semantic boundaries between classes are less distinct.

The geometric framework employing *Euclidean Distance* to determine optimal weights for voting strategies, adapted from Wu and colleagues (2023) to our setting – as described in *Section 3* – yields performance improvements in SciCite compared to unweighted voting strategies, but produces mixed results in ACL-ARC. This discrepancy may stem from differences in the alignment of base models' predictions between validation and test splits in ACL-ARC, probably due to the low number of instances, whereas SciCite exhibits greater consistency[23]. Accordingly, we argue that the geometric framework's efficacy in finding optimal weights is related to the stability of base classifiers' outputs, which is more easily achievable in datasets with a larger number of datapoints.

Overall, these results emphasize the necessity of selecting aggregation strategies meaningful w.r.t. the characteristics of the dataset and the desired level-0 aggregation. While FFNN's adaptability makes it proficient for solving our task on SciCite and ACL-ARC by effectively combining level-0 models meant to work in a complementary manner, ensemble methods like *StackingC* and the *Geometric Framework* may be advantageous in scenarios where base models are not meant to provide synergistic representations of the same object (e.g., in case of homogeneous base models), or in situations in which the dataset do not suffer from severe class imbalance.

### 4.2.1 Computational Instability: Observations on Robustness and Replicability

To further demonstrate the robustness of our strategy, beside its application to two different datasets in two different settings, we computed the average scores and their standard deviations across multiple runs of our classifiers in WS setting (see ***Table 5***). For the SciCite dataset, mean accuracy and Macro-F1 scores are $90.53$ ($\sigma = 0.14$) and $89.34$ ($\sigma = 0.18$), respectively. These results not only surpass the previous SOTA model's performances (***Table 6***) when averaged, but also demonstrate remarkable consistency, as our strategy surpasses SOTA in all ten experimental runs, demonstrating both replicability and reproducibility for SciCite (as outlined in *Section 3.4*).

Similarly, for the ACL-ARC dataset, the mean accuracy and Macro-F1 scores are $80.44$ ($\sigma = 0.44$) and $73.71$ ($\sigma = 1.61$), respectively. Despite the relatively high standard deviation in Macro-F1, CiteFusion consistently outperformed SOTA models across all ten runs in both metrics (see ***Tables 5*** and *7*), further confirming the replicability and dependent reproducibility of our framework and results, and additionally demonstrating conceptual replicability (see *Section 3.4*). The high standard deviation in Macro-F1 for ACL-ARC can be attributed to the extreme imbalance and variability among the 6 classes in the dataset, which poses challenges for classification tasks. Specifically, *Motivation* and *Future*, both underrepresented classes in ACL-ARC, exhibit greater fluctuations in SciBERT-based models' performance across runs, contributing to the higher variance observed in Macro-F1 scores. Nevertheless, the consistent superiority of our strategy highlights its ability to effectively handle such complexities and provide reliable performances even when applied to extremely imbalanced datasets.

Additionally, we collected intermediate training results of level-0 models across all ten runs for both datasets. The validation loss trends over these ten runs are visualized for each of the 6 level-0 models fine-tuned on SciCite in ***Figure 3*** and for each of the 12 level-0 models fine-tuned on ACL-ARC in ***Figure 4***.

[23] To clarify this point, it is important to note that the SciCite dataset contains a significantly larger number of instances in both the validation and test splits compared to ACL-ARC (see *Table 1*). This difference in dataset size has implications for the robustness of weight computation through the geometric framework which, we recall, is computed by means of the validation split for each dataset. In the case of SciCite, the larger sample size ensures that even if outliers are present within the dataset, their impact on the class-specific weight calculation remains minimal due to the averaging effect across a greater number of instances. Conversely, in ACL-ARC, the relatively smaller number of instances makes the weight computation more sensitive to outliers and even a small number of anomalous predictions from level-0 models can disproportionately influence the results.

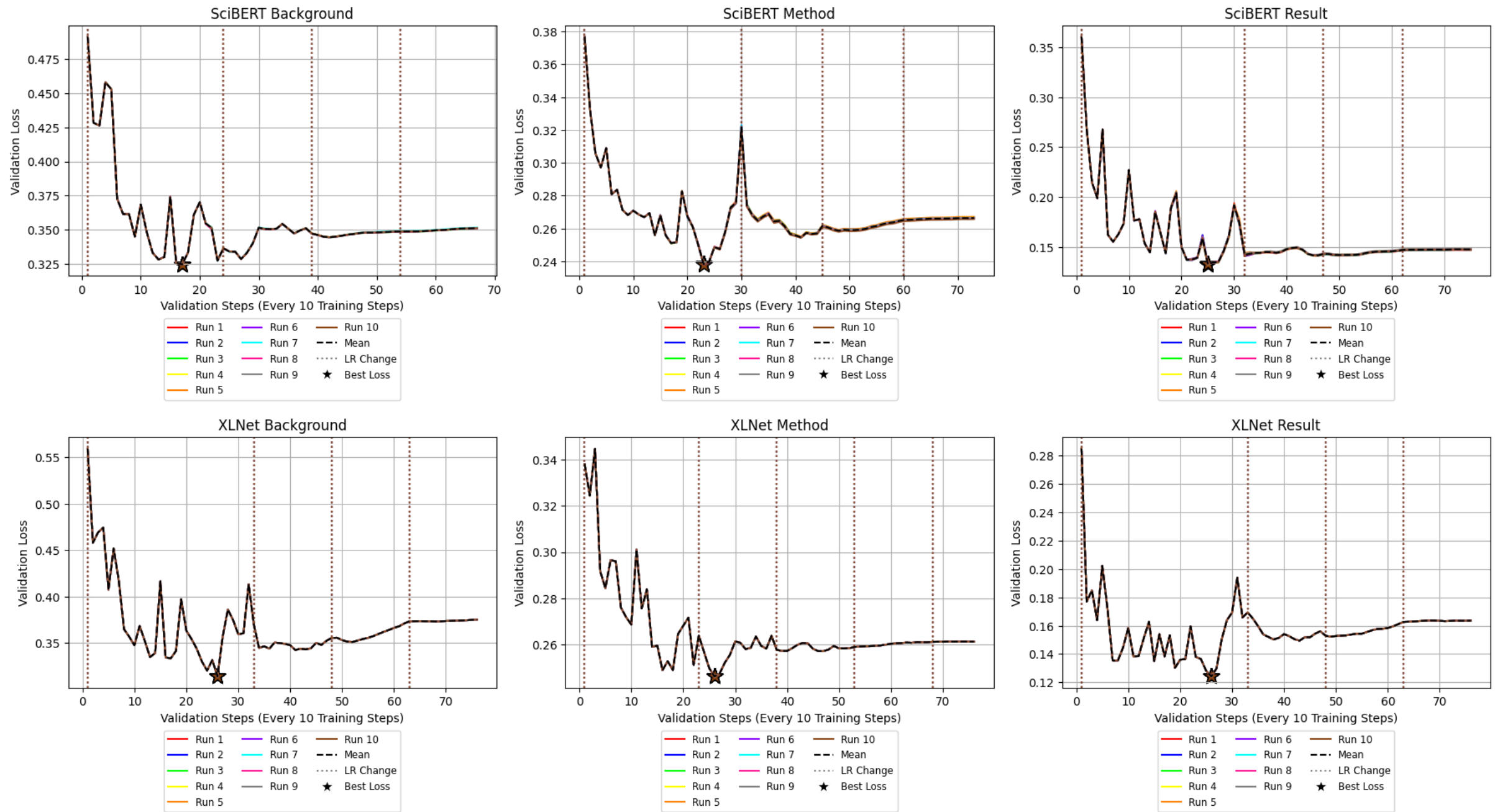

***Figure 3***. *Visual comparison of validation losses across the fine-tuning steps for level-0 models trained on the SciCite dataset. The y-axis represents the validation loss scores, while the x-axis denotes the fine-grained evaluation steps. All models exhibited stable performance across the ten runs, as evidenced by the near-perfect overlap of the 11 lines in each plot (10 corresponding to individual models and one dashed line representing the mean). This overlap indicates consistent performance across validation steps. However, minor instability is observable in the SciBERT-based models after reaching the minimum validation loss. Upon closer inspection, slight separations between the lines at certain points suggest minimal variability in performance, though the overall stability remains completely unaffected since the best state of the models was yet saved when they reached their respective minimum.*

***Figure 3*** illustrates that, across the repeated experimental runs on the SciCite dataset, all the ten base models within each class consistently converge to the same minimum in validation loss, with minimal variations recorded for SciBERT-based models (for full detailed reports of loss minima across runs, we refer the reader to ***Appendix A.2***). Consequently, their performance remains stable across various applications of the entire ES, with a small standard deviation in final performance scores. Since we used fine-grained evaluations, the final loaded state of these models is the one recorded at the evaluation step in which each of them reaches the minimum in validation loss. Thereby, the fluctuations in ***Figure 3*** are either negligible or minimal, resulting in equally robust ECs.

A similar trend is evident in ***Figure 4*** for most of the models trained on the ACL-ARC dataset, but with slight variations that account for the relatively higher standard deviation observed in this dataset. While the plots in ***Figure 4*** confirm that XLNet models exhibit perfect stability across different runs for different classes, they also reveal differences in training dynamics for certain classes. Minor discrepancies can be noted in SciBERT's performances for the *Uses* and *Extends* classes, even though these do not significantly impact the overall performance of the ECs. Indeed, these plots indicate that the loss diverges at certain points, but they also confirm that minima were consistently reached prior to such divergence. The variations in performance should be primarily attributed to SciBERT's handling of the *CompareOrContrast*, *Motivation*, and *Future* classes, where higher differences in minima are observed across runs. Class-specific minima in validation loss for each run are provided in ***Tables A.1 and A.2*** of ***Appendix A.2*** to facilitate a detailed comparison with the visual trends depicted in ***Figures 3 and 4***.

Our results and analyses of computational instability confirm the robustness and adaptability of our strategy across both datasets, assessing conceptual replicability. Furthermore, the use of fine-grained evaluations and mixed precision demonstrates minimal impact on performance, robustness, and replicability of CiteFusion in imbalanced settings. These findings also highlight the reliability of our binarized couple-based ensemble strategy under conditions of severe imbalance, and even with a relatively low number of datapoints to tune level-0 models.

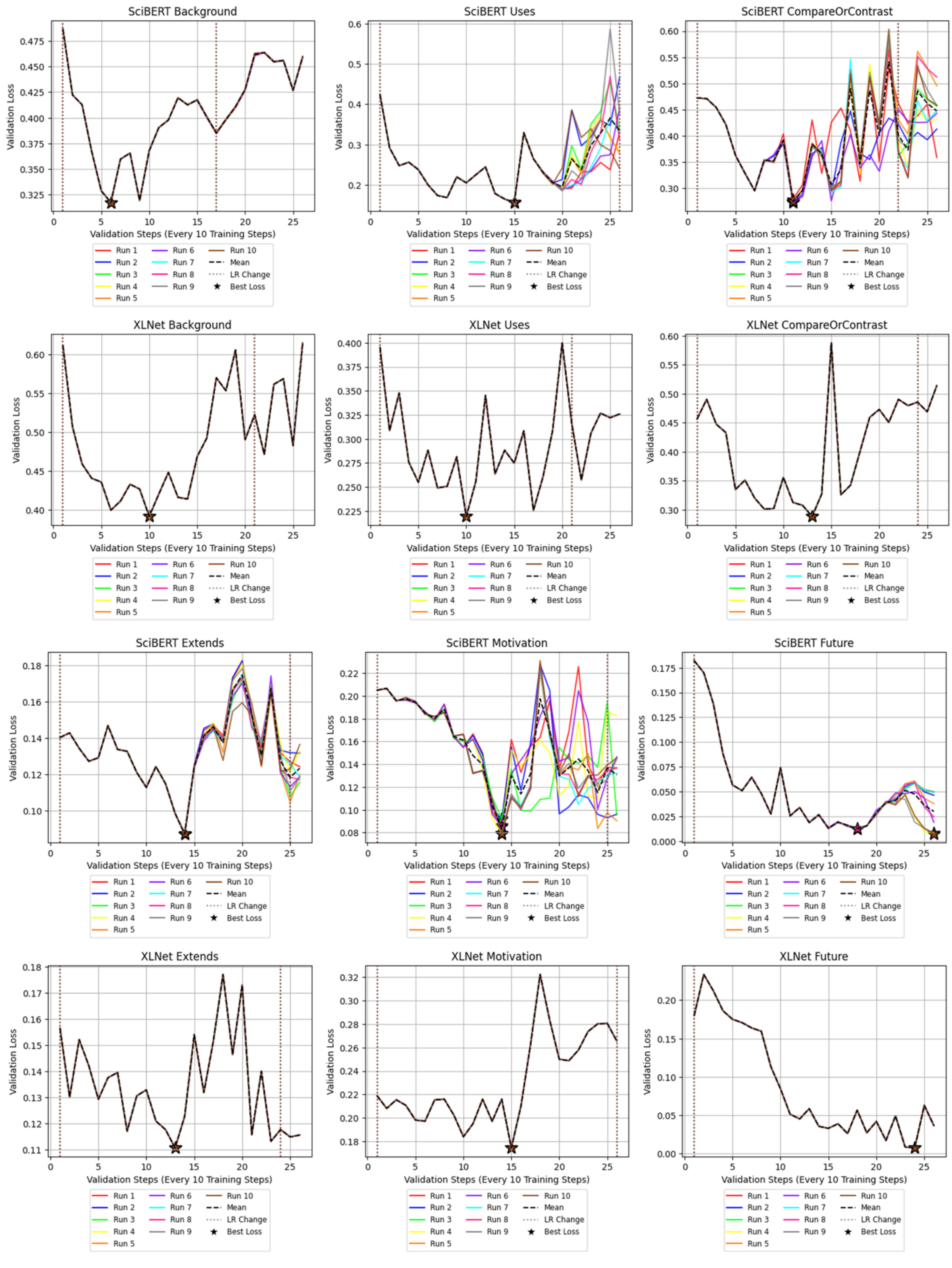

***Figure 4***. *A visual comparison of validation losses across the fine-tuning steps for level-0 models trained on the ACL-ARC dataset reveals notable patterns. Specifically, in certain SciBERT-based models, variations in validation loss can be observed across different evaluations and experimental runs. For instance, the plots for SciBERT Motivation and SciBERT Future illustrate how the loss diverges during training in some instances, highlighting inconsistencies in convergence behavior across runs. These observations underscore the variability in training dynamics among specific model-class combinations.*

### 4.3 Discussing the Classification Process of CiteFusion

Our analyses of SHAP results (reported in ***Appendix A.1***) provide the top 15 tokens used by each model as features to positively classify an input sentence to their class. We can formally distinguish the two datasets in this level-0 discussion for a better understanding of the outcomes and to better asses the contributions of adding section titles within input sentences. Then, the discussion of the results related to the metaclassification process will be presented for both datasets together.

**SciCite**

Our analysis reveals that XLNet, with its standard vocabulary, prioritizes more general tokens, while SciBERT, leveraging domain-specific training and its specialized SciVOCAB (Beltagy et al., 2019), focuses more on tokens specific to the scientific discourse. Accordingly, SciBERT identifies a broader and more specialized set of features (over 11,000), compared to XLNet's smaller and more generic set (about 9,700). This contrast is particularly evident when examining the most influential tokens for each class (see ***Figures A.1*** and ***A.2***). The implications of this divergence will become clearer in the subsequent section, where we analyze some misclassified citations.

We also find that models trained on the *Background* class identify less informative and less intent-specific features compared to those for *Method* and *Result*. For example, top-contributing tokens for *Background* – such as "*circumference*", "*ramification*", or "*difficulty*" – lack clear semantic attribution. In contrast, *Method* and *Result* classes are marked by semantically coherent terms like "*methodology*" or "*used*" (*Method*), and "*agrees*" or "*confirms*" (*Result*), which are closely tied to their respective conceptual domains[24] (Trier, 1931). This pattern is consistent across both WS and WoS settings. In addition, mean SHAP values for *Background* are negative in all cases, while *Method* and *Result* classes show positive means – particularly in WS experiments. This suggests that *Background* sentences often lack strong discriminative features, possibly due to higher variability in their contexts and fewer intent-specific cues.

SHAP analyses also show distinct patterns across settings and model configurations. For *Background*, the total SHAP attribution is negative for both XLNet and SciBERT, indicating that models tend to identify exclusionary features ("*what is not*" *Background*) rather than defining cues for this class. A similar trend is also observed in SciBERT's *Method* and *Result* models under the WoS setting, where the total SHAP attributions are negative; totals for these two classes are instead positive for XLNet-based models. However, when section titles are included (WS setting), SciBERT models better capture positive, intent-specific features, shifting SHAP values toward more positive attributions for all classes – including *Background*. In contrast, XLNet exhibits only minor changes while still maintaining the positive sums for *Method* and *Result*, likely because its standard vocabulary does not emphasize the contextual significance of section titles.

After incorporating section titles, we also observe a reduction in the overall SHAP scores for the XLNet-*Method* model: its total score drops from $+32.12$ (WoS) to $+25.65$ (WS), indicating that tokens that were previously influential, now become less prominent. Similar but smaller shifts are seen for XLNet-*Background* (from $-40.45$ to $-39.3$) and XLNet-*Result* (from $+11.71$ to $+10.96$). These suggest that while the impact of section titles on token contributions varies across models, it is consistently present.

**ACL-ARC**

The analyses of SHAP values at level-0 for the ACL-ARC dataset reveal patterns that are mostly consistent with those observed in SciCite**,** but also highlight several unique behaviors specific to the different citation functions provided in the schema of this dataset. Importantly, our findings confirm that many of the most influential features are aligned with the class-specific grammatical cues identified by Jurgens et al. (2018), reinforcing the validity of our results.

[24] The intended conceptual domains are the ones described by the authors as descriptions of the labels used for their schemes in SciCite (Cohan et al., 2019) and ACL-ARC (Jurgens et al., 2018).

In ACL-ARC, SciBERT-based models still exhibit domain specificity in the tokens they identify, even though this effect is less pronounced than in SciCite, and often appears more subtle. In the WS setting, SciBERT models tend to assign higher SHAP values to intent-relevant tokens, a trend that becomes particularly evident when comparing common tokens between WS and WoS settings. For example, the token "*Following*" for the *Uses* class receives a higher mean score in WS ($+0.45$) compared to WoS ($+0.36$), while "*limitation*" for *CompareOrContrast* rises from $+0.03$ (WoS) to $+0.14$ (WS), underscoring the growing interpretive role of these features from WoS to WS. Notably, such intent-specific tokens are absent among XLNet's top features, as XLNet models prioritize broader and more generic linguistic terms to guide their classifications. This is evident in the use of terms like "*analogous*" or "*contrast*" for *CompareOrContrast*, and "*utilize*" or "*uses*" for the *Uses* class, reflecting XLNet's reliance on broader linguistic features rather than intent-specific cues.

Our analysis revealed that the *Background* class lacks a precise characterization of its intent also in ACL-ARC, mirroring the ambiguity previously observed in SciCite. By contrast, the other five classes are clearly defined by their top features. For example, the *Uses* class in ACL-ARC is strongly associated with terms like those that define the *Method* class in SciCite (e.g., "*utilize/s*" or "*applied*"), supporting the functional alignment between these categories that we described through CiTO (see ***Table 2***). Similarly, the other classes exhibit distinct token-based characterizations, like "*share*" or "*similar*" for the *CompareOrContrast* intent, "*elsewhere*" or "*earlier*" for the *Extends* class, "*inspired*" or "*motivated*" for *Motivation*, and "*promising*" or "*direction*" for the *Future* intent.

Our analysis of model-specific tokens also reveals significant overlap with the grammatical features identified by Jurgens et al. (2018) for four classes in the ACL-ARC dataset. Specifically, the authors highlighted terms such as "*inspire/d*" for the *Motivation* class, which aligns with our findings in the top contributing features of our *Motivation*-specific models. Similarly, we observed comparable patterns for the *Uses* class with terms like "*use/s*" and "*follow/ing*", as well as for *CompareOrContrast* with "*similar*", and for the *Extends* intent with "*previous/ly*" and "*extend*". Beyond these base grammatical features, our analysis extends to include additional terms that enrich the semantic characterization of each class. For instance, the *Motivation* class incorporates terms such as "*address*", "*claim*", and "*find*", while the *Uses* class includes "*implementation*", "*apply/ied*", and "*employ*". In the case of *CompareOrContrast*, we identified terms like "*advantage*", "*outperform*", and "*analogous*", and for *Extends*, we found "*earlier*" and "*follows*". However, for the *Background* and *Future* classes, no grammatical features were provided by Jurgens and colleagues (2018), rendering a direct comparison with our results unfeasible.

Finally, the analysis of the overall token contributions is consistent with our previous observations regarding the inclusion of section titles. For most classes, in WS setting, SciBERT models show an improved ability to capture intent-specific positive features, likely due to their domain-focused vocabulary and training. However, improvements for *Extends* and *Future* are limited, likely due to the low number of training examples for these classes. This underrepresentation may limit the model's capacity to precisely define these classes, even after the incorporation of section titles, although it still assists in identifying what they do not represent, further supporting the decrease of the recall score for *Future* in WS setting when compared with WoS, as noted before. For XLNet, the main change after incorporating section titles is seen in the *Background* class, where a substantial negative shift in SHAP values (over 8 points) suggests a major recalibration in how this model interprets *Background* instances.

## MetaClassification

The analyses of the FFNN metaclassifiers offer key insights into the aggregation of predictions from base models. In SciCite, across both WS and WoS settings, the metaclassifier reliably selects the appropriate level-0 models to determine citation intents. As illustrated in ***Figures 5*** and ***6***, the FFNN assigns strong positive weights to the outputs of the PLMs specialized for the target class, while applying negative weights to predictions from models focused on other classes. This behavior demonstrates effective integration and discrimination among the ensemble's constituent models.

The only exception is in classifying *Method* citations, where the metaclassifier tends to assign a relatively low positive importance to XLNet-*Result*. A plausible explanation is that the XLNet-*Result* model has identified certain textual patterns or signals that also characterize *Method*-type citations. As a result, when the metaclassifier encounters a high *Result*-class score from XLNet, it interprets this as weak but positive evidence for categorizing the citation as *Method*. Essentially, some linguistic cues used by XLNet-*Result* might overlap with the way in which *Method*-type citations are described. This misinterpretation is corrected with the inclusion of section titles, further demonstrating their utility. This is depicted in ***Figure 6***, where the metaclassifier no longer misinterprets XLNet-*Result* predictions as pertaining to the *Method* class.

In classifying *Method*-type citations, the FFNN's reliance on XLNet-*Method* also becomes less significant after section titles are included. As noted before, this adjustment corresponds to a general reduction in the token-level scores produced by XLNet-*Method*, resulting in a more balanced and conservative contribution to the ensemble. Consequently, while binary classification by XLNet-*Method* is less dominant, the overall ensemble benefits from reduced misclassification. This trend is mirrored in the class-specific precision and recall scores in ***Table 8***, in which the *Method* class is the only intent to improve in both metrics from WoS to WS.

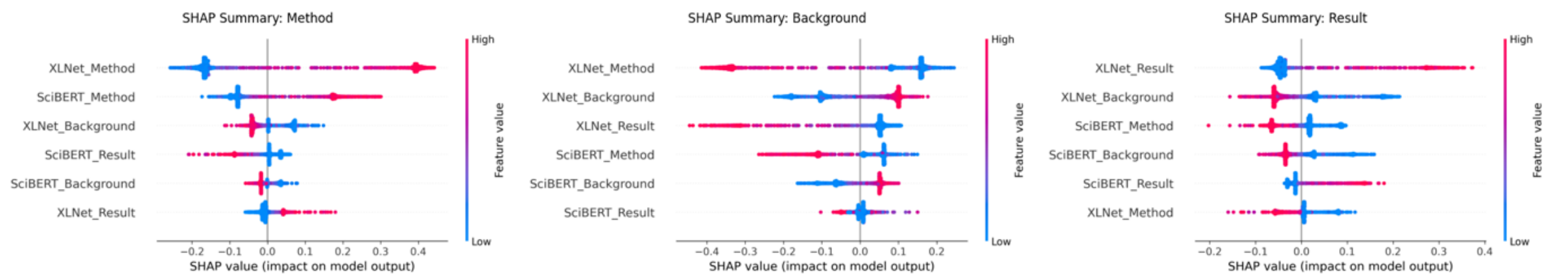

***Figure 5***. *SHAP summary plots illustrating the contribution of each base model in classifying various intents within the SciCite dataset for the* ***WoS setting***.

An illustrative example of the "*what is not*" kind of classification can be seen with XLNet-*Method* when it helps to identify *Background*-class sentences (see ***Figure 5***). Here, low XLNet-*Method* scores (blue dots) are used as strong evidence for the *Background* class, while high scores (red dots) serve as evidence against it. This demonstrates that the metaclassifier leverages negative evidence – absence of *Method*-specific features – to inform its decisions about the classification of *Background*-type citations, highlighting its efficiency in defining classification boundaries, but also reinforcing the hypothesis on the lack of intent-specific cues for the *Background* class.

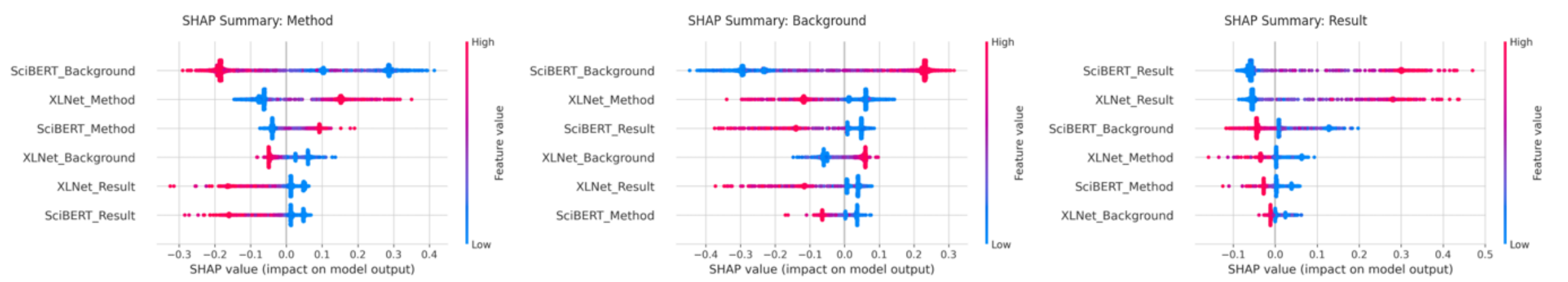

***Figure 6***. *SHAP summary plots illustrating the contribution of each base model in classifying various intents within the SciCite dataset for the* ***WS setting***. *As you can notice when comparing with Figure 5, XLNet-Result model is no more misinterpreted when it comes to classify Method class instances.*

This precise identification of the appropriate models to classify each class generally extends also to ACL-ARC, with few notable exceptions (see ***Figure 7***). For the *Background* class, the metaclassifier incorrectly attributes negative importance to XLNet-*Background* predictions. Conversely, in classifying *CompareOrContrast* instances, models such as XLNet-*Future*, XLNet-*Background*, and XLNet-*Extends* contribute positively alongside the XLNet-based model specifically tuned to recognize this class, while SciBERT-*CompareOrContrast* plays a smaller role than expected, suggesting a potential overlap in features across different intent categories.

Examining ***Figure 7*** further, the ensemble model demonstrates effective use of both positive and negative contributions from different features in the most common classes (*Background*, *Uses*, *CompareOrContrast*), reflecting its reliance on a diverse set of signals. However, for underrepresented classes (*Extends*, *Motivation*, and *Future*) the metaclassifier relies almost exclusively on the expert models trained for those intents, with little input from other models. This suggests a potential limitation in the model's ability to generalize or utilize broader contextual signals for less frequently occurring classes, which is probably shaped from an overweighing of positive probabilities produced by the base models specifically tuned on these classes due to their underrepresentation.

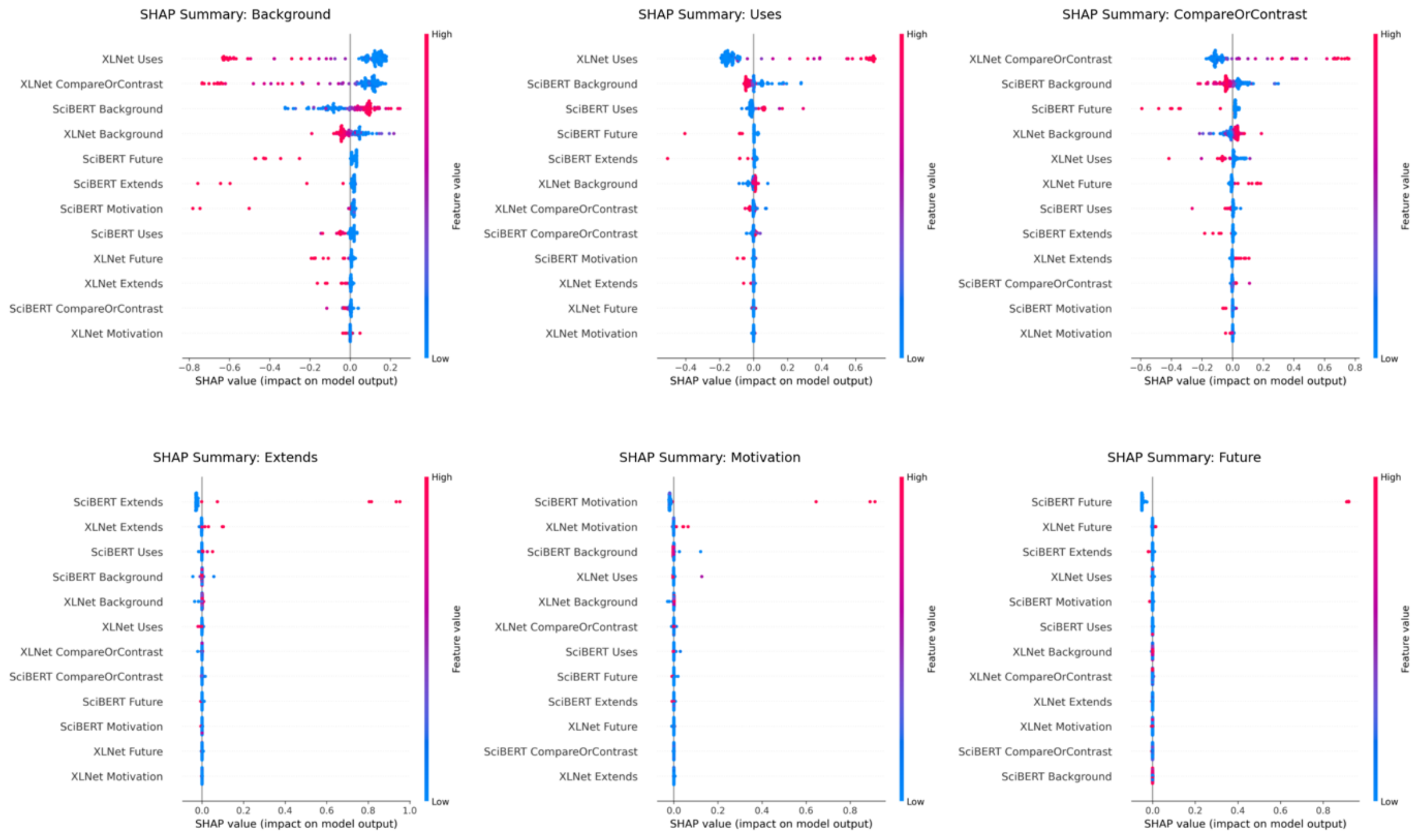

***Figure** 7. SHAP summary plots depicting the contribution of each base model in classifying various intents within the ACL-ARC dataset for the **WoS setting**.*

Once again, incorporating section titles results beneficial for the overall predictions by mitigating the influence of individual token contributions and improving the alignment between base models and target classes. In ***Figure 8***, the prior misuse of XLNet-*Background* as a negatively contributing feature to *Background*-class identification is corrected after section titles are included. Similarly, for *CompareOrContrast*, the misinterpretations involving XLNet-*Future*, XLNet-*Background*, and XLNet-*Extends* are fully resolved for the first two models, and substantially mitigated for the third.

However, a challenge persists with XLNet-*Extends*, which contributes negatively to *Extends* classifications after section titles are introduced. While this problem did not exist in the WoS setting, it does not notably affect overall classification, as class-specific scores for *Extends* still improve in WS w.r.t. WoS setting. Additionally, earlier challenges where SciBERT-*Uses* was incorrectly aligned with *Extends* have been fixed in the WS setting. Nonetheless, a new issue arises with XLNet-*Uses* since its predictions are positively aligned with the *Extends* class, despite expected to be the opposite, but the associated SHAP values are small and therefore minimally impactful.

Two notable issues emerge instead for the *Motivation* and *Future* classes, where the metaclassifier's ability to correctly classify instances diminishes, as reflected in the performance scores presented in ***Tables** 7* and ***9***. In these cases, even if the metaclassifier appears to appropriately utilize relevant features and negatively weigh additional ones, its performance for these classes declines. This suggests that, for rare intents, a simpler, more direct reliance on expert models (as in WoS) might be preferable to a broader ensemble strategy.

In summary, these findings enable us to assess the role of section titles before proceeding to examine some misclassifications and errors. Our analyses demonstrate that incorporating section titles consistently improves classification performance across all the experiments, boosting both accuracy and macro-F1. Acting as *framing devices*, section titles influence how base models process input sentences – an effect particularly strong in SciBERT-based models, which benefit from domain-specific vocabulary and extended context and undergo a major perceptual shift. While XLNet models also improve, the gains are more modest, likely due to the general-domain pretraining and standard vocabulary of this PLM. Overall, the WS setting consistently outperforms the WoS configuration across both datasets.

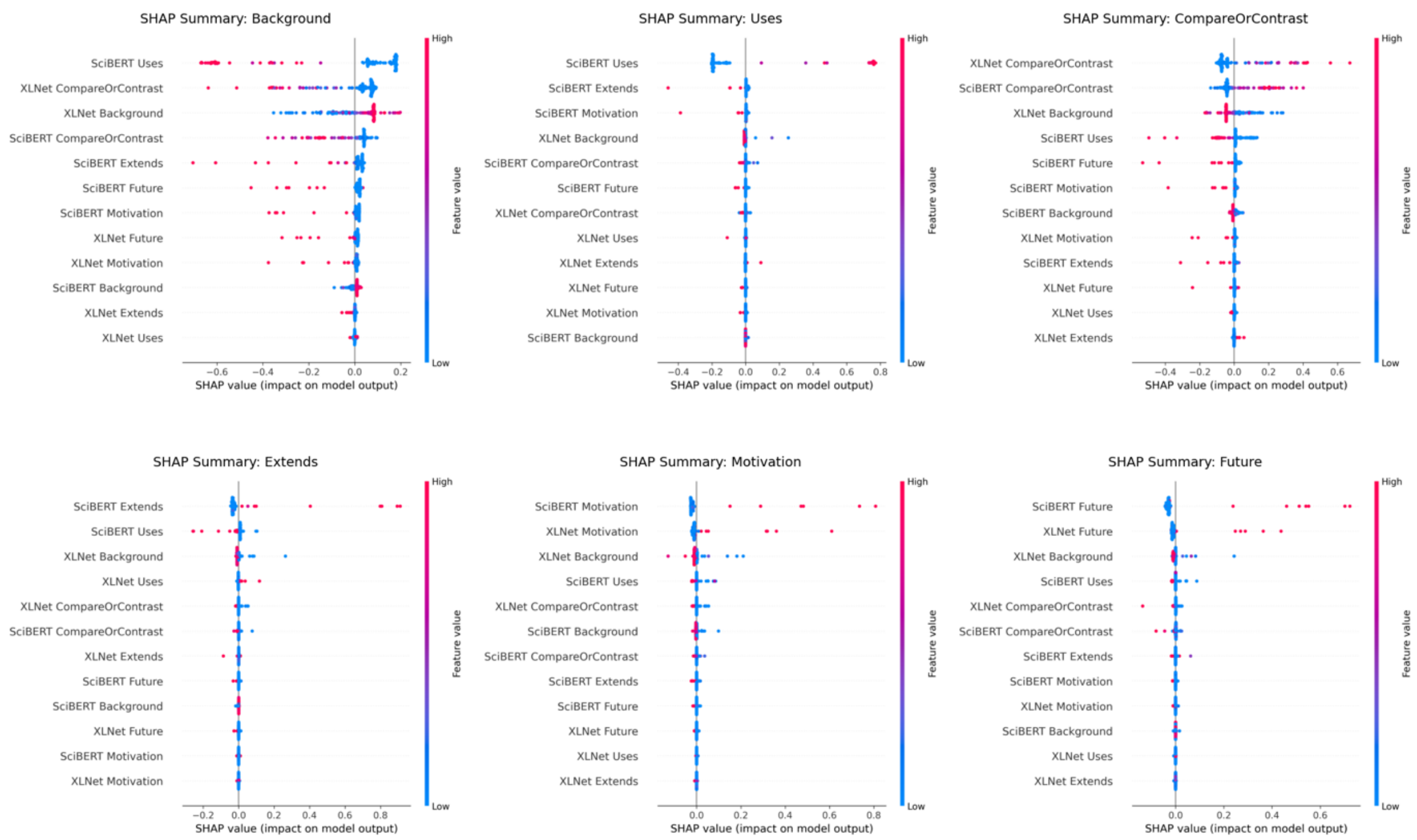

***Figure 8**. SHAP summary plots depicting the contribution of each base model in classifying various intents within the ACL-ARC dataset for the **WS setting**.*

### 4.4 Insights from SHAP: Misclassification Analysis

SHAP values, as utilized so far, demonstrated and provided support for elucidating the general classification dynamics of CiteFusion, while supporting the use of section titles as framing devices. This section, however, discusses CiteFusion's strengths and limitations, also presenting explanations of misclassified instances. We conducted a comprehensive error analysis using both interpretability metrics (attribution masses and their correlation in class-specific settings) and classic evaluation tools (confusion matrix). For simplicity, all the experiments discussed in this section were conducted on the CiteFusion ensemble model developed for the SciCite dataset in WS setting.

#### 4.4.1 General Classification Patterns of CiteFusion

In *Section 4.3*, we observed that SciBERT and XLNet differ in their processing of citation contexts at token level. While SciBERT tends to focus on specialized scientific terms, XLNet relies on broader and more general language features. Furthermore, we noted that the *Background* class exhibits lower and often negative SHAP attribution sums, with less distinctive or semantically coherent feature sets than the *Method* and *Result* classes, which benefit from stronger class-specific signals.

To frame these token-level observations in a more global perspective, aimed at better explaining the classification dynamics of CiteFusion and the nature of intent-specific citation contexts in SciCite, we now examine *attribution mass* – defined as the sum of the SHAP values assigned to all tokens in a specific citation sentence by a given base binary classifier for its predicted class. Intuitively, a higher absolute attribution mass indicates that a level-0 model has found stronger evidence in the input sentence to justify its binary decision. In contrast, low attribution mass may signal uncertainty or diffused evidence, coinciding with ambiguous and challenging cases. This measure thereby provides a sense of how confidently a model supports its predictions for a class on a given input, and enables us to investigate model agreement and typical patterns of error.

***Table 10*** reports the *mean positive*, *negative*, and *signed* attribution masses computed per expert and grouped by the ensemble's predicted class. Each *expert* corresponds to a level-0 binary model trained to detect a specific SciCite's intent (*Method*, *Background*, or *Result*), and produces an attribution mass for every input sentence, regardless of the ensemble's decision. This table demonstrates that for each predicted class, the corresponding class-specific expert tends to produce the highest positive mean attribution mass and the lowest negative mean mass – resulting in the largest mean signed mass values. This trend is especially pronounced for the *Method* and *Result* classes.

These findings are also illustrated in the boxplot shown in ***Figure A.7*** (***Appendix A.3***), which offers a clear, visual comparison of mean signed attribution masses among experts for each predicted class.

| **Predicted Class** | **Intent-Specific Expert** | **Mean Positive Mass ($\pm$std)** | **Mean Negative Mass ($\pm$std)** | **Signed Mass ($\pm$std)** |
|---|---|---|---|---|
| **Method** | **SciBERT-Method** | 0.8346 ($\pm$ 0.136) | 0.0204 ($\pm$ 0.0367) | 0.8143 ($\pm$ 0.1607) |
| | **XLNet-Method** | 0.9102 ($\pm$ 0.122) | 0.0707 ($\pm$ 0.0694) | 0.8394 ($\pm$ 0.1551) |
| | SciBERT-Background | 0.0349 ($\pm$ 0.0428) | 0.6593 ($\pm$ 0.1067) | -0.6244 ($\pm$ 0.1357) |
| | XLNet-Background | 0.0540 ($\pm$ 0.0522) | 0.9408 ($\pm$ 0.0931) | -0.8869 ($\pm$ 0.12) |
| | SciBERT-Result | 0.0268 ($\pm$ 0.0663) | 0.0551 ($\pm$ 0.0205) | -0.0283 ($\pm$ 0.0584) |
| | XLNet-Result | 0.0381 ($\pm$ 0.1166) | 0.0215 ($\pm$ 0.0491) | 0.0166 ($\pm$ 0.0802) |
| **Background** | SciBERT-Method | 0.0895 ($\pm$ 0.1177) | 0.1045 ($\pm$ 0.0286) | -0.0150 ($\pm$ 0.1192) |
| | XLNet-Method | 0.0906 ($\pm$ 0.1556) | 0.0551 ($\pm$ 0.0704) | 0.0355 ($\pm$ 0.106) |
| | **SciBERT-Background** | 0.2355 ($\pm$ 0.0635) | 0.1012 ($\pm$ 0.1172) | 0.1343 ($\pm$ 0.1656) |
| | **XLNet-Background** | 0.0738 ($\pm$ 0.0678) | 0.2291 ($\pm$ 0.2574) | -0.1553 ($\pm$ 0.2275) |
| | SciBERT-Result | 0.0568 ($\pm$ 0.1116) | 0.0497 ($\pm$ 0.0185) | 0.0072 ($\pm$ 0.1076) |
| | XLNet-Result | 0.0851 ($\pm$ 0.1781) | 0.0278 ($\pm$ 0.0522) | 0.0572 ($\pm$ 0.1431) |
| **Result** | SciBERT-Method | 0.0712 ($\pm$ 0.0984) | 0.1115 ($\pm$ 0.0304) | -0.0403 ($\pm$ 0.1001) |
| | XLNet-Method | 0.0725 ($\pm$ 0.1591) | 0.0566 ($\pm$ 0.0708) | 0.0158 ($\pm$ 0.1091) |
| | SciBERT-Background | 0.0656 ($\pm$ 0.0673) | 0.6358 ($\pm$ 0.1749) | -0.5702 ($\pm$ 0.226) |
| | XLNet-Background | 0.0417 ($\pm$ 0.042) | 0.9493 ($\pm$ 0.097) | -0.9076 ($\pm$ 0.1112) |
| | **SciBERT-Result** | 0.8003 ($\pm$ 0.1623) | 0.0206 ($\pm$ 0.0267) | 0.7797 ($\pm$ 0.1816) |
| | **XLNet-Result** | 0.9342 ($\pm$ 0.1374) | 0.0535 ($\pm$ 0.0609) | 0.8807 ($\pm$ 0.1576) |

***Table 10****. Mean positive, negative, and signed attribution mass, with std, for each level-0 model, grouped by the predicted class assigned by the ensemble. Positive mass corresponds to the expert's support for the assigned label, while negative mass reflects its rejection (values must be considered as negative). Signed mass is computed as the difference between positive and negative values.*

Accordingly, when the ensemble predicts *Method*, SciBERT- and XLNet-*Method* experts produce high positive signed attribution masses of $0.8143$ and $0.8394$ respectively. This demonstrates that these level-0 models find, on average, strong positive discriminative signals that helps them to identify this class, and that they learned some intent-specific patterns. In contrast, *Background* experts exhibit high negative signed masses ($-0.6244$ and $-0.8869$), reflecting a confident rejection of the *Background* hypothesis for these instances. *Result*-specific experts, meanwhile, produce near-zero attribution masses ($-0.0283$ and $0.0166$), suggesting minimal influence. A similar dynamic holds also for *Result*-type predictions, where class-specific experts find supporting evidence, while *Background*-based models find contrasting features for their class, and *Method* experts play a marginal role.

However, for *Background* predictions, the signed attribution masses are more sparse and generally low. While SciBERT-*Background* reports the highest signed mass for this class ($0.1343$), it remains relatively low, and XLNet-*Background* identifies a negative signed mass ($-0.1553$) on average. This pattern aligns with our token-level findings (see *Section 4.3*), where *Background* was associated with weaker discriminative signals. In addition, non-*Background* experts, such as XLNet-*Result* ($0.0572$) or SciBERT-*Method* ($-0.015$), also produce low and near-zero values. Together, these results support the hypothesis that *Background* is primarily identified through exclusion – therefore by the absence of strong evidence for the other classes.

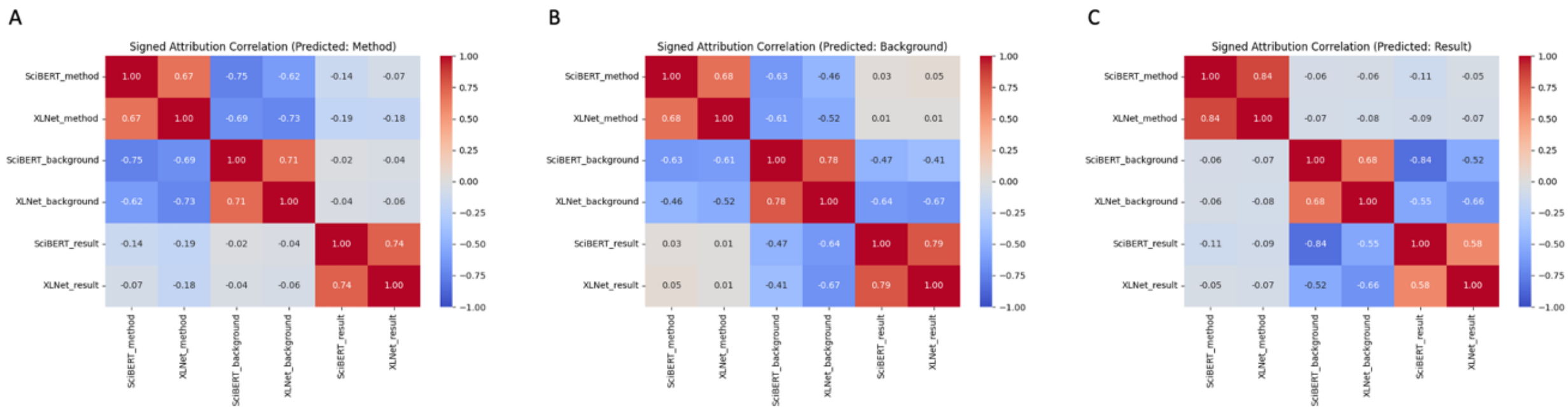

***Figure 9***. *Signed attribution mass correlations between experts, conditioned on the ensemble's predicted class. Each heatmap shows the pairwise Pearson correlation of signed attribution masses across all six expert models, restricted to sentences classified as* ***(A)*** *Method,* ***(B)*** *Background, or* ***(C)*** *Result by the ensemble. High positive correlations (red) indicate agreement in attribution direction, while negative values (blue) denote contrasting patterns.*

These observations are further confirmed by ***Figure 9***, which presents class-specific correlation heatmaps between signed attribution masses. When the ensemble predicts *Method* instances (***Figure 9.A***), *Method* experts are strongly correlated with each other and negatively correlated with *Background*-tuned models, supporting the previously discussed mutual rejection dynamic. Similarly, for *Background*-type predictions (***Figure 9.B***), *Background* models are positively aligned, while *Method* and *Result* experts show negative correlation with them – indicating that, even with low signed attribution masses (see ***Table 10***), these experts play a significant exclusionary role in classifying *Background* instances. For *Result* classifications (***Figure 9.C***), the heatmap reveals a negative correlation between *Result* and *Background* experts, along with the agreement between *Result* models.

Overall, ***Figure 9*** shows that each intent category displays internal agreement among models tuned on the same class, while attribution masses across different categories are generally mutually exclusive and negatively correlated. This effect is most pronounced for the *Background* experts, which consistently exhibit negative correlation with both *Result*- and *Method*-tuned models; in contrast, *Method* and *Result* models have only a limited influence on each other, as suggested also by the average attribution masses in ***Table 10***. These patterns confirm that *Background* models primarily function as rejectors when not predicting the *Background* class, and that the absence of both *Method*- and *Result*-specific features has a significant role in identifying *Background* instances. These dynamics are effectively captured by the FFNN aggregator, as demonstrated in ***Figure 6***.

### 4.4.2 Attribution Mass and Misclassification

To better connect model interpretability with classification outcomes, we relate attribution masses to prediction errors at the ensemble level. ***Table 11*** presents the confusion matrix for *CiteFusion* (FFNN aggregator), highlighting that most misclassifications occur between *Method* and *Background*, and between *Background* and *Result*.

| | Method | Background | Result |
|---|---|---|---|
| Method | 538 | 61 | 5 |
| Background | 33 | 902 | 61 |
| Result | 2 | 17 | 240 |

***Table 11***. *Confusion Matrix of the FFNN-based CiteFusion (WS)*

Since ***Figure 9*** demonstrates that *Method* and *Background* models are generally negatively correlated in their attribution masses across the entire test set – mainly when CiteFusion predicts *Method* and *Background* intents – , we hypothesize that *Method*-to-*Background* misclassifications arise when *Method*-tuned models fail to provide strong positive evidences for the citation context at hand. This hypothesis is consistent with the previously discussed exclusionary classification mechanism for the *Background* class – i.e., sentences are assigned to *Background* when they lack distinctive *Method*- or *Result*-type cues.

Therefore, we conducted a brief analysis[25] on some misclassified citation contexts (***Table 12***). This ultimately confirmed our hypothesis, and revealed that *Method*-to-*Background* errors usually occur when *Method* models assign low signed attribution masses – significantly below the average (± std) observed for the prediction of *Method* instances (see ***Table 10***) – to these sentences, while *Background* models, for these same sentences, assign signed attribution masses consistent with their typical range of values. As a result, these misclassifications are due to the absence of *Method*-specific evidences within these sentences, which allows the *Background* models to dominate the ensemble's prediction even with their low signed attribution scores.

| ID | Gold | Predicted | Section | Citation Context | SciBERT MET Masses | XLNet MET Masses | SciBERT BKG Masses | XLNet BKG Masses | SciBERT RES Masses | XLNet RES Masses |
|---|---|---|---|---|---|---|---|---|---|---|
| *1* | *Method* | *Background* | Introduction | Central serotonin (5-HT) depletion induced by electrolytic lesions as well as 5,7- dihydroxy-tryptamine (5,7-DHT) lesions to the medial raphe nucleus attenuate LI (Asin et al. 1980; Lorden et al. 1983; Solomon et al. 1980). | -0.0582 | +0.0052 | +0.2147 | +0.004 | -0.0486 | -0.0029 |
| *2* | *Method* | *Background* | Discussion | All insertions consist of direct repeats of the DNA sequence, which are removed precisely, presumably by recombination or 'slippage' during DNA replication or repair (Farabaugh et al., 1978; Albertini et al., 1982). | -0.0487 | -0.0071 | +0.2315 | -0.0057 | -0.0448 | -0.0036 |
| *3* | *Method* | *Background* | DISCUSSION | Third, AOSD patients in active stage had increased concentrations of IL-6, TNF-, or INF-(11). | -0.0606 | -0.0339 | +0.2337 | -0.0205 | -0.0345 | +0.001 |
| *4* | *Method* | *Background* | Introduction | Several previous studies have indicated that κ and δ opioid receptors can exist not only as monomers but also as multi-protein structures including homodimers, heterodimers, and larger oligomer complexes... | -0.0682 | -0.0152 | +0.2411 | -0.1758 | -0.0235 | +0.0615 |
| *5* | *Method* | *Background* | 1. Introduction | Furthermore, SelM knock-down resulted in decreased cell viability and increased ROS, further demonstrating the functional importance of SelM in preventing oxidative stress [12]. | -0.0352 | -0.0187 | +0.2299 | -0.0256 | -0.0403 | -0.0002 |
| *6* | *Method* | *Background* | DISCUSSION | Magnetobacterium bavaricum" (49) but have recently been recognized to contain bacterial species showing a broad range of morphological properties and with high phylogenetic diversity (10, 25, 27, 30, 31). | -0.0624 | -0.023 | +0.237 | +0.0021 | -0.0403 | -0.0009 |
| *7* | *Method* | *Background* | Results | After secondary review, 93 studies were included in the final report.(5-97) There were no randomized clinical trials. | +0.0436 | +0.0284 | +0.1224 | -0.3975 | +0.0524 | -0.0644 |
| *8* | *Method* | *Background* | Results | Finally, 82 articles [9– 18,32–103] were selected in the meta-analysis, of which 9 articles [14,34,37,42,61,64,71,84,88] had two independent studies and were considered separately. | +0.0699 | +0.1002 | +0.0301 | -0.2667 | -0.0017 | +0.0137 |

***Table 12****. Examples of Method-type citations, misclassified as Background by CiteFusion, presented with expert-specific signed attribution masses. In the headers, we used MET for Method, BKG for Background, and RES for Result.*

Upon inspecting ***Table 12*** and additional related errors, we identified a recurring misclassification pattern: citation contexts that reference papers used as data sources in the citing work's literature review are frequently mislabeled as *Background*, rather than being correctly classified under the *Method* intent (see ***Table 12***, citation contexts ***7*** and ***8***). Such structures, typically, do not provide enough distinctive *Method*-specific features to be accurately categorized as *Method*, therefore explaining model's failure considering the exclusionary classification performed for *Background* instances. This is confirmed when looking at expert-specific signed mass attributions that show extremely low attribution masses from *Method* experts (w.r.t. their average – see ***Table 10***), and in range attribution masses from *Background*-specific models when predicting the *Background* intent.

A closer examination of these misclassified citations highlights the inherent challenges and subjectivity involved in annotating citation intent, for example when sentences labeled as *Method* appear within Introduction or

[25] For the following analyses we consider the description of the SciCite intents provided in the original work: "*The citation states, mentions, or points to the background information giving more context about a problem, concept, approach, topic, or importance of the problem in the field.*" (Background); "*Making use of a method, tool, approach or dataset.*" (Method), "*Comparison of the paper's results/findings with the results/findings of other work*" (Result) (Cohan et al., 2019).

Discussion sections. In these cases, citations may serve to provide background or interpretation rather than to describe methodological details, and this blurs the line between *Method* and *Background* intent categories, resulting in ambiguous annotations. This overlap suggests that current annotation protocols may be too rigid to fully capture the rhetorical functions found in scientific writing, and motivates the consideration of more flexible frameworks – such as allowing for multi-label annotations of citation intents, where sentences can be tagged with more than one function (Lauscher et al., 2021).

Now, we also provide additional examples of misclassifications to further clarify the relationship between the weaknesses of CiteFusion and the expert-specific attribution masses. ***Table 13*** presents some examples from the second largest source of misclassifications, which is from *Background* annotated labels to *Result* predicted intents, as shown in ***Table 11***.

| ID | Gold | Predicted | Section | Citation Context | SciBERT MET Mass | XLNet MET Mass | SciBERT BKG Mass | XLNet BKG Mass | SciBERT RES Mass | XLNet RES Mass |
|---|---|---|---|---|---|---|---|---|---|---|
| 1 | *Background* | *Result* | RESULTS | Consistent with previous observations, all of these factors have been shown to associate with replicated viral DNA (14, 15). | -0.0579 | -0.0404 | -0.6612 | -0.8765 | +0.7714 | +0.4462 |
| 2 | *Background* | *Result* | Discussion | Other authors (Moayeri et al., 2007) compared the monozygotic twinning rate after blastocyst transfer with previously published results from the same institute (Milki et al. | -0.0603 | -0.0063 | -0.6425 | -0.9641 | +0.8373 | +0.9663 |
| 3 | *Background* | *Result* | Discussion | In further contrast to the results of Brown et al. (2008, 2009), a recent magnetoencephalography study (Miyaji et al. 2014) showed a more inferior central sulcus activation related to laryngeal airpuff stimulation. | -0.0678 | -0.0083 | -0.7049 | -0.9565 | +0.9014 | +0.9747 |
| 4 | *Background* | *Result* | Discussion | Similar results between Gla-300 and Gla-100 were also found in other glycaemic response parameters... | -0.0535 | -0.0004 | -0.6951 | -0.9612 | +0.8707 | +0.9636 |
| 5 | *Background* | *Result* | Discussion | Compared with Caucasians, Asians are generally smaller [29], however waist circumference results from the present study contradict this expectation. | -0.0665 | -0.0163 | -0.4760 | -0.5238 | +0.8586 | +0.9125 |
| 6 | *Background* | *Result* | Discussion | It was reported that this final result excluded nine patients who showed sentinel nodes negative intraoperatively but positive postoperatively [20]. | -0.0586 | -0.0071 | -0.1079 | -0.6478 | +0.5314 | +0.7631 |
| 7 | *Background* | *Result* | RESULTS | In general, these observations are consistent with earlier time-resolved FTIR spectra of hR on this time scale (Dioumaev and Braiman, 1997a). | -0.0493 | -0.0192 | +0.0127 | -0.8798 | +0.4980 | +0.9026 |
| 8 | *Background* | *Result* | Discussion | This result deals with carrot, where DcSERK expression is characteristic of embryogenic development up to the globular stage and stops thereafter [17]. | -0.0626 | -0.0120 | -0.6929 | -0.9602 | +0.9043 | +0.9497 |
| 9 | *Background* | *Result* | Discussion | Oh and Barrett-Connor indicated that the BsmI polymorphism may be associated with insulin resistance in a nondiabetic Caucasian population [10]. | -0.0611 | -0.0139 | +0.0479 | -0.8740 | +0.1371 | +0.8897 |

***Table 13***. *Examples of Background-type citations, misclassified as Result by CiteFusion, presented with expert-specific signed attribution masses. In the headers, we used MET for Method, BKG for Background, and RES for Result.*

***Table 13*** shows that most of *Background*-to-*Result* misclassifications are shaped by high attribution masses from the *Result*-specific models, while *Method* experts contribute only marginally. This behavior of *Method*-tuned models is in line with the classification of both *Background* and *Result* instances, according to ***Table 10*** and ***Figure 9***, therefore it does not provide any useful information for the analysis. In contrast, *Background*-specific models behave differently than expected: their attribution patterns are more consistent with their average behavior ($\pm$ std) when identifying *Result* instances. Interestingly, the *Background* models appear to reinforce the *Result* prediction in all these misclassified cases by assigning strong negative attribution to their own class, effectively acting as rejecting experts.

The behavior of *Result* and *Background* experts in ***Table 13*** can be attributed to the nature of these citation contexts. Most of the misclassified contexts (except for instances ***6***, ***8***, and ***9***) involve some form of "*comparison*" which, according to the SciCite schema, is a key aspect of the *Result* intent – even if the definition seem to describe a direct comparison of the cited source with the citing work. Furthermore, many of these citation contexts include token-level features that, as identified by CiteFusion (see *Section 4.3* and ***Figure A.2***), are characteristic of the *Result* class. We hypothesize also that this additional factor may contribute to *Background*-to-*Result* misclassifications, as the model interpret these token-level cues as strong indicators for the *Result* intent. Since the *Result* class is underrepresented in SciCite, both *Background* and *Result* experts may fail to learn the distinctions in how these token-level features (such as "*results*", "*contrast*", "*consistent*", …) and the concept of comparison are used across different contexts, thereby conflating different cases under the same category. As a result, CiteFusion struggles to differentiate between real *Result*-type citations – according to SciCite's schema – and those that only superficially share similar features, leading to these misclassifications.

In conclusion, we further demonstrated that the *Background* intent is characterized by lower attribution masses and less distinctive feature patterns within SciCite. In addition, the misclassified *Method*-type citations, as seen in ***Table 12***, often display weak and non-discriminative evidences for their true class, reinforcing the idea that these errors reflect true ambiguity in the underlying data rather than simple model failure. In addition, we discussed some misclassifications related to the *Result* intent and to its underrepresentation in the dataset. These insights highlight the need for richer annotation frameworks and more balanced datasets to enable the development of more robust and interpretable models for the CIC task.

#### 4.4.3 Local Interpretations of Misclassified Citations

SciCite includes additional metadata related to the annotation process, specifically the confidence levels of annotators for certain sentences. To further enrich our analysis of misclassified citations and deepen the discourse on ambiguity, we extracted all the errors made by CiteFusion (WS)[26] and focused on those instances accompanied by these metadata. This allowed us to evaluate the performance of our model in relation to both less certain annotations and those with complete agreement.

The first sentence under examination is illustrated in ***Figure 10***, where the element preceding the first period (".") represents the title of the section containing the subsequent citation context. The citation is labelled as *Background* with a confidence score of 1 (on a scale from 0 to 1). The plots presented in ***Figure 10*** elucidates how different base models within CiteFusion process and interpret (i.e., weigh) the same citation text, and demonstrates how scores are assigned at both level-0 (binary classification) and level-1 (multi-class metaclassification) stages.

A key insight derived from the analysis in ***Figure 10*** is the different contribution of the same input sentence across various model architectures. As previously discussed, we can further confirm that XLNet tends to emphasize more general tokens, whereas SciBERT assigns greater importance to domain-specific tokens. This distinction is particularly evident when examining, for example, the phrase "*FTIR spectra of hR on this time scale*". SciBERT-based models attribute significant importance – either positive or negative, according to the specific expert analyzing it – to this segment, likely because it relates directly to a scientific domain covered by SciBERT's vocabulary. In contrast, XLNet-based models assign little to no importance to this phrase.

Similarly, elements of the citation such as author names or publication years are considered by SciBERT-based experts to have a slight influence on the final prediction, while they are mostly disregarded by XLNet models. This pattern becomes even more pronounced when analyzing the role of section titles[27]. XLNet models completely overlook the section title in this case, whereas it plays a pivotal role in the predictions generated by SciBERT-based models. Specifically, for SciBERT models, the section title emerges as the most contributing token in classifying the sentence as not pertaining to the *Background* class, while ranks as the third most influential token in classifying it as belonging to the *Result* class. This further underscores the critical role of these contextual elements in enhancing the performance of domain-specific models like SciBERT.

[26] Out of the total 1,859 test instances, our model misclassified 179 instances.

[27] In ***Figure 10***, the input sentence reported under the XLNet models contains certain "grammatical" errors. It is important to note that these discrepancies are not related to the classification process or the computation of SHAP values. Instead, they arise during the detokenization of the input performed by SHAP to generate the text plot. Consequently, these errors have no bearing on the classification outcomes or the associated scores.

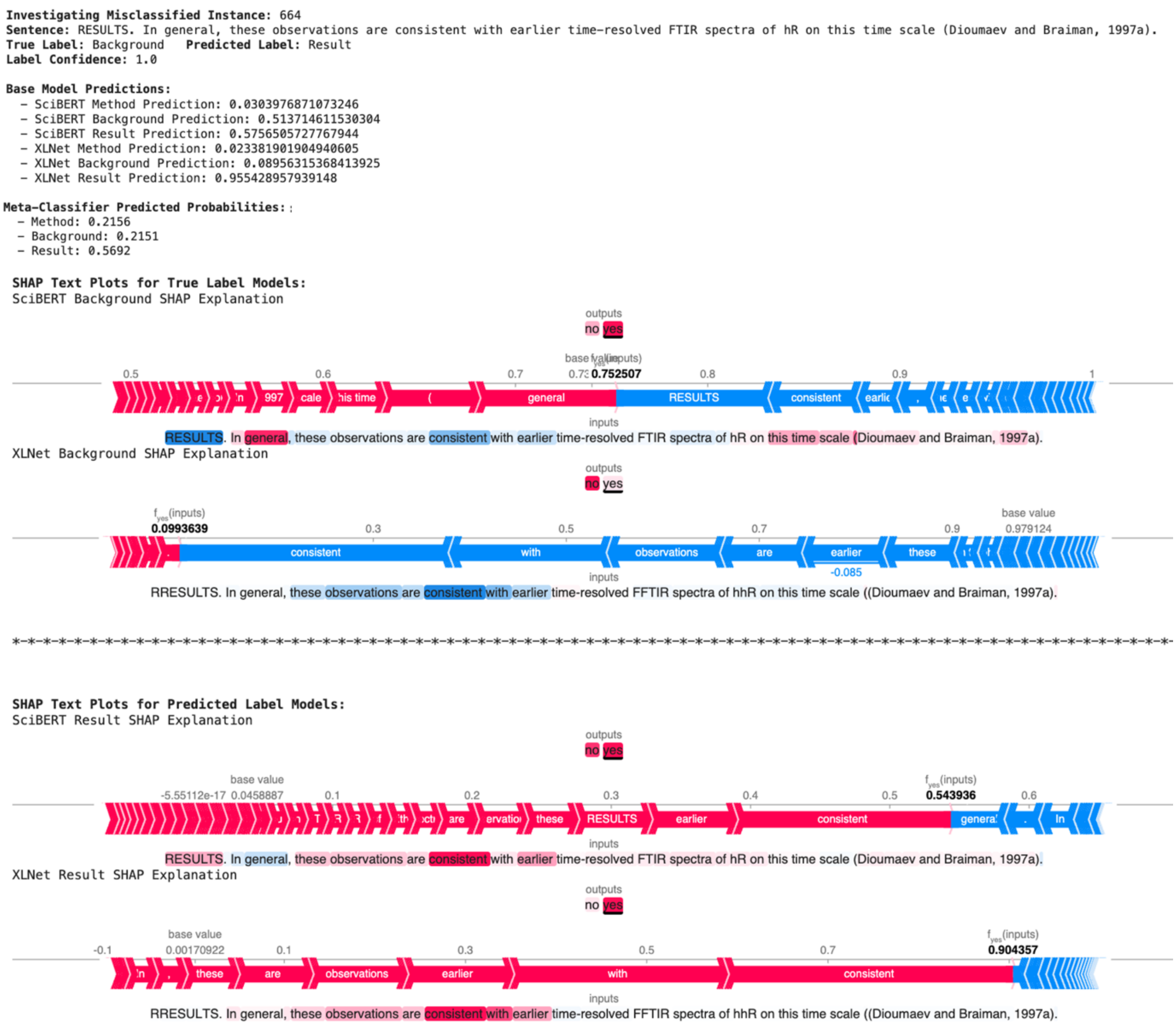

***Figure 10***. *Analysis of the citation context depicted in the image, which was misclassified as Result by our model despite belonging to the Background class. The report begins by presenting the sentence to be classified, accompanied by the individual prediction scores from the base models and the final output scores from the FFNN meta-classifier. The four text plots are organized as follows: the first two plots illustrate the tokens with the highest SHAP values as attributed by the two base models specifically trained to identify the true label (in this case, the Background label). The remaining two plots represent how the models trained for the incorrectly predicted label (in this case, Result) classified the same instance. Red elements indicate positive contributions toward the positive binary classification decision ("yes"), while blue elements highlight negative contributions to it. These color-coded features reveal which parts of the input influenced the classification outcome for each respective model, and their width reveal how strong their influence was.*

Furthermore, although the citation is labeled as a *Background*-type instance, we have reservations regarding the confidence score assigned by the annotators, further supporting the idea of possible subjectivity and inherent challenges in labelling ambiguous citation contexts through a rigid annotation schema, as stated in the previous section. Indeed, in our opinion, the sentence depicted in ***Figure 10*** seems to demonstrate a better alignment with the class identified by CiteFusion.

Another example of a misclassified citation is presented in ***Figure 11***, this time annotated with a confidence score of 0.61. CiteFusion incorrectly predicts the citation as belonging to the *Result* class, whereas it is labeled as a *Background* citation. Consistent with previous observations, SciBERT demonstrates alignment with the scientific vocabulary and domain-specific terminology, while XLNet models continue to disregard section titles, instead assigning higher scores to more general textual features. Furthermore, ***Figure 11*** also demonstrates visually our previous hypothesis on strong token-level attributions to features that are superficially related with the *Result* class, which is evident when looking at the scores attributed to "*consistent*" and "*support*" by both *Result* experts.

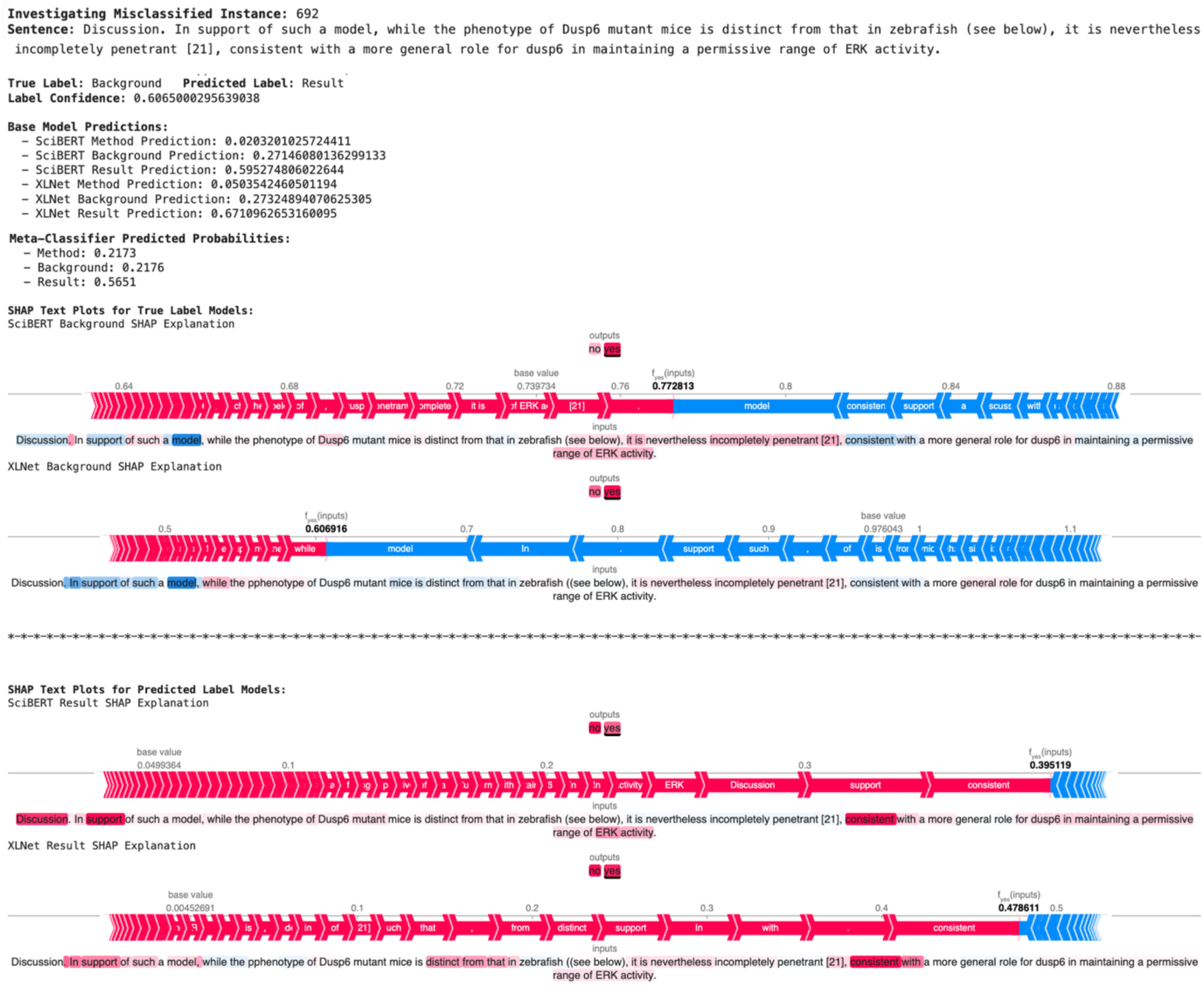

***Figure 11****. Analysis of the citation context depicted in the image, which was misclassified as Result by our model despite belonging to the Background class. Notably, the main difference with Figure 9 is the confidence score of the annotators for this citation context.*

Additionally, ***Table 14*** presents further misclassified citations along with their annotation confidence.

| ID | Section Title | Citation Context | Gold Label | Predicted Label | Annotation Confidence |
|---|---|---|---|---|---|
| *1* | Discussion | Several studies have now shown that DNA methylation changes with age at different genomic locations, the direction, and rate of change [9, 14, 33]. | *Method* | *Background* | 76% |
| *2* | Methods | Finally, the data regarding the obstetric history of the subjects were retrieved from the Danish Medical Birth Registry (MBR) which includes data on all live births, stillbirths and infant deaths in Denmark (Knudsen and Olsen, 1998). | *Background* | *Method* | 100% |
| *3* | Discussion | The rapid disappearance of cyclin mRNA after RNAi treatment is consistent with the currently-held notion that dsRNA targets the gene-specific destruction of mRNA in a wide variety of eukaryotic cells (Fire et al., 1998; Tabara et al., 1998; Tabara et al., 1999; Sanchez-Alvorado and Newmark, 1999; Grishok et al., 2000; Boscher and Labouesse, 2000; Klink and Wolniak, 2000). | *Background* | *Result* | 76% |
| *4* | 2. Methodology | …d (subject to g > 0) gives the analytic solution\nĝ = max {\n1 d 1 c (βML − ν) >(X>WX)(βML − ν)− 1, 0 }\n≈ max{zobs/d− 1, 0}, (5)\nhere zobs denotes the observed deviance (relative to the null model with β = 0), obtained from fitting a standard GLM to the data at hand (Copas, 1983, 1997). | *Method* | *Background* | 59% |

| ID | Section Title | Citation Context | Gold Label | Predicted Label | Annotation Confidence |
|---|---|---|---|---|---|
| *5* | Method | …allopathic effects of root exudates due to its high capacity to absorb organic compounds, which could thus alleviate the negative effects of allelochemicals and differentiate between their effect and that of resource depletion by neighbours (Callaway and Aschehoug 2000; Inderjit and Callaway 2003). | *Method* | *Background* | 58% |
| *6* | Introduction | The mineral compositions and oxygen isotope ratios of the particles studied by the PE team are very similar to those of the equilibrated LL chondrites (Nakamura et al. 2011; Yurimoto et al. 2011), which match the spectroscopic observations by the Hayabusa spacecraft (e. | *Background* | *Result* | 56% |
| *7* | Discussion | AGPs were proposed also to promote cell division and/or suppress cell elongation (Nothnagel 1997; Šamaj et al. 1998; 1999a; Serpe and Nothnagel 1999; Showalter 2001). | *Method* | *Background* | 62% |

***Table 14**. Additional examples of citation contexts from SciCite test split misclassified by CiteFusion (WS).*

In conclusion, while CiteFusion (WS) occasionally misclassifies certain instances due to mathematical elements and formulas contained in it (see ***Table 14***, context ***4***), or due to ambiguities arising from sentence structures or patterns of over-generalization as discussed in *Section 4.4.2*, it is important to note that the reliability of the annotations themselves may also be a contributing factor. This observation holds true even for instances where the confidence scores associated with the labels are relatively high.

Such findings underscore the need for continuous refinement and improvement in the quality of schemas and datasets employed within the domain of Citation Intent Classification. Furthermore, they highlight the absence of a definitive gold standard that can serve as a robust foundation for training and developing highly reliable models. Establishing such a standard remains an open challenge and a critical area for future research, emphasizing the importance of collaborative efforts to enhance dataset accuracy and consistency in this field.

## 4.5 Concluding Remarks on CiteFusion and the Role of Explainers

Although explainability was not the primary focus of this work, and despite SHAP having certain limitations – particularly its inability to fully account for feature dependencies in transformer-based models (Gohel et al., 2021) – our analyses provide valuable insights into the classification dynamics of CiteFusion. SciBERT and XLNet demonstrate complementary strengths at the base level: SciBERT excels in capturing domain-specific vocabulary, while XLNet is more sensitive to general terms. The metaclassifier effectively integrates these perspectives, weighing base model predictions according to their class alignment, and thereby achieving robust performance even in the presence of class imbalance – differently from other strategies focused on the specificity of each class, such as *StackingC* (Seewald, 2002) or the *ED+MLR adaptation of the Geometric Framework* (Wu et al., 2023) when applied to OVA frameworks.

Our error analysis, based on attribution mass, model agreement, and confusion matrices, shows that most misclassifications occur between *Method* and *Background*, and between *Background* and *Result*. These errors often arise when the rhetorical role of citation contexts is ambiguous. This underscores not only the inherent subjectivity of manual annotation, but also the limitations of current annotation schemes and datasets.

Finally, to the best of our knowledge, no prior studies have systematically explored the semantic features that characterize citation functions based on the intent categories described by SciCite and ACL-ARC schemes. Furthermore, we extended from the explanations of raw citation contexts by assessing the role of structural features (i.e., section titles) when provided as framing devices. Section titles are analyzed when employed by PLMs as semantical components of citation contexts, and their presence demonstrated to generally amplify the ability of base models to retrieve the constituting features of what class-specific citations are, also for underrepresented classes, thereby reducing models' reliance on exclusionary cues. Although this investigation falls outside the primary scope of our work, we ultimately provided model-dependent generalizations regarding the types of words, or tokens, that influence the understanding and perception of different citation intents across two different PLM architectures. These insights lay foundational groundwork for future research in this domain.

# 5 CIC Application

This chapter outlines the implementation of a web application designed to deploy the state-of-the-art (SOTA) ensemble models developed in this study. We decided to focus exclusively on CiteFusion models trained using SciCite. This choice is motivated by the greater reliability demonstrated by the models tuned on this dataset relative to ACL-ARC. Moreover, the classification schema adopted in SciCite is more robust and better suited to the requirements of our task, further supporting its selection for deployment.

The application is built using *Flask*[28], a lightweight and flexible web framework, which enables the seamless integration of various model variants, including those trained in WoS (without section titles) setting, into a user-friendly web-based platform, currently available[29]. Additionally, this chapter provides a general overview of the application's development process and its functional capabilities.

## 5.1 General Aim and Description

The application was developed within the *OpenCitations* infrastructure[30] as part of the *GraspOS* project[31]. This initiative aims to establish a robust data infrastructure and promote an ethical research assessment system grounded in Open Science (OS) principles across Europe. GraspOS contributes to the *European Open Science Cloud* (EOSC) ecosystem by integrating tools that monitor research service usage and advocate for the adoption of OS principles. Within this context, citation data plays a pivotal role in fostering openness, legitimacy, and knowledge sharing among academic communities, aligning closely with the objectives of both GraspOS and the broader Open Science movement.

The classifier presented in this study forms part of a larger application designed to automate the extraction and classification of citation intents from PDF files of scholarly works. Specifically, the classifier component, which was developed as part of this research and detailed in Paolini (2024a), encompasses a backend system capable of loading the various models generated in this work, preprocessing input data into a compatible format, and ultimately classifying citation contexts. The tool was implemented using Flask, Python, HTML, CSS, and JavaScript, and is currently accessible to the public in its Beta version. This development represents a significant step toward enhancing the automation and accuracy of citation intent classification within academic research workflows.

## 5.2 Design and Implementation

The central component of the software is the *Predictor* object, which manages prediction tasks, allocates GPU resources when available, and performs tokenization processes specifically tailored for both section-based and non-section-based data. Additionally, it generates a downloadable JSON file containing the classification results.

The backend architecture comprises several key components:

- *EnsembleClassifier*: This module loads base models from the server in accordance with the instructions provided by the *Predictor*.
- *DataProcessor*: Responsible for data preprocessing tasks, including reading, formatting, and performing structural integrity checks on the input data.
- *MetaClassifierSection* and *MetaClassifierNoSection*: These modules define the metaclassifier architectures designed to handle the two distinct data scenarios – section-based and non-section-based inputs.

The backend system processes the input data and employs ensemble models to classify citation contexts based on the selected operational mode. To ensure the reliability of classifications, and thereby avoid ambiguous cases such as the ones presented in *Section 4.4*, a human-defined threshold is integrated into the system. This threshold evaluates the output probabilities generated by the classification process. If none of the three main classes achieves a confidence score exceeding 90%, the classification is deemed *unreliable*. In such cases, the result is mapped to http://purl.org/spar/cito/citesForInformation, a general-purpose object property within the Citation Typing Ontology (CiTO) used to represent general-intent citations. This mapping enhances interoperability while addressing cases where classification confidence is insufficient because of ambiguous citation contexts.

[28] https://flask.palletsprojects.com/en/3.0.x/
[29] http://137.204.64.4:81/cic/
[30] https://opencitations.net/
[31] https://www.graspos.eu/

### 5.3 User Interface

The interface developed using Flask, HTML, CSS, and JavaScript, adopts a minimalistic design while ensuring sufficient user-friendliness in its initial release (see ***Figure 12***). It enables users to input data either as a list of text tuples, or as JSON files, to then select from the following analysis modes:

- *Mixed Mode*: This mode utilizes both ensembles to dynamically identify citations with and without section titles and applies the corresponding model for classification.
- *With Section Titles*: Designed for sentences that include section titles, this mode employs the ensemble trained on data containing section titles.
- *Without Section Titles*: Intended for raw citation contexts, this mode uses the ensemble trained on pure citations without section titles.

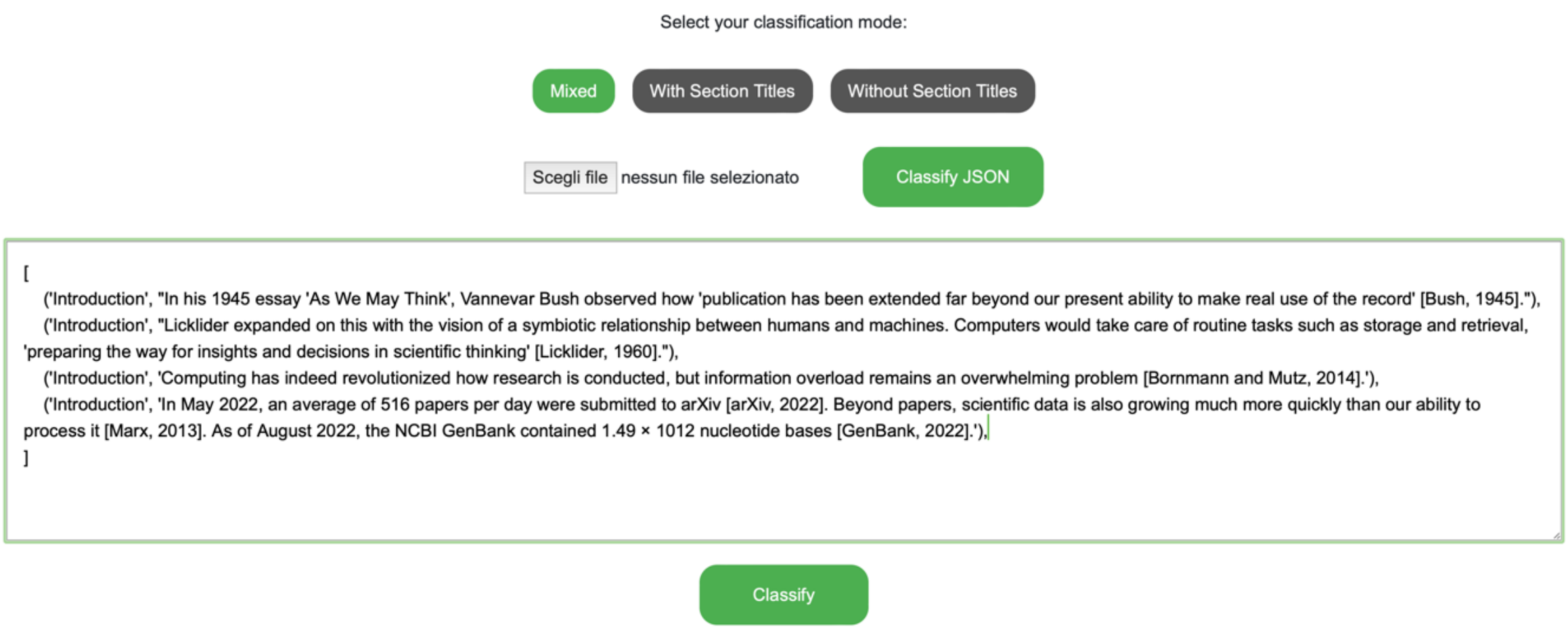

***Figure 12***. *User input interface. The user can decide the classification mode, and whether to upload data in text (with a list of tuples containing section titles – if possible – and citation contexts) or JSON format.*

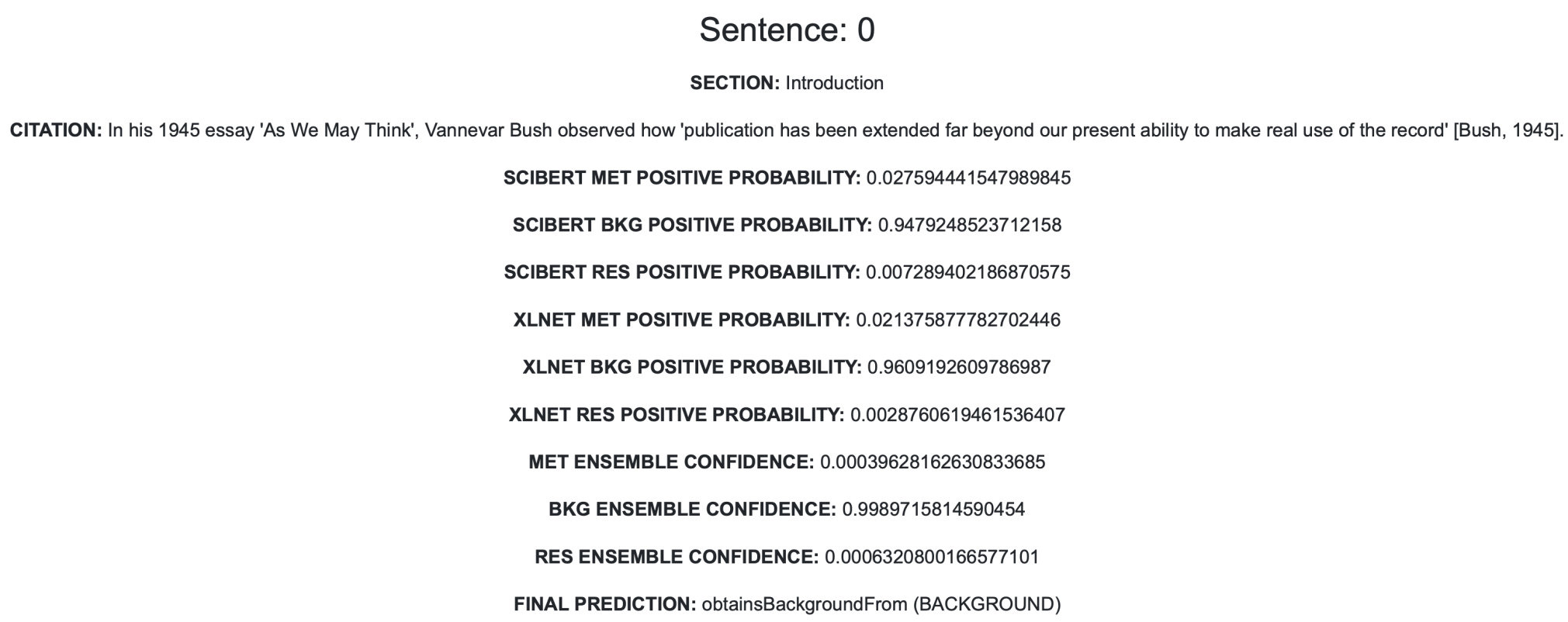

***Figure 13***. *Visualization of the classification results for a single sentence.*

After the backend loads the models and processes the input citation sentences, the classification results are displayed. Upon completion of the analysis, the application provides detailed results for all predictions. These results include confidence scores from level-0 models, the metaclassifier's confidence, and the final classification mapped to CiTO (***Figure 13***). Furthermore, the tool offers an option to download comprehensive JSON files containing both the citation data and the generated classification metadata.

# 6 Conclusion and Future Works

This concluding section provides a synthesis of the theoretical motivations, empirical findings, and broader implications of this work. It presents a conclusive analysis on the contribution of section titles as framing devices, examines the challenges and downstream consequences of misclassification in citation intent classification, and outlines ongoing and future improvements to the web-based application. In addition, the chapter discusses the technical implications of the proposed framework, and concludes with a summary of the research contributions and their significance for the advancement of automated citation analysis.

## 6.1 CiteFusion: A High-Level Summary

Understanding the motivations behind academic citations is crucial for analyzing scholarly discourse. Citation Intent Classification (CIC) aims to provide deeper insights into the underlying reasons for citations, thereby enhancing research evaluation and improving the transparency and reliability of academic communication. This study advances this objective by introducing a novel, couple-based one-vs-all (OVA) ensemble framework, which systematically pairs a domain-specific (*SciBERT*) and a general-purpose (*XLNet*) pretrained language model (PLM) for each intent class. The couple-based architecture is intended to harness the complementary strengths of the two base models, enabling the extraction of both domain-specific and general-purpose perspectives for each citation context, thereby providing a comprehensive interpretation of each input sentence. Furthermore, the design of *CiteFusion*, given the skewed distributions of the datasets employed in this study, is theoretically motivated by ensemble learning literature, which demonstrates that combining diverse models enhances robustness and generalizability in imbalanced settings (Khan et al., 2024; Liu et al., 2022). Additionally, the use of section titles as framing devices and their inclusion as prefixes to citation contexts is mainly motivated by frame theory (Goffman, 1974), as well as by other related works, as discussed in *Section 2.4*.

As illustrated in ***Tables 6*** and *7*, the ensemble models developed in this work outperform the previous state-of-the-art (SOTA) models for CIC on both the ***SciCite*** (**89, 60 macro-F1**) – surpassing the previous SOTA model *ImpactCite* (Mercier et al., 2021) – and ***ACL-ARC*** (**76, 24 macro-F1**) benchmarks. Empirically, the results obtained in this study provide strong support for our theoretical assumptions regarding the inclusion of section titles, which consistently improves performance in both datasets. Furthermore, the ensemble strategy and the OVA decomposition described in this work have proven to be particularly effective in addressing challenges associated with class imbalance. The findings reported in *Section 4.2* underscore the critical role of the stacked and binary-coupled architecture employed in this study. Additionally, we observed that the results remain stable and our models consistently outperform SOTA across all the 10 runs of the experiment for both SciCite and ACL-ARC in WS (with section titles) setting, even when mixed-precision training is utilized together with fine-grained evaluations, further supporting the robustness of our strategy.

To provide greater transparency into model decisions and annotation schemes, we used SHAP (Lundberg & Lee, 2017) to interpret both the level-0 binary classifiers and the FFNN metaclassifier components of CiteFusion, and mapped the original class labels to standard object properties from the Citation Typing Ontology (CiTO) (Peroni & Shotton, 2012). SHAP-based analyses revealed marked differences in how SciBERT and XLNet assign importance to tokens, reflecting their respective domain-specific versus general language training. SciBERT distributes attribution across a broader scientific vocabulary, while XLNet focuses on a smaller set of more generic terms. These complementary strengths are evident both globally and locally – as demonstrated and discussed in *Sections 4.3* and *4.4* –, and SHAP allowed us to clarify how the FFNN effectively merges predictions from the base models, contributing to CiteFusion's robust performance, and empirically supporting our assumption regarding the complementary strengths of the base PLMs employed. Importantly, SHAP also demonstrated to be valuable for interpreting and understanding misclassifications, providing insight into ambiguous cases.

While our use of SHAP was mainly aimed at explaining and validating CiteFusion's behavior, XAI tools could also be employed to refine model architectures, guide future development of annotation tools and datasets, or assess annotation quality and consistency. For example, attribution mass evaluations could support the development of synthetic citation intent datasets, an ongoing effort in this field (Zhang, Wang, Sheng, et al., 2023), or help to ensure consistency in annotation schemes. Uncovering deeper linguistic patterns linked to citation intent also represents a promising future direction, which may prove beneficial for research aimed at integrating citations and Large Language Models (Zhang, Wang, Wang, et al., 2023), as well as for other NLP-based research works on citations. However, these applications lie beyond the scope of this study and are left for future work.

In addition to these contributions, we developed and publicly released a preliminary version of a web-based application designed to automatically classify citation contexts using CiteFusion models trained on SciCite in both WS and WoS settings. Building upon the original SciCite schema, we extended the framework by mapping the three initial labels to CiTO object properties to enhance the interoperability of this application. Furthermore, guided by our SHAP analyses, we also introduced a more general citation function to classify citation contexts for which our model lacks sufficient confidence, with the aim to avoid ambiguous classifications, as described in *Section 5*.

### 6.2 Section Titles as Framing Devices: Future Directions

Our experimental evaluations on both SciCite and ACL-ARC datasets demonstrate that incorporating section titles as part of the input text significantly enhances the performance of citation intent classification. Across all experimental configurations, the WS setting consistently surpasses the WoS baseline. Including section titles within citation contexts provide for additional semantical cues that help in reshaping – and improving – the way in which level-0 models perceive input sentences. Overall, section titles effectively act as framing devices when prepended to input sentences, enhancing the models' ability to resolve possible ambiguities in citation contexts. This contextual enrichment contributes to the robust state-of-the-art performances achieved by CiteFusion.

The importance of section titles suggests several avenues for enhancing performance in the CIC domain and achieving more reliable systems. One promising direction involves integrating PLMs and ensemble strategies into the broader framework of *Neuro-Symbolic Systems* (Yu et al., 2023). Such frameworks enable classifications that are not solely reliant on the intrinsic semantics of textual contexts but also incorporate structured data sources, such as Knowledge Graphs (KGs), which facilitate logical reasoning operations. Knowledge Graphs constructed from citation data can encapsulate logical rules and relationships, which may prove to be highly informative for the CIC task. Given our findings on the effectiveness of integrating structural elements like section titles as framing devices, leveraging both the semantic and structural significance of these features within a broader context may theoretically yield further performance improvements.

Thereby, the generation of Knowledge Graph Embeddings (KGEs) (Cao et al., 2024) offers a practical means to integrate structured data with embeddings derived from PLMs (Pan et al., 2024). This integration can be achieved through various methodologies, although a detailed exploration of these techniques falls outside the scope of this study. Nevertheless, combining logical rules, symbolic representations, and sub-symbolic elements with the intrinsic semantics of citation contexts extracted via PLMs should theoretically provide a more structured and interpretable approach to CIC.

Enhanced explainability of model predictions, as emphasized by Pan et al. (2024), is also crucial for developing robust and reliable systems suitable for deployment in production environments. By bridging the gap between semantic understanding and structural/logical reasoning, such an approach holds significant potential for advancing the state-of-the-art in CIC and fostering greater transparency in automated citation analysis systems.

### 6.3 Citation Intents: Issues and Downstream Implications of Misclassifications

As discussed in the introductory part of this work, citations are inherently complex and heterogeneous: they reflect a wide range of motivations and rhetorical functions within scholarly writing (Pride, 2022; Tahamtan & Bornmann, 2019), and for this reason it is essential to avoid treating them all in the same way. The use of traditional citation-based metrics, which consider all citation objects as equivalent, leads to the risk of losing important facets of the scientific discourse, ultimately undermining the accuracy and fairness of bibliometric analyses.

Despite significant advances in schema design and annotation guidelines, the context-sensitive nature of citations continues to challenge automated classification systems, and this complexity is amplified by the diversity of scientific communication practices across disciplines, as well as by the existence of ambiguous cases where the boundaries between intent categories are blurred, as observed in *Section 4.4*.

In addition, misclassifying citation intents is not only a technical concern, but has significant implications for the interpretation and evaluation of scientific work. Indeed, as automated systems become increasingly employed in research assessment pipelines, errors in classifying citation intents may modify the perceived impact of scientific outputs (Nicholson et al., 2021; Tahamtan & Bornmann, 2019) . The downstream implications of these misclassifications can influence various applications and tasks, such as literature reviews, research assessment practices, or citation network analyses, where reliable intent annotation is essential to accurately analyze the research landscape (Nicholson et al., 2021; Pride, 2022; Tahamtan & Bornmann, 2019).

Despite recent advancements in NLP and Machine Learning, many challenges remain in the conceptual definition of a reliable annotation framework for the field, but also in the construction of a diverse, complete, and interoperable dataset for CIC. The inherent subjectivity of annotations and the lack of a well-defined and interoperable schema, together with the subtleties of the rhetorical functions of citations, still lead to various misclassifications that propagate through automated systems. We argue that a standardized definition of citation intents and a broader understanding of their impact are essential to recognize the potential and limitations of these conceptual tools. In addition, a transparent description of the classification dynamics of NLP models designed to solve the CIC task would largely benefit the field by offering reliable classification outcomes while explaining model failures, possibly helping researchers also in assessing the validity and reliability of the tools and applications they develop and use.

### 6.4 Citation Intent Classifier: Future Releases

The web-based application discussed in *Section 5* is currently in its initial release and has undergone significant improvements since its preliminary version, which was developed as part of the master's thesis "*Enhanced Citation Intent Classification with Population-based Training, Ensemble Strategies, and Language Models*" (Paolini, 2024c). Building on the work presented thus far, we aim to enhance the tool by incorporating explanations for classified sentences using SHAP, which provides tools and functions that enable dynamic and engaging visualization of results, such as the ones shown in ***Figures 10*** and ***11***. Furthermore, users will be provided with the possibility to manually set different thresholds to distinguish between reliable and unreliable classifications, and thereby account for a certain level of ambiguity in classification. By integrating these explanations with user-defined thresholds tailored to specific needs, we aim to create a more robust, transparent, and user-friendly tool that can potentially be adapted to a variety of use cases.

In addition to these improvements, we plan to further enhance the performance of the backend models by leveraging an expanded dataset for training, and incorporating KGEs into the classification pipeline. This integration is expected to improve the system's ability to capture structured relationships within citation data, thereby enhancing classification accuracy and interpretability.

### 6.5 Final Remarks

In summary, this research positions the field of Citation Intent Classification at a critical juncture between the need to deepen the understanding of academic discourse and the advancements in Natural Language Processing and Artificial Intelligence. A central contribution of this study is the emphasis on the necessity of a comprehensive and meticulously curated dataset for the CIC task, supported by empirical evidence demonstrating the key role that data quantity and quality play in determining outcomes. For instance, the low scores observed across all models for the *Background* classes in both SciCite and ACL-ARC in terms of comprehensibility and attribution mass highlights this dependency. Similarly, the challenges associated with the underrepresented classes (*Result* in SciCite, and *Extends*, *Motivation*, and *Future* in ACL-ARC) underscore the limitations posed by an insufficient number of data points for developing robust models. This is evident not only in the class-specific performance metrics for these intents, but also in the aggregated explanations generated using SHAP.

Despite these challenges, another key contribution of this work is the development of *CiteFusion*, an ensemble strategy that achieves state-of-the-art performance for the Citation Intent Classification task on both the SciCite and ACL-ARC datasets. From a technical perspective, this framework demonstrates that combining domain-specific and general-purpose language models within an ensemble with OVA decomposition can robustly address class imbalance and semantic ambiguity in multi-class classification tasks. This architecture, together with SHAP-based analyses and the explicit use of section titles as framing devices, offers a blueprint for enhancing performance and transparency in other challenging NLP applications.

To ensure transparency and reproducibility, we publicly release all code (Paolini, 2024d) and models developed during this study, including versions tailored for both with section titles (WS) and without section titles (WoS) settings:

- CiteFusion $_{\text{SciCite}}$ (WS) (Paolini, 2024b),
- CiteFusion $_{\text{SciCite}}$ (WoS) (Paolini, 2024a),
- CiteFusion $_{\text{ACL-ARC}}$ (WS) (Paolini, 2025b),
- CiteFusion $_{\text{ACL-ARC}}$ (WoS) (Paolini, 2025a).

On a practical level, the public release of our code, models, and web-based application empowers researchers and practitioners to implement and adapt these solutions in real-world bibliometric analysis, digital libraries, and research evaluation systems, thereby advancing the automated study of scholarly communication.

This research establishes a foundation for advancing the Citation Intent Classification task and guiding the development of future datasets. These efforts should be closely aligned with the evolving field of Explainable AI, as this study demonstrates that XAI not only enhances the interpretability of model behavior but also plays a crucial role in identifying potential biases within the developed tools. Additionally, we provide a clearer – even if model-dependent – understanding of citation intents by highlighting the tokens and words that different models consider relevant, as well as providing an average mass of distinctive intent-specific features revealed by our models. This analysis enables a preliminary semantic characterization of the intents used in SciCite and ACL-ARC, while also revealing significant overlaps between classes (e.g., *Background* with *Background*, and *Method* with *Uses*) across datasets. These overlaps are further enriched through annotation with CiTO object properties (Peroni & Shotton, 2012), facilitating their intersection and comparative analysis. In conclusion, by contributing to the field of Citation Intent Classification, this work fosters interdisciplinary dialogue and lays the groundwork for future innovations and investigations in the domain.

## Declarations

### Funding and Competing Interests

This work has been partially funded by the European Union's Horizon Europe framework programme under Grant Agreements No 101095129 (*GraspOS Project*) and No 101188018 (*GRAPHIA Project*).

The authors have no relevant financial or non-financial interests to disclose.

# Appendix

## A.1 XAI Experimental Results

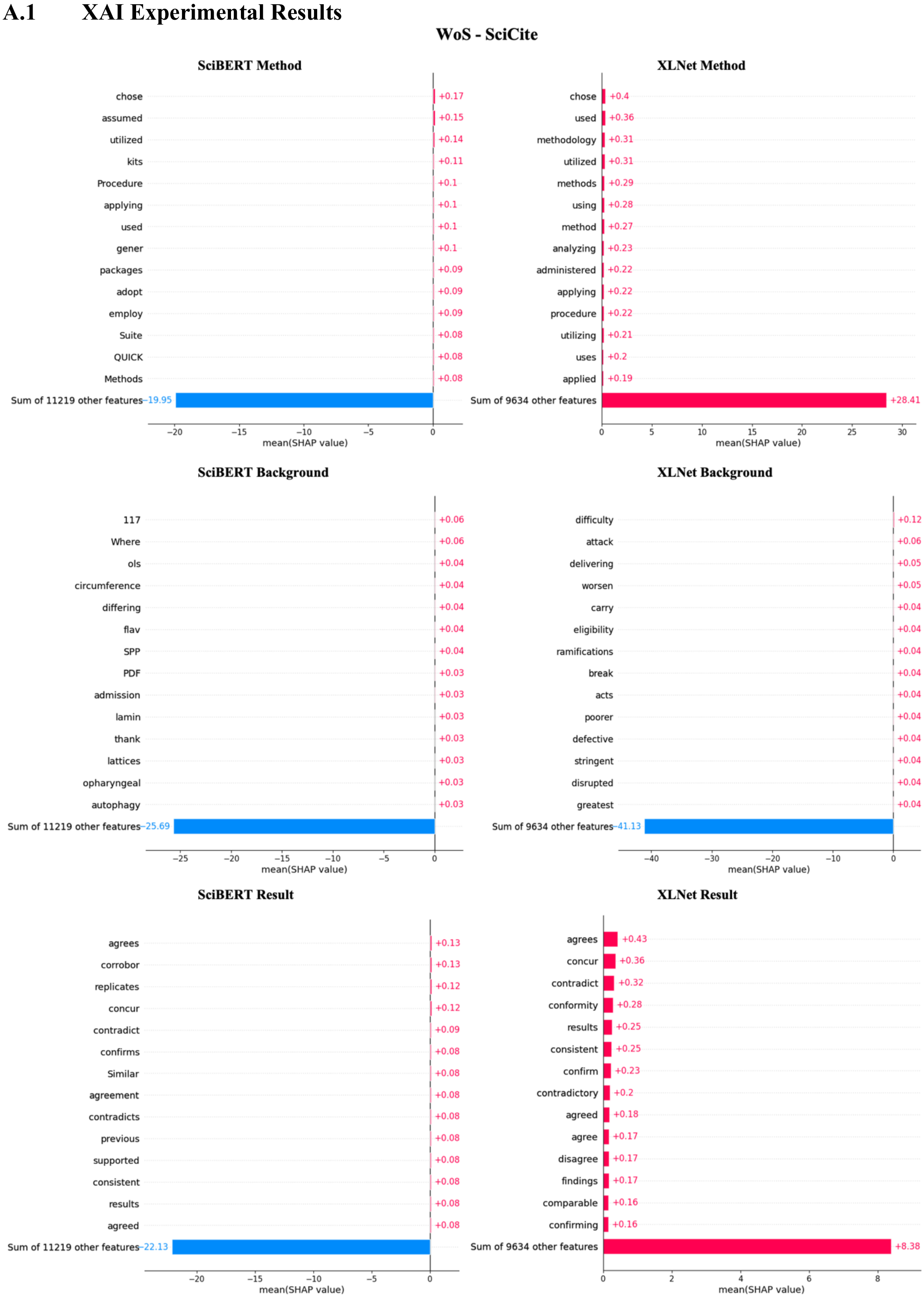

***Figure A.1.*** *This figure displays the top 15 tokens, considered as features based on their SHAP scores, that contribute positively to the classification of citations into each specific class of the SciCite dataset in WoS setting. Each plot illustrates the contributions of a particular model architecture and the tokens to the classification of the specific class, reported in each title (6 figures: 2 PLMs and 3 classes).*

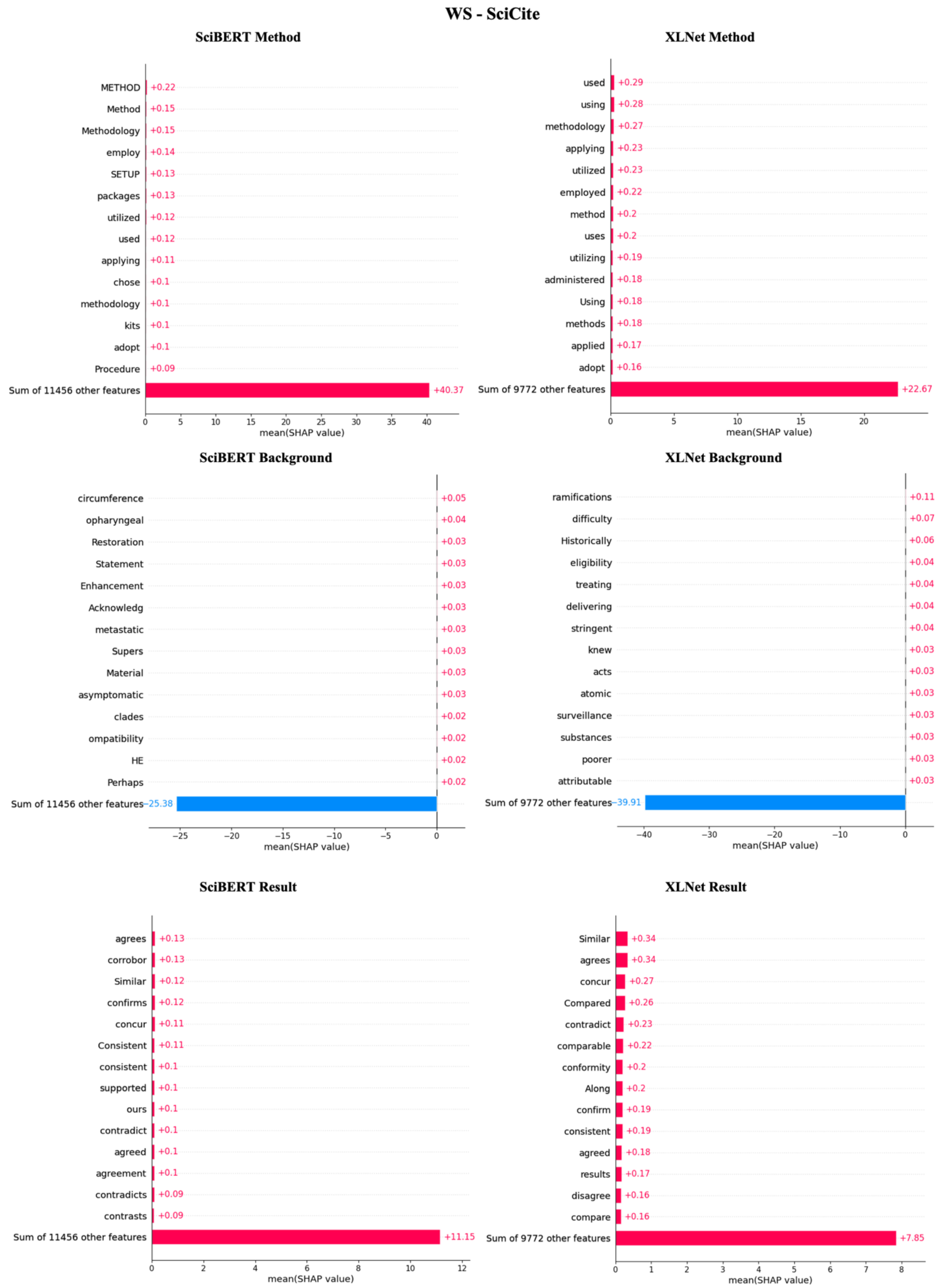

***Figure A.2.*** *This figure displays the top 15 tokens, considered as features based on their SHAP scores, that contribute positively to the classification of citations into each specific class of the SciCite dataset in WS setting. Each plot illustrates the contributions of a particular model architecture and the tokens to the classification of the specific class, reported in each title (6 figures: 2 PLMs and 3 classes).*

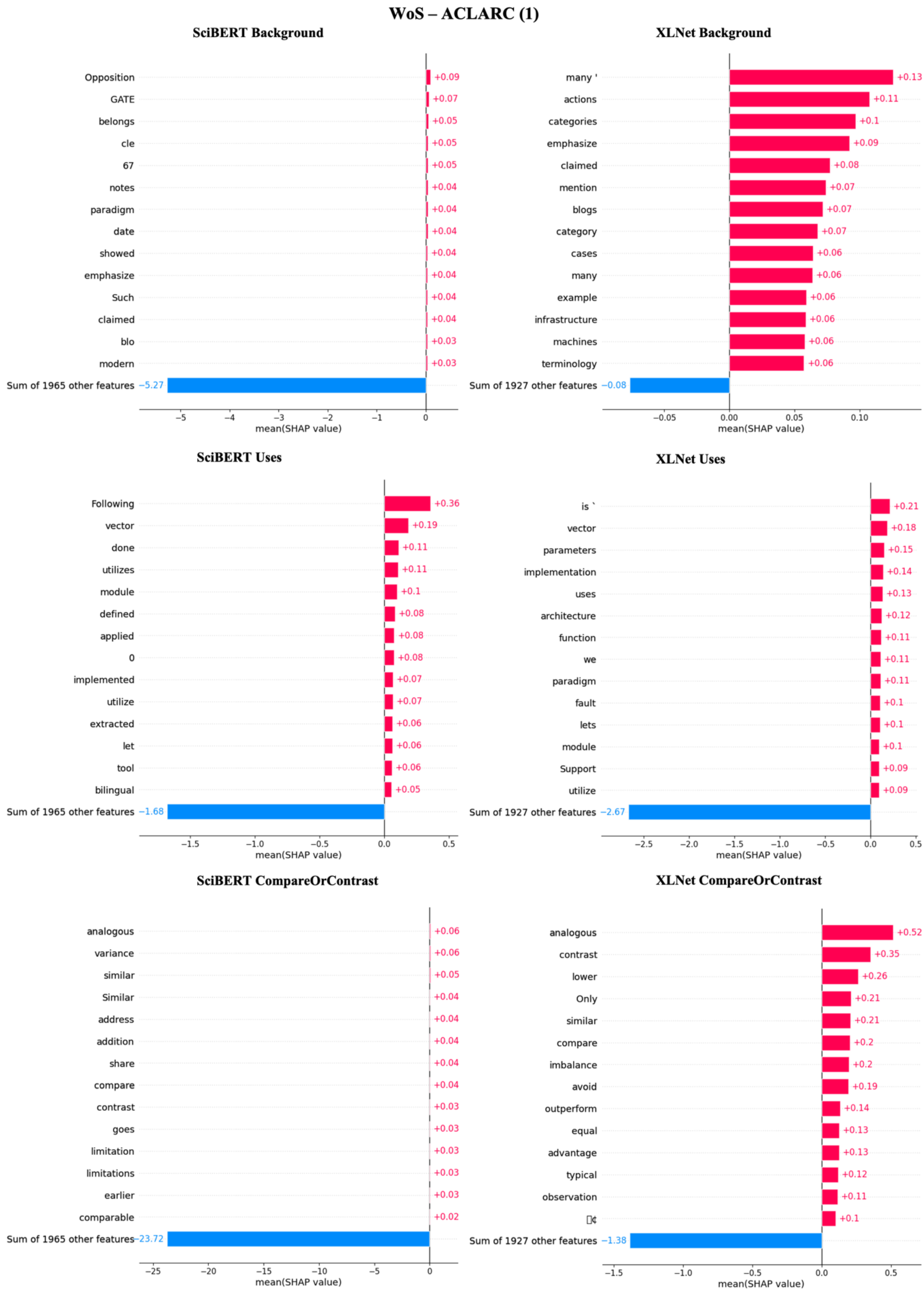

***Figure A.3.*** *This figure displays the top 15 tokens, considered as features based on their SHAP scores, that contribute positively to the classification of citations into each of the first 3 classes (Background, Uses, CompareOrContrast) of the ACL-ARC dataset in WoS setting. Each plot illustrates the contributions of a particular model architecture and the tokens to the classification of the specific class, reported in each title (6 figures: 2 PLMs and 3 classes). Second part in Figure A4.*

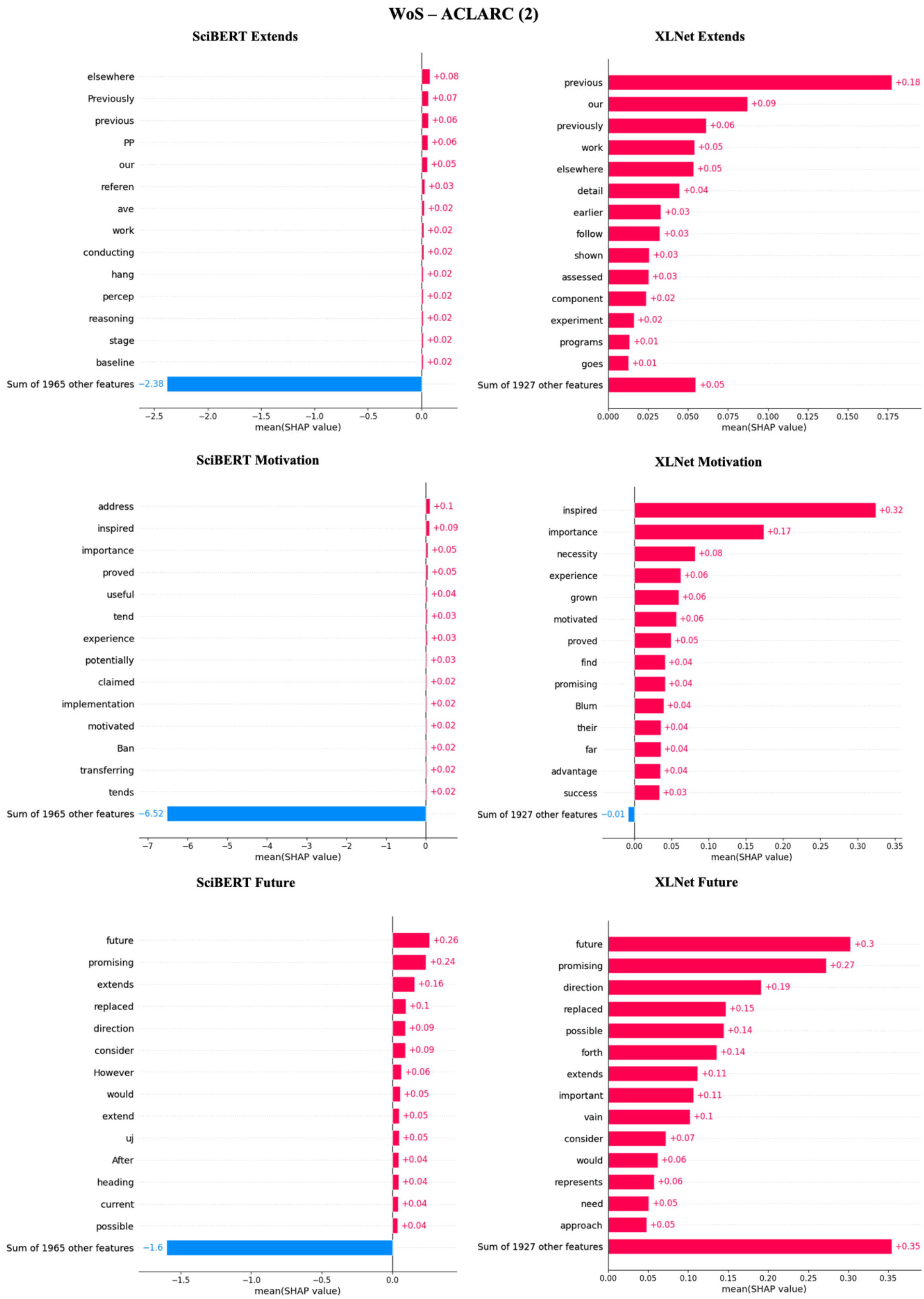

***Figure A.4.*** *This figure displays the top 15 tokens, considered as features based on their SHAP scores, that contribute positively to the classification of citations into each of the remaining 3 classes (Extends, Motivation, Future) of the ACL-ARC dataset in WoS setting, w.r.t. the 3 reported in Figure A3. Each plot illustrates the contributions of a particular model architecture and the tokens to the classification of the specific class, reported in each title (6 figures: 2 PLMs and 3 classes). First part in Figure A3.*

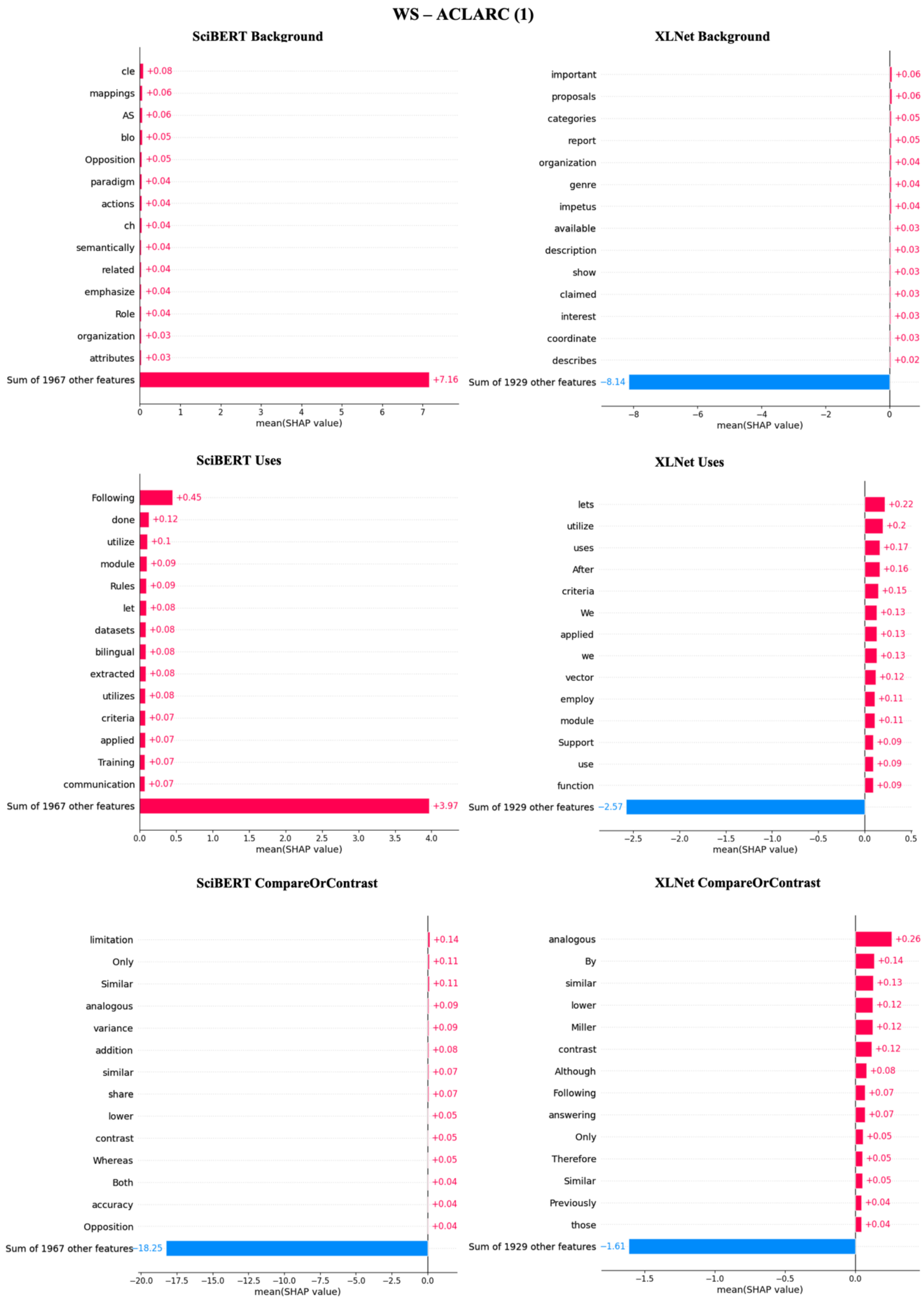

***Figure A.5.*** *This figure displays the top 15 tokens, considered as features based on their SHAP scores, that contribute positively to the classification of citations into each of the first 3 classes (Background, Uses, CompareOrContrast) of the ACL-ARC dataset in WS setting. Each plot illustrates the contributions of a particular model architecture and the tokens to the classification of the specific class, reported in each title (6 figures: 2 PLMs and 3 classes). Second part in Figure A6.*

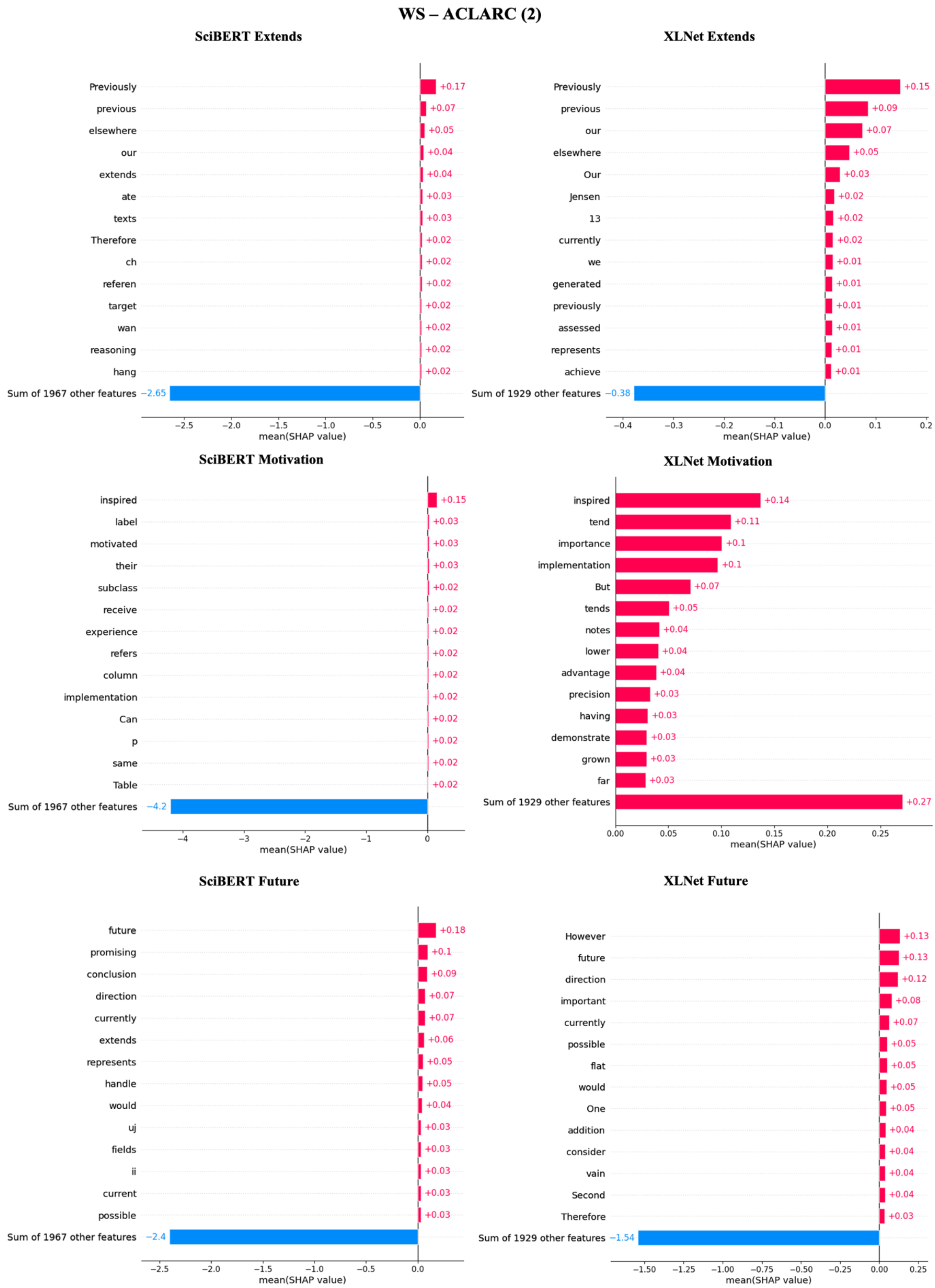

***Figure A.6.*** *This figure displays the top 15 tokens, considered as features based on their SHAP scores, that contribute positively to the classification of citations into each of the remaining 3 classes (Extends, Motivation, Future) of the ACL-ARC dataset in WS setting, w.r.t. the 3 reported in Figure A5. Each plot illustrates the contributions of a particular model architecture and the tokens to the classification of the specific class, reported in each title (6 figures: 2 PLMs and 3 classes). First part in Figure A5.*

## A.2 Computational Instability Details on Minima

SciCite

| Run | SciBERT-based Models | | | XLNet-based Models | | |
|---|---|---|---|---|---|---|
| | Met | Bkg | Res | Met | Bkg | Res |
| 0 | 0.2385 | 0.3245 | 0.1330 | 0.3139 | 0.2461 | 0.1242 |
| 1 | 0.2381 | 0.3245 | 0.1331 | 0.3139 | 0.2461 | 0.1242 |
| 2 | 0.2387 | 0.3245 | 0.1329 | 0.3139 | 0.2461 | 0.1242 |
| 3 | 0.2386 | 0.3244 | 0.1327 | 0.3139 | 0.2461 | 0.1242 |
| 4 | 0.2383 | 0.3245 | 0.1328 | 0.3139 | 0.2461 | 0.1242 |
| 5 | 0.2385 | 0.3245 | 0.1329 | 0.3139 | 0.2461 | 0.1242 |
| 6 | 0.2385 | 0.3246 | 0.1330 | 0.3139 | 0.2461 | 0.1242 |
| 7 | 0.2383 | 0.3245 | 0.1330 | 0.3139 | 0.2461 | 0.1242 |
| 8 | 0.2383 | 0.3245 | 0.1329 | 0.3139 | 0.2461 | 0.1242 |
| 9 | 0.2382 | 0.3244 | 0.1331 | 0.3139 | 0.2461 | 0.1242 |

***Table A.1.*** *Detailed results of computational instability analyses for minima in validation losses within the entire fine-tuning loops recorded for all the level-0 models in each different run (across the 10 total runs) of the same experiment on the SciCite dataset. Models are described by the class on which they were tuned, and by their architecture. Abbreviations reported in the table for these classes are: Met (Method), Bkg (Background), and Res (Result Comparison).*

ACL-ARC

| Run | SciBERT-based Models | | | | | | XLNet-based Models | | | | | |
|---|---|---|---|---|---|---|---|---|---|---|---|---|
| | Bkg | Use | CoC | Ext | Mot | Fut | Bkg | Use | CoC | Ext | Mot | Fut |
| 0 | 0.3173 | 0.1565 | 0.2788 | 0.0872 | 0.0861 | 0.0127 | 0.3913 | 0.2195 | 0.2887 | 0.1106 | 0.1745 | 0.0075 |
| 1 | 0.3172 | 0.1574 | 0.2724 | 0.0871 | 0.0916 | 0.0126 | 0.3913 | 0.2195 | 0.2887 | 0.1106 | 0.1745 | 0.0075 |
| 2 | 0.3171 | 0.1570 | 0.2738 | 0.0871 | 0.0924 | 0.0126 | 0.3913 | 0.2195 | 0.2887 | 0.1106 | 0.1745 | 0.0075 |
| 3 | 0.3171 | 0.1572 | 0.2747 | 0.0873 | 0.0792 | 0.0086 | 0.3913 | 0.2195 | 0.2887 | 0.1106 | 0.1745 | 0.0075 |
| 4 | 0.3171 | 0.1573 | 0.2745 | 0.0872 | 0.0792 | 0.0126 | 0.3913 | 0.2195 | 0.2887 | 0.1106 | 0.1745 | 0.0075 |
| 5 | 0.3172 | 0.1568 | 0.2745 | 0.0872 | 0.0852 | 0.0132 | 0.3913 | 0.2195 | 0.2887 | 0.1106 | 0.1745 | 0.0075 |
| 6 | 0.3172 | 0.1569 | 0.2748 | 0.0872 | 0.0796 | 0.0126 | 0.3913 | 0.2195 | 0.2887 | 0.1106 | 0.1745 | 0.0075 |
| 7 | 0.3172 | 0.1566 | 0.2745 | 0.0871 | 0.0788 | 0.0127 | 0.3913 | 0.2195 | 0.2887 | 0.1106 | 0.1745 | 0.0075 |
| 8 | 0.3172 | 0.1570 | 0.2747 | 0.0873 | 0.0787 | 0.0084 | 0.3913 | 0.2195 | 0.2887 | 0.1106 | 0.1745 | 0.0075 |
| 9 | 0.3172 | 0.1574 | 0.2745 | 0.0874 | 0.0792 | 0.0073 | 0.3913 | 0.2195 | 0.2887 | 0.1106 | 0.1745 | 0.0075 |

***Table A.2.*** *Detailed results of computational instability analyses for minima in validation losses within the entire fine-tuning loops recorded for all the level-0 models in each different run (across the 10 total runs) of the same experiment on the ACL-ARC dataset. Models are described by the class on which they were tuned, and by their architecture. Abbreviations reported in the table for these classes are: Bkg (Background), Use (Uses), CoC (CompareOrContrast), Ext (Extends), Mot (Motivation), and Fut (Future).*

### A.3 Visual Comparison of Mean Signed Attribution Masses in Intent-specific Predictions

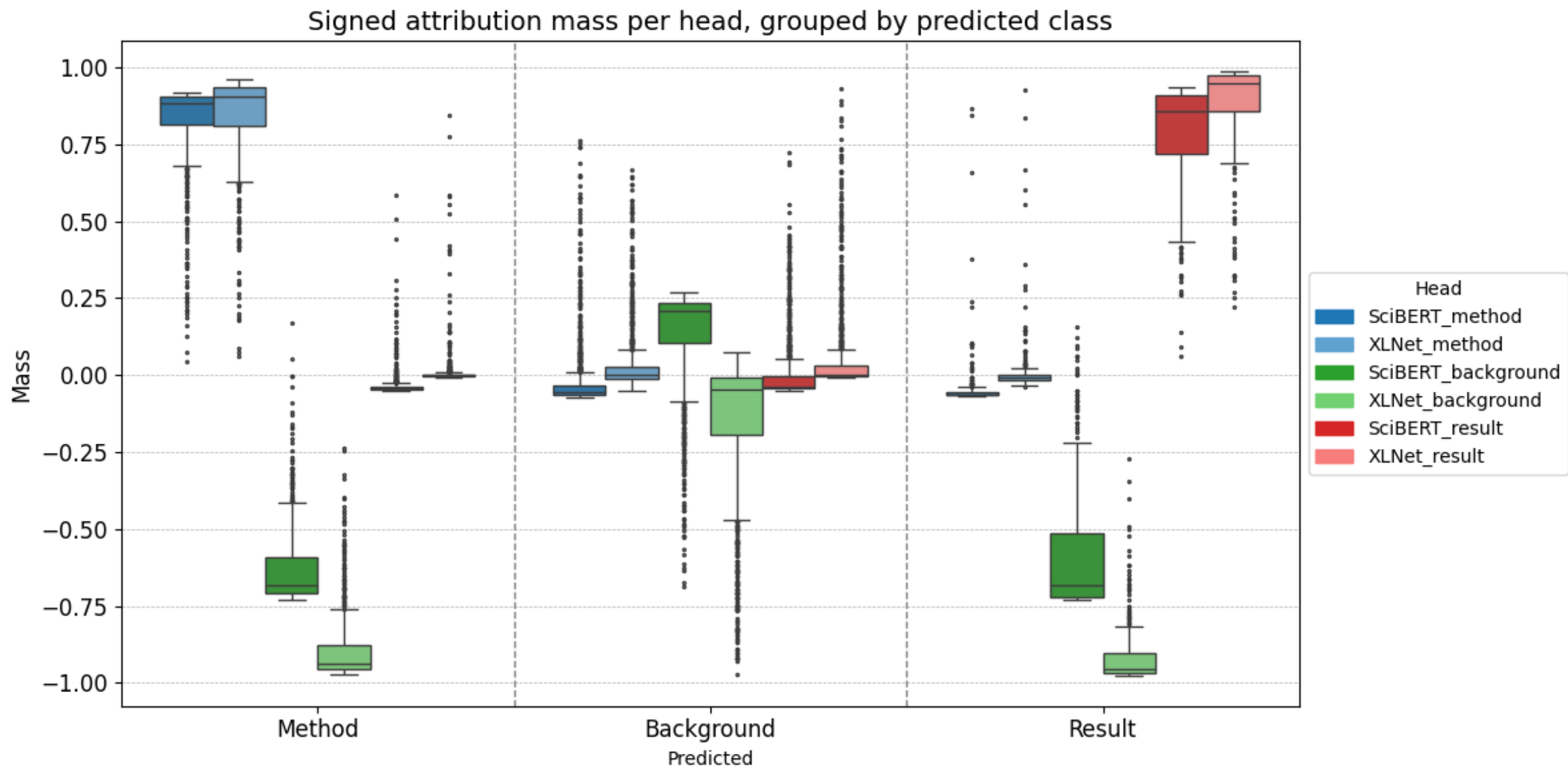

***Figure A.7.*** *This figure presents boxplots of signed attribution mass per expert, grouped by predicted class, as described in **Table 10** (Section 4.2.1). Each box represents the distribution of attribution masses assigned by SciBERT and XLNet models, each fine-tuned for one of the three citation intent in SciCite, when the ensemble predicts a specific class label.*
*When the predicted class is <u>Method</u>, attribution masses from Method-tuned experts (in blue) are strongly positive and show tight internal agreement, while Background-tuned models (in green) assign consistently negative masses, indicating mutual rejection. Result experts (in red) display instead near-zero attributions, reflecting their marginal role in Method predictions.*
*A similar pattern emerges for the <u>Result</u> class: Result experts show high positive attribution, Background models assign negative values, and Method experts remain close to zero.*
*For <u>Background</u> predictions, attribution masses from Background-tuned models are positive but with greater variance, while Method- and Result-tuned experts generally assign negative or near-zero attributions which, as demonstrated by **Figure 9** (Section 4.4.1), do not represent marginal values but negatively correlated scores.*